\documentclass[runningheads]{llncs}
\usepackage{graphicx}
\usepackage{amsmath,amssymb} % define this before the line numbering.
\usepackage{color}
\usepackage{mathrsfs}

\usepackage{epstopdf}
\usepackage{epsfig}
\usepackage{graphicx}
\usepackage{algorithm}
\usepackage{algorithmic}
\usepackage{amssymb}
\usepackage{mathrsfs} % For LaTeX2e
\usepackage{amsmath,bm}
\usepackage{array,multirow}
\usepackage{mdwlist}
\usepackage{paralist}
\usepackage{url}
\usepackage{color}
\usepackage{setspace}
\usepackage{textcomp}
\usepackage{bbding}
\usepackage{booktabs}
\usepackage{threeparttable}
\usepackage{float}
\usepackage{subfig}

\newcommand\Tstrut{\rule{0pt}{2.6ex}}       % "top" strut
\newcommand\Bstrut{\rule[-0.6ex]{0pt}{0pt}} % "bottom" strut
\newcommand{\TBstrut}{\Tstrut\Bstrut} % top&bottom struts

\usepackage{multirow}

\usepackage[width=122mm,left=12mm,paperwidth=146mm,height=193mm,top=12mm,paperheight=217mm]{geometry}
\newcommand{\mathbbm}[1]{\text{\usefont{U}{bbm}{m}{n}#1}} % from mathbbm.sty

\begin{document}
\title{Distractor-aware Siamese Networks for Visual Object Tracking}
% Replace with your title

\titlerunning{DaSiameseRPN}
% Replace with a meaningful short version of your title
%
\authorrunning{Zheng Zhu, Qiang Wang, Bo Li, Wei Wu, Junjie Yan and Weiming Hu}
% Replace with shorter version of the author list. If there are more authors than fits a line, please use A. Author et al.

\author{Zheng Zhu$^{* 1,2}$ \and Qiang Wang$^{* 1,2}$ \and Bo Li$^{* 3}$ \and Wei Wu$^{3}$ \and \\
Junjie Yan$^{3}$ \and Weiming Hu$^{1,2}$}
% Replace with shorter version of the author list. If there are more authors than fits a line, please use A. Author et al.
%
\footnotetext{*The first three authors contributed equally to this work. This work is done when Zheng Zhu and Qiang Wang are interns at SenseTime Group Limited.}
\institute{$^{1}$University of Chinese Academy of Sciences, Beijing, China\\
$^{2}$Institute of Automation, Chinese Academy of Sciences, Beijing, China\\
$^{3}$SenseTime Group Limited, Beijing, China}
\maketitle              % typeset the header of the contribution
\begin{abstract}

Recently, Siamese networks have drawn great attention in visual tracking community because of their balanced accuracy and speed.
However, features used in most Siamese tracking approaches can only discriminate foreground from the non-semantic backgrounds.
The semantic backgrounds are always considered as distractors, which hinders the robustness of Siamese trackers.
In this paper, we focus on learning distractor-aware Siamese networks for accurate and long-term tracking.
To this end, features used in traditional Siamese trackers are analyzed at first.
We observe that the imbalanced distribution of training data makes the learned features less discriminative.
During the off-line training phase, an effective sampling strategy is introduced to control this distribution and make the model focus on the semantic distractors.
During inference, a novel distractor-aware module is designed to perform incremental learning, which can effectively transfer the general embedding to the current video domain.
In addition, we extend the proposed approach for long-term tracking by introducing a simple yet effective local-to-global search region strategy.
Extensive experiments on benchmarks show that our approach significantly outperforms the state-of-the-arts, yielding 9.6\% relative gain in VOT2016 dataset and 35.9\% relative gain in UAV20L dataset.
The proposed tracker can perform at 160 FPS on short-term benchmarks and 110 FPS on long-term benchmarks. The code is available at \url{https://github.com/foolwood/DaSiamRPN}.
\keywords{Visual Tracking \and Distractor-aware \and Siamese Networks}
\end{abstract}

\section{Introduction}

Visual object tracking, which locates a specified target in a changing video sequence automatically, is a fundamental problem in many computer vision topics such as visual analysis, automatic driving and pose estimation. A core problem of tracking is how to detect and locate the object accurately and efficiently in challenging scenarios with occlusions, out-of-view, deformation, background cluttering and other variations~\cite{OTB2015}.

Recently, Siamese networks, which follow a tracking by similarity comparison strategy, have drawn great attention in visual tracking community because of favorable performance~\cite{SINT,GOTURN,SiamFC,RASNet,CFNet,DSiam,EDCF,SiamRPN}. SINT~\cite{SINT}, GOTURN~\cite{GOTURN}, SiamFC~\cite{SiamFC} and RASNet~\cite{RASNet} learn a priori deep Siamese similarity function and use it in a run-time fixed way. CFNet~\cite{CFNet} and DSiam~\cite{DSiam} can online update the tracking model via a running average template and a fast transformation learning module, respectively. SiamRPN~\cite{SiamRPN} introduces a region proposal network after the Siamese network, thus formulating the tracking as a one-shot local detection task.

Although these tracking approaches obtain balanced accuracy and speed, there are 3 problems that should be addressed: firstly, features used in most Siamese tracking approaches can only discriminate foreground from the non-semantic background.
The semantic backgrounds are always considered as distractors, and the performance can not be guaranteed when the backgrounds are cluttered.
Secondly, most Siamese trackers can not update the model~\cite{SINT,GOTURN,SiamFC,RASNet,SiamRPN}. Although their simplicity and fixed-model nature lead to high speed, these methods lose the ability to update the appearance model online which is often critical to account for drastic appearance changes in tracking scenarios. %Even in CFNet~\cite{CFNet} and DSiam~\cite{DSiam}, updating methods may result in the drift of the tracker.
Thirdly, recent Siamese trackers employ a local search strategy, which can not handle the full occlusion and out-of-view challenges.

In this paper, we explore to learn Distractor-aware Siamese Region Proposal Networks (DaSiamRPN) for accurate and long-term tracking. SiamFC uses a weighted loss function to eliminate class imbalance of the positive and negative examples. However, it is inefficient as the training procedure is still dominated by easily classified background examples. In this paper, we identify that the imbalance of the \textit{non-semantic} background and \textit{semantic} distractor in the training data is the main obstacle for the representation learning. As shown in Fig.~\ref{fig:visual}, the response maps on the SiamFC can not distinguish the people, even the athlete in the white dress can get a high similarity with the target person. High quality training data is crucial for the success of end-to-end learning tracker. We conclude that the quality of the representation network heavily depends on the distribution of training data. In addition to introducing positive pairs from existing large-scale detection datasets, we explicitly generate diverse semantic negative pairs in the training process. To further encourage discrimination, an effective data augmentation strategy customizing for visual tracking are developed.

After the offline training, the representation networks can generalize well to most categories of objects, which makes it possible to track general targets.
During inference, classic Siamese trackers only use nearest neighbour search to match the positive templates, which might perform poorly when the target undergoes significant appearance changes and background clutters.
Particularly, the presence of similar looking objects (distractors) in the context makes the tracking task more arduous.
To address this problem, the surrounding contextual and temporal information can provide additional cues about the targets and help to maximize the discrimination abilities.
In this paper, a novel distractor-aware module is designed, which can effectively transfer the general embedding to the current video domain and incrementally catch the target appearance variations during inference.

Besides, most recent trackers are tailored to short-term scenario, where the target object is always present.
These works have focused exclusively on short sequences of a few tens of seconds, which is poorly representative of practitioners' needs.
Except the challenging situations in short-term tracking, severe out-of-view and full occlusion introduce extra challenges in long-term tracking.
Since conventional Siamese trackers lack discriminative features and adopt local search region, they are unable to handle these challenges. Benefiting from the learned distractor-aware features in DaSiamRPN, we extend the proposed approach for long-term tracking by introducing a simple yet effective local-to-global search region strategy. This significantly improves the performance of our tracker in out-of-view and full occlusion challenges.

We validate the effectiveness of proposed DaSiamRPN framework on extensive short-term and long-term tracking benchmarks: VOT2016~\cite{VOT2016}, VOT2017~\cite{VOT2017}, OTB2015~\cite{OTB2015}, UAV20L and UAV123~\cite{UAV}.
On short-term VOT2016 dataset, DaSiamRPN achieves a 9.6\% relative gain in Expected Average Overlap compared to the top ranked method ECO~\cite{ECO}.
On long-term UAV20L dataset, DaSiamRPN obtains 61.7\% in Area Under Curve which outperforms the current best-performing tracker by relative 35.9\%.
Besides the favorable performance, our tracker can perform at far beyond real-time speed: 160 FPS on short-term datasets and 110 FPS on long-term datasets.
All these consistent improvements demonstrate that the proposed approach establish a new state-of-the-art in visual tracking.

\subsection{Contributions}

The contributions of this paper can be summarized in three folds as follows:

1, The features used in conventional Siamese trackers are analyzed in detail. And we find that the imbalance of the \textit{non-semantic} background and \textit{semantic} distractor in the training data is the main obstacle for the learning. %To the best

2, We propose a novel Distractor-aware Siamese Region Proposal Networks (DaSiamRPN) framework to learn distractor-aware features in the off-line training,  and explicitly suppress distractors during the inference of online tracking.

3, We extend the DaSiamRPN to perform long-term tracking by introducing a simple yet effective local-to-global search region strategy, which significantly improves the performance of our tracker in out-of-view and full occlusion challenges. In comprehensive experiments of short-term and long-term visual tracking benchmarks, the proposed DaSiamRPN framework obtains state-of-the-art accuracy while performing at far beyond real-time speed.

\section{Related Work}

%Visual tracking is a significant problem in computer vision systems and a series of approaches have been proposed in recent years. Since our main contribution is a distractor-aware Siamese framework for accurate and long-term tracking, we give a brief review on three directions closely related to this work.

\subsubsection{Siamese Networks based Tracking.}
Siamese trackers follow a tracking by similarity comparison strategy.
The pioneering work is SINT~\cite{SINT}, which simply searches for the candidate most similar to the exemplar given in the starting frame, using a run-time fixed but learns a priori deep Siamese similarity function.
As a follow-up work, Bertinetto et.al \cite{SiamFC} propose a fully convolutional Siamese network (SiamFC) to estimate the feature similarity region-wise between two frames. RASNet \cite{RASNet} advances this similarity metric by learning the attention mechanism with a Residual Attentional Network.
Different from SiamFC and RASNet, in GOTURN tracker \cite{GOTURN}, the motion between successive frames is predicted using a deep regression network. These threee trackers are able to perform at 86 FPS, 83FPS and 100 FPS respectively on GPU because no fine-tuning is performed online.
CFNet \cite{CFNet} interprets the correlation filters as a differentiable layer in a Siamese tracking framework, thus achieving an end-to-end representation learning. But the performance improvement is limited compared with SiamFC.
%DSiam~\cite{DSiam} proposes a dynamic Siamese network by performing appearance variation learning and background suppression at the same time.
FlowTrack~\cite{FlowTrack} exploits motion information in Siamese architecture to improve the feature representation and the tracking accuracy.
It is worth noting that CFNet and FlowTrack can efficiently online update the tracking model.
Recently, SiamRPN~\cite{SiamRPN} formulates the tracking as a one-shot local detection task by introducing a region proposal network after a Siamese network, which is end-to-end trained off-line with large-scale image pairs.

\subsubsection{Features for Tracking.}
Visual features play a significant role in computer vision tasks including visual tracking.
Possegger et.al~\cite{DAT} propose a distractor-aware model term to suppress visually distracting regions, while the color histograms features used in their framework are less robust than the deep features.
DLT~\cite{DLT} is the seminal deep learning tracker which uses a multi-layer autoencoder network. The feature is pretrained on part of the 80M Tiny Image dataset~\cite{80MTinyImage} in an unsupervised fashion. Wang et al.~\cite{wang2015video} learn a two-layer neural network on a video repository, where temporally slowness constraints are imposed for feature learning.
DeepTrack~\cite{DeepTrack} learns two-layer CNN classifiers from binary samples and does not require a pre-training procedure. UCT~\cite{UCT} formulates the features learning and tracking process into a unified framework, enabling learned features are tightly coupled to tracking process.

%Recently, considerable attention has been paid to Correlation Filters (CF) based trackers due to their high computational efficiency with the use of fast Fourier transforms\cite{KCF,Staple,DeepSRDCF,HCF,HDT,gladh2016deep,UCT}. Hand-crafted features used in CF trackers are mainly HOG and color-attributes~\cite{KCF,Danelljan_2014_CVPR,Staple}. Inspired by the success of CNN in object classification, detection and segmentation, the visual tracking community has started to focus on the deep trackers that exploit the strength of CNN features in CF tracking approaches. In HCF \cite{HCF}, HDT \cite{HDT} and DeepSRDCF \cite{DeepSRDCF}, CNN are employed to extract features instead of handcrafted features. Gladh et.al~\cite{gladh2016deep} use CNN to extract deep motion features from optical flow feature maps.
%It is worth noting that the chosen CNN features are always pre-trained in different tasks and individual components in tracking systems are learned separately. So the achieved tracking results may be suboptimal.

\subsubsection{Long-term Tracking.}
Traditional long-term tracking frameworks can be divided into two groups: earlier methods regard tracking as local key point descriptors matching with a geometrical model~\cite{pernici2014object,nebehay2015clustering,maresca2013matrioska}, and recent approaches perform long-term tracking by combining a short-term tracker with a detector.
The seminal work of latter categories is TLD~\cite{TLD}, which proposes a memory-less flock of flows as a short-term tracker and a template-based detector run in parallel. Ma et al. ~\cite{LCT}propose a combination of KCF tracker and a random ferns classifier as a detector that is used to correct the tracker. Similarly, MUSTer~\cite{muster} is a long-term tracking framework that combines KCF tracker with a SIFT-based detector that is also used to detect occlusions. Fan and Ling~\cite{PTAV} combines a DSST tracker~\cite{DSST} with a CNN detector~\cite{SINT} that verifies and potentially corrects proposals of the short-term tracker.
%For the methods ,,in former category, Pernici et al.~\cite{pernici2014object} propose to reduce contamination of the keypoints model that occurs at adaptation during occlusion. Nebehay et al.~\cite{nebehay2015clustering} utilize a keypoint tracker without updating and using pairs of correspondences in a Generalized Hough Transform (GHT) framework to track deformable models. Maresca and Petrosino~\cite{maresca2013matrioska} extend the GHT approach by integrating various descriptors and introducing a conservative updating mechanism. These keypoint based methods require a large and well textured target, which constrains their application scenarios.

\section{Distractor-aware Siamese Networks}

%In this section, we first provide an in-depth analysis of features in Siamese trackers. Then we introduce an effective sampling strategy to balance the distributions of training data. In order to make full use of the information in the context, we propose a distractor-aware module that can explicitly suppress the distractors. Finally, we extend the DaSiamRPN framework with iterative growing search region to adapt for long-term tracking.

\begin{figure}[t]
\centering
\subfloat[ROI]{%
  \begin{tabular}{c}
   \includegraphics[width=0.18\linewidth]{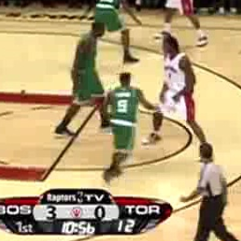}
  \end{tabular}
 }%
 \subfloat[SiamFC]{%
  \begin{tabular}{c}
   \includegraphics[width=0.18\linewidth]{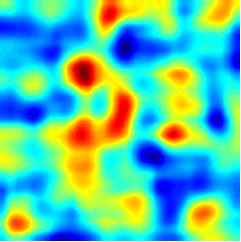}
  \end{tabular}
 }%
\subfloat[SiamRPN]{%
  \begin{tabular}{c}
   \includegraphics[width=0.18\linewidth]{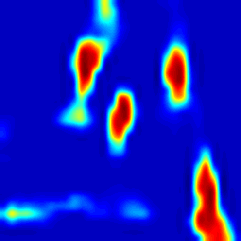}
  \end{tabular}
 }%
 \subfloat[SiamRPN+]{%
  \begin{tabular}{c}
  \includegraphics[width=0.18\linewidth]{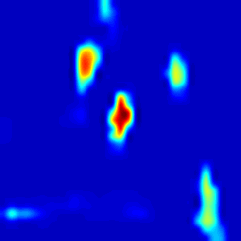}
  \end{tabular}
 }%
  \subfloat[Ours]{%
  \begin{tabular}{c}
  \includegraphics[width=0.18\linewidth]{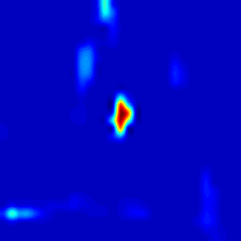}\label{fig:balldat}
  \end{tabular}
 }%
\caption{Visualization of the response heatmaps of Siamese network trackers. (a) shows the search images. (b-e) show the heatmaps that produced by SiamFC, SiamRPN, SiamRPN+ (trained with distractors) and the DaSiamRPN.
}
\label{fig:visual}
\end{figure}

\subsection{Features and Drawbacks in Traditional Siamese Networks}

\label{sect:figures}
Before the detailed discussion of our proposed framework, we first revisit the features of conventional Siamese network based tracking~\cite{SiamFC,SiamRPN}. Siamese trackers use metric learning at their core. The goal is to learn an embedding space that can maximize the interclass inertia between different objects and minimize the intraclass inertia for the same object. The key contribution leading to the popularity and success of Siamese trackers is their balanced accuracy and speed.

%SiamFC~\cite{SiamFC} learns discriminating embedding with pairs of template, search region and corresponding prediction map. SiamRPN~\cite{SiamRPN} significantly boosts the tracking accuracy by predicting both confidences and offsets for multi anchor boxes. The feature representation networks in these methods are frozen during the online tracking. Without online fine-tune, these methods heavily depend on the off-line training process.

Fig.~\ref{fig:visual} visualizes of response maps of SiamFC and SiamRPN. It can be seen that for the targets, those with large differences in the background also achieve high scores, and even some extraneous objects get high scores. The representations obtained in SiamFC usually serve the discriminative learning of the categories in training data.
In SiamFC and SiamRPN, pairs of training data come from different frames of the same video, and for each search area, the \emph{non-semantic} background occupies the majority, while semantic entities and distractor occupy less. This imbalanced distribution makes the training model hard to learn instance-level representation, but tending to learn the differences between foreground and background.

During inference, nearest neighbor is used to search the most similar object in the search region, while the background information labelled in the first frame are omitted. The background information in the tracking sequences can be effectively utilized to increase the discriminative capability as shown in Fig.~\ref{fig:balldat}.

To eliminate these issues, we propose to actively generate more semantics pairs in the offline training process and explicitly suppress the distractors in the online tracking.

\subsection{Distractor-aware Training}

\label{sect:train}
High quality training data is crucial for the success of end-to-end representation learning in visual tracking. We introduce series of strategies to improve the generalization of the learned features and eliminate the imbalanced distribution of the training data.

\begin{figure}[t]
\setlength{\abovecaptionskip}{-0.2cm}
\setlength{\belowcaptionskip}{-0.5cm}
\begin{center}
\includegraphics[width=0.9\linewidth]{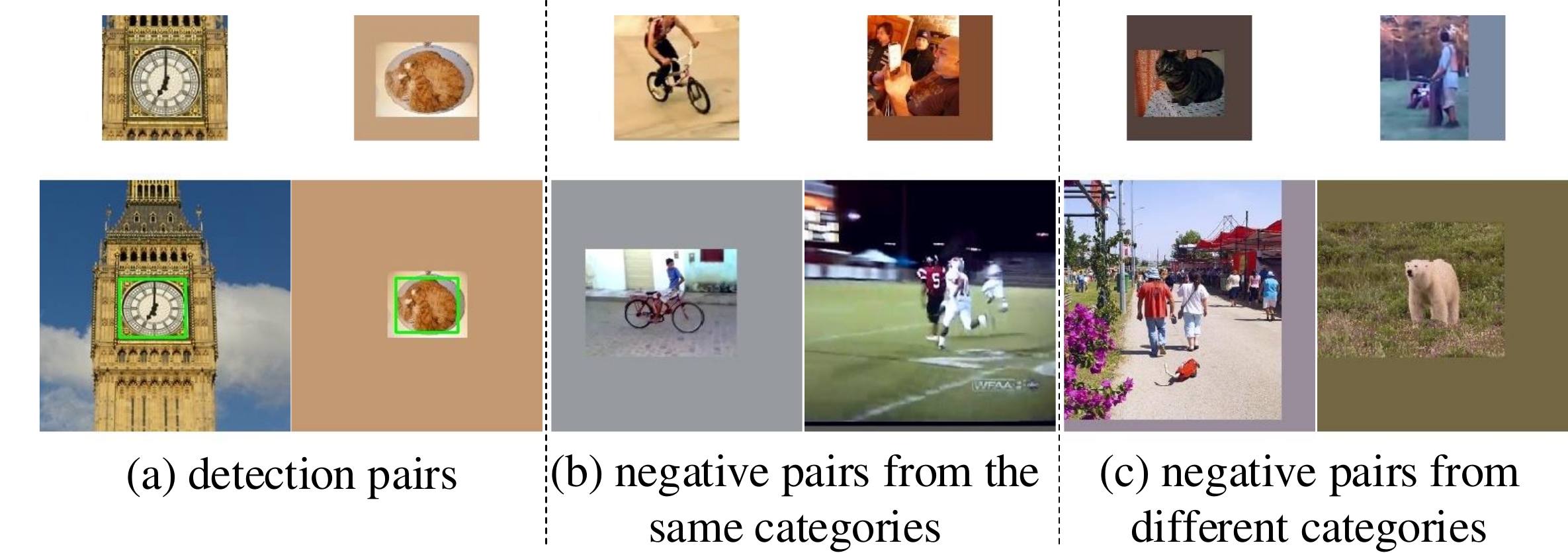}
\end{center}
\caption{(a) positive pairs generated from detection datasets through augmenting still images. (b) negative pairs from the same category. (c) negative pairs from different categories.}
\label{fig:data}
\end{figure}

\subsubsection{Diverse categories of positive pairs can promote the generalization ability.}
The original SiamFC is trained on the ILSVRC video detection datasets, which consists of only about 4,000 videos annotated frame-by-frame \cite{ILSVRC15}. Recently, SiamRPN \cite{SiamRPN} explores to use sparsely labelled Youtube-BB \cite{YouTubeBB} videos which consists of more than 200,000 videos annotated once in every 30 frames. In these two methods, target pairs of training data come from different frames in the same video. However, these video detection datasets only contain few categories (20 for VID \cite{ILSVRC15}, 30 for Youtube-BB \cite{YouTubeBB}), which is not sufficient to train high-quality and generalized features for Siamese tracking.
Besides, the bounding box regression branch in the SiamRPN may get inferior predictions when encountering new categories.
Since labelling videos is time-consuming and expensive, in this paper, we greatly expand the categories of positive pairs by introducing large-scale ImageNet Detection \cite{ILSVRC15} and COCO Detection \cite{COCO} datasets. As shown in Fig.~\ref{fig:data}(a), through augmentation techniques (translation, resize, grayscale et.al), still images from detection datasets can be used to generate image pairs for training.
The diversity of positive pairs is able to improve the tracker's discriminative ability and regression accuracy.

\subsubsection{Semantic negative pairs can improve the discriminative ability.}
We attribute the less discriminative representation in SiamFC~\cite{SiamFC} and SiamRPN~\cite{SiamRPN} to two level of imbalanced training data distribution. The first imbalance is the rare semantic negative pairs. Since the background occupies the majority in the training data of SiamFC and SiamRPN, most negative samples are non-semantic (not real object, just background), and they can be easily classified.
That is to say, SiamFC and SiamRPN learn the differences between foreground and background, and the losses between semantic objects are overwhelmed by the vast number of easy negatives. Another imbalance comes from the intraclass distractors, which usually perform as hard negative samples in the tracking process. In this paper, semantic negative pairs are added into the training process. The constructed negative pairs consist of labelled targets both in the same categories and different categories. The negative pairs from different categories can help tracker to avoid drifting to arbitrary objects in challenges such as out-of-view and full occlusion, while negative pairs from the same categories make the tracker focused on fine-grained representation. The negative examples are shown in Fig.~\ref{fig:data}(b) and Fig.~\ref{fig:data}(c).

\subsubsection{Customizing effective data augmentation for visual tracking.}
To unleash the full potential of the Siamese network, we customize several data augmentation strategies for training. Except the common translation, scale variations and illumination changes, we observe that the motion pattern can be easily modeled by the shallow layers in the network. We explicitly introduce motion blur in the data augmentation.

\subsection{Distractor-aware Incremental Learning}

\label{sect:DAT}

\begin{figure}[t]

\subfloat[General Siamese tracker]{\includegraphics[width=0.49\linewidth]{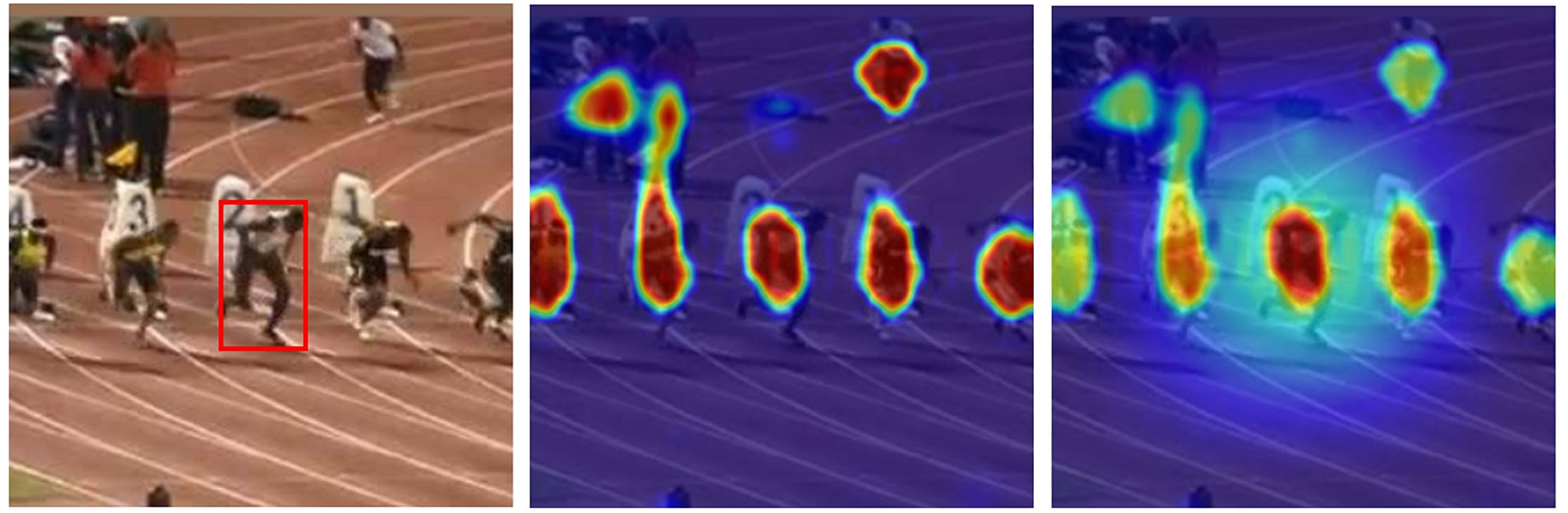}\label{fig:boltfc}}
\hfil
\subfloat[Distractor-aware Siamese tracker]{\includegraphics[width=0.49\linewidth]{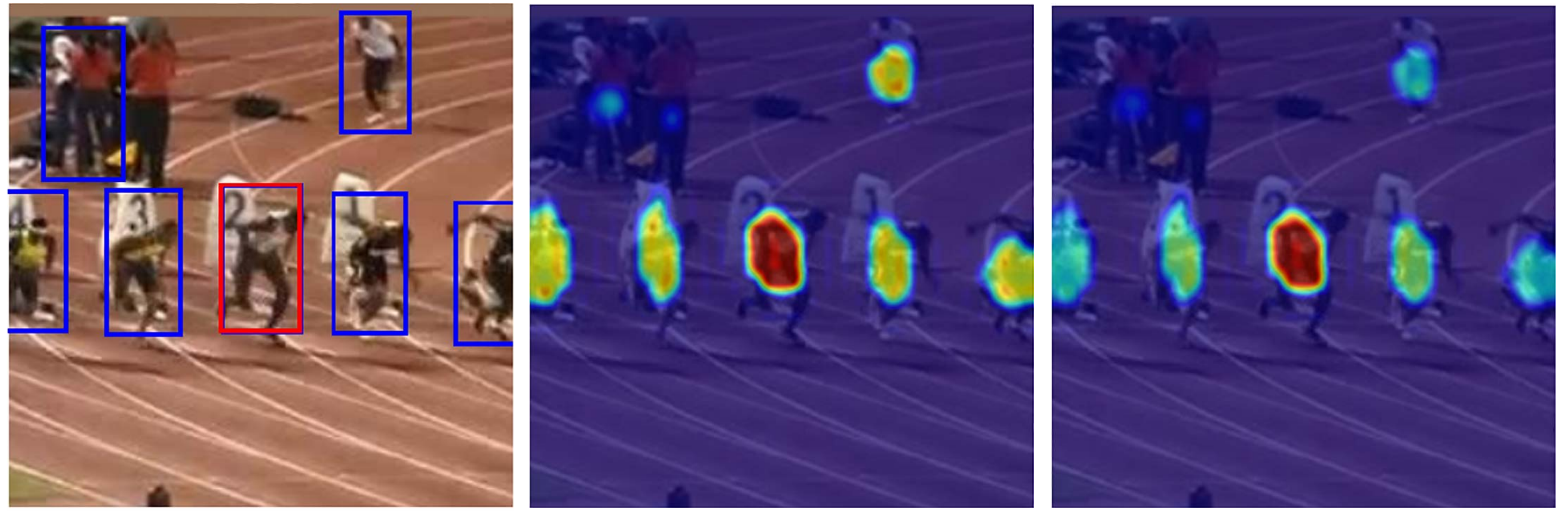}\label{fig:boltdat}}
\caption{Illustrations of our proposed Distractor-aware Siamese Region Proposal Networks (DaSiamRPN). The target and the background information are fully utilized in DaSiamRPN, which can suppress the influence of distractor during tracking.}
\label{fig:dat}
\end{figure}

The training strategy in the last subsection can significantly improve the discrimination power on the offline training process.
However, it is still hard to distinguish two objects with the similar attributes like Fig.~\ref{fig:boltfc}. SiamFC and SiamRPN use a cosine window to suppress the distractors. In this way, the performance is not guaranteed when the motion of objects are messy. Most existing Siamese network based approaches provide inferior performance when encountering with fast motion or background clutter. In summary, the potential flaw is mainly due to the misalignment of the general representation domain and the specifical target domains. In this section, we propose a distractor-aware module to effectively transfer the general representation to the video domain.

The Siamese tracker learns a similarity metric $f(z,x)$ to compare an exemplar image $z$ to a candidate image $x$ in the embedding space $\varphi$:
\begin{equation}
f(z,x)=\varphi(z)\star \varphi(x) + b \cdot \mathbbm{1}
\label{eq:siamfc}
\end{equation}
where $\star$ denotes cross correlation between two feature maps, $b \cdot \mathbbm{1}$ denotes a bias which is equated in every location. The most similar object of the exemplar will be selected as the target.

To make full use of the label information, we integrate the hard negative samples (distractors) in context of the target into the similarity metric.
In DaSiamRPN, the Non Maximum Suppression (NMS) is adopted to select the potential distractors $d_i$ in each frames, and then we collect a distractor set $\mathcal{D}:=\{\forall d_i\in \mathcal{D}, f(z,d_i) > h \cap d_i \neq z_t\}$, where $h$ is the predefined threshold, $z_t$ is the selected target in frame $t$ and the number of this set $|\mathcal{D}|=n$. Specifically, we get $17*17*5$ proposals in each frame at first, and then we use NMS to reduce redundant candidates. The proposal with highest score will be selected as the target $z_t$. For the remaining, the proposals with scores greater than a threshold are selected as distractors.

After that, we introduce a novel distractor-aware objective function to \emph{re-rank} the proposals $\mathcal{P}$ which have \emph{top-k} similarities with the exemplar. The final selected object is denoted as $q$:
\begin{equation}
q = \underset{p_k \in \mathcal{P}}{argmax} ~~f(z,p_k)- \frac{\hat{\alpha}\sum_{i=1}^{n} \alpha_i f(d_i,p_k)}{\sum_{i=1}^{n}\alpha_i }
\label{eq:datobj}
\end{equation}
the weight factor $\hat{\alpha}$ control the influence of the distractor learning, the weight factor $\alpha_i$ is used to control the influence for each distractor $d_i$. It is worth noting that the computational complexity and memory usage increase $n$ times by a direct calculation.
Since cross correlation operation in the Equation~(\ref{eq:siamfc}) is a linear operator, we utilize this property to speed up the distractor-aware objective:
\begin{equation}
q = \underset{p_k \in \mathcal{P}}{argmax} ~~ (\varphi(z) - \frac{\hat{\alpha}\sum_{i=1}^{n} \alpha_i \varphi(d_i)}{\sum_{i=1}^{n}\alpha_i} )\star \varphi(p_k)
\end{equation}
it enables the tracker run in the comparable speed in comparisons with SiamRPN. This associative law also inspires us to incrementally learn the target templates and distractor templates with a learning rate $\beta_t$:
\begin{equation}
q_{T+1} = \underset{p_k \in \mathcal{P}}{argmax} ~~ (\frac{\sum_{t=1}^{T} \beta_t \varphi(z_t)}{\sum_{t=1}^{T} \beta_t} - \frac{\sum_{t=1}^{T}\beta_t\hat{\alpha}\sum_{i=1}^{n} \alpha_i \varphi(d_{i,t})}{\sum_{t=1}^{T} \beta_t\sum_{i=1}^{n}\alpha_i} )\star \varphi(p_k)
\label{eq:dat_update}
\end{equation}
This distractor-aware tracker can adapt the existing similarity metric (general) to a similarity metric for a new domain (specific).
The weight factor $\alpha_i$ can be viewed as the dual variables with sparse regularization, and the exemplars and distractors can be viewed as positive and negative samples in correlation filters. Actually, an online classifier is modeled in our framework. So the adopted classifier is expected to perform better than these only use general similarity metric.

\subsection{DaSiamRPN for Long-term Tracking}

% \begin{figure}[t]
%   \centering
%   \includegraphics[width=0.9\linewidth]{person7.pdf}
% \caption{The tracking snapshots of video \emph{person7} in out-of-view challenge.
% First rows: results of SiamRPN. Second rows: detection scores and according overlaps of the proposed DaSiamRPN and SiamRPN~\cite{SiamRPN}. The overlaps are difined as intersection-over-union (IOU) between tracking results and ground truth. The arrows indicate tracking results and according overlaps in line charts. Third rows: results of DaSiamRPN.
% \color{red} Red: ground truth. \color{green} Green: tracking box.  \color{blue} Blue: Search region box.}
%   \label{out-of-view}

% \end{figure}

\begin{figure}[t]
\centering
\subfloat[scores and overlaps in SiamRPN]{\includegraphics[width=0.48\linewidth,trim=3cm 1cm 1cm 4cm]{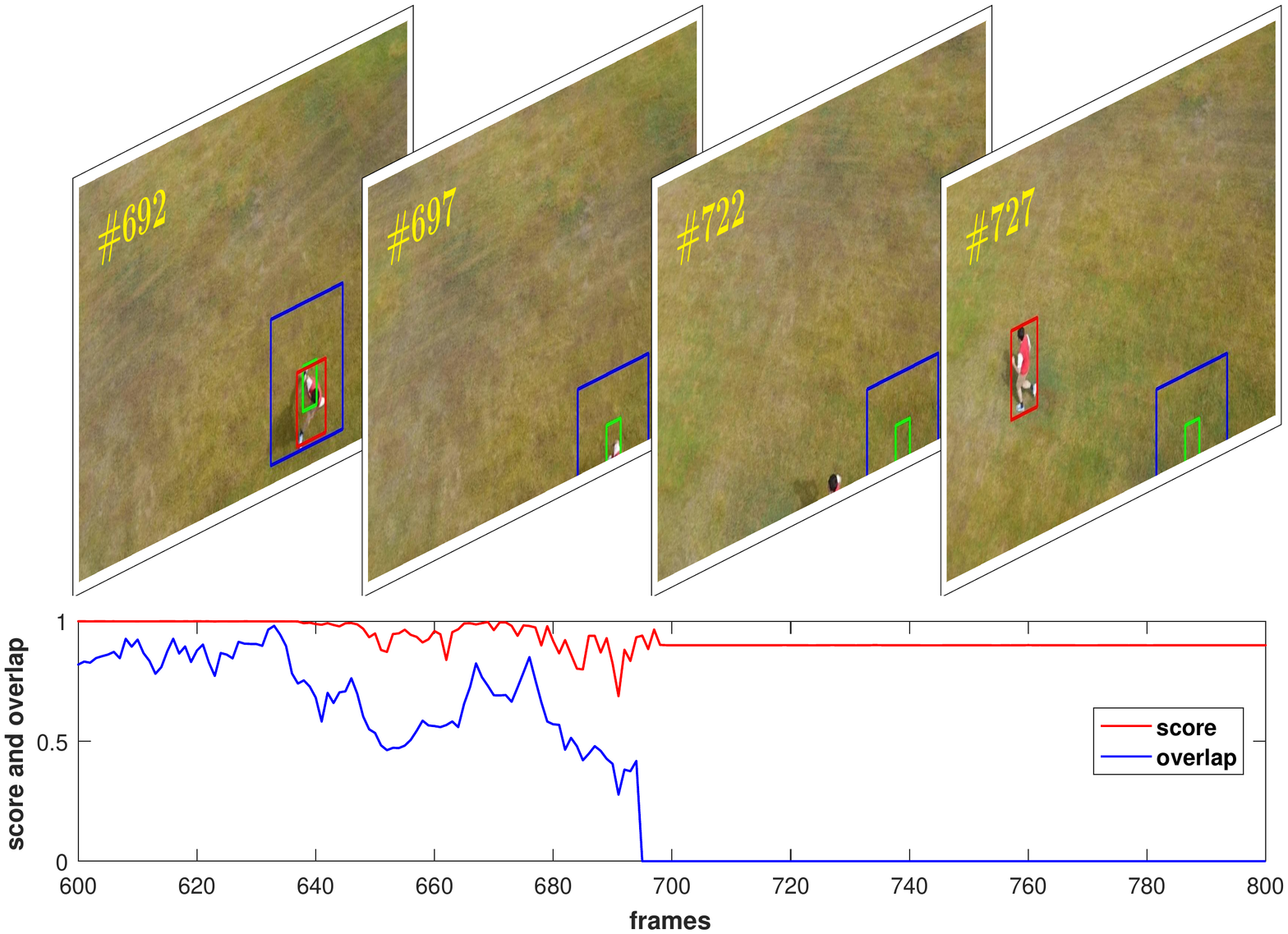}}
\subfloat[scores and overlaps in DaSiamRPN]{\includegraphics[width=0.48\linewidth, trim=3cm 1cm 1cm 4cm]{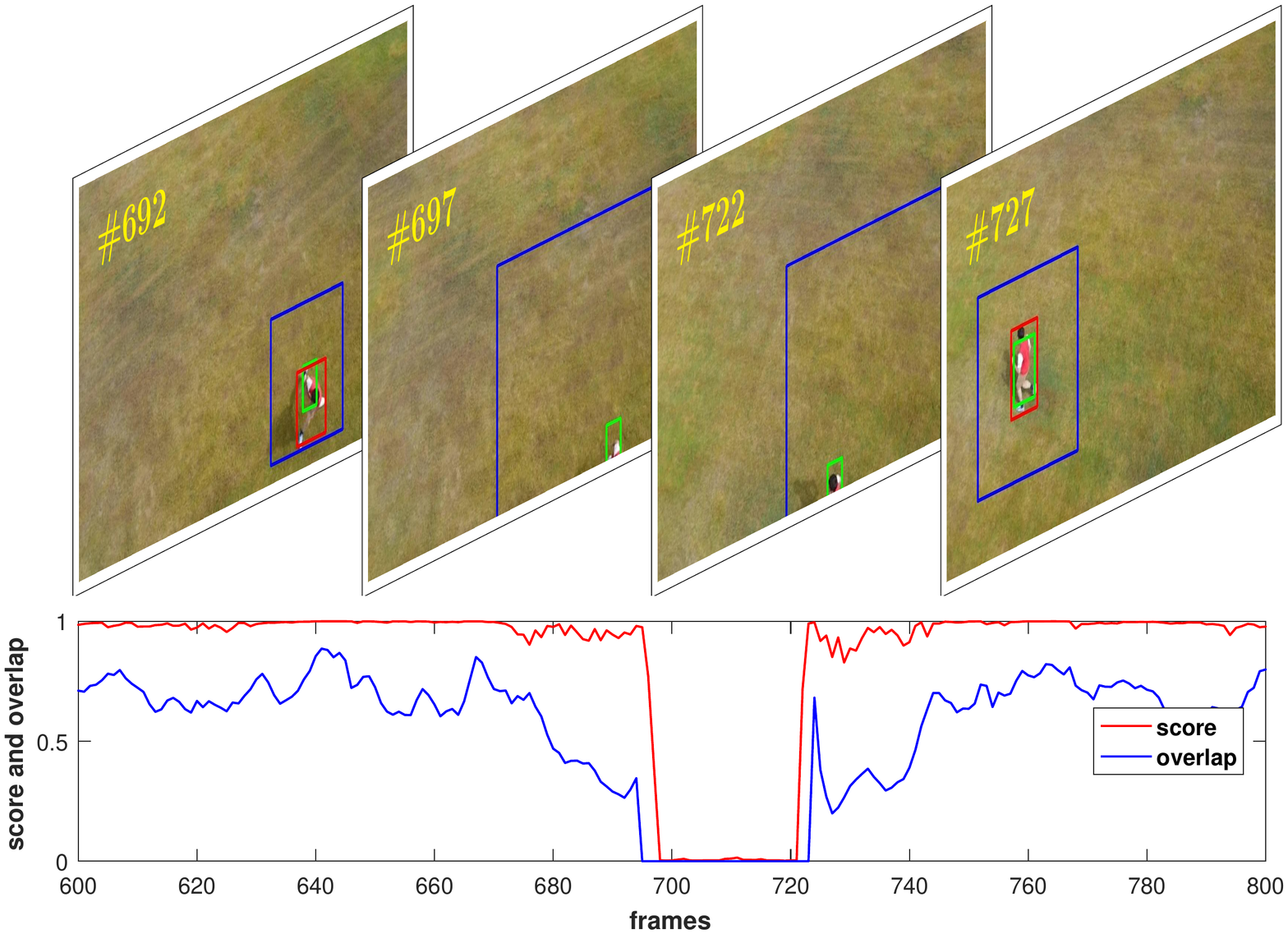}}
\caption{The tracking results of video \emph{person7} in out-of-view challenge.
 First row: tracking snapshots of SiamRPN and DaSiamRPN. Second row: detection scores and according overlaps of the two methods. The overlaps are defined as intersection-over-union (IOU) between tracking results and ground truth. \color{red} Red: ground truth. \color{green} Green: tracking box.  \color{blue} Blue: Search region box.}
 \label{out-of-view}

\end{figure}

In this section, the DaSiamRPN framework is extended for long-term tracking. Besides the challenging situations in short-term tracking, severe out-of-view and full occlusion introduce extra challenges in long-term tracking, which are shown in Fig.~\ref{out-of-view}. The search region in short-term tracking (SiamRPN) can not cover the target when it reappears, thus failing to track the following frames.
We propose a simple yet effective switch method between short-term tracking phase and failure cases. In failure cases, an iterative local-to-global search strategy is designed to re-detect the target.

In order to perform switches, we need to identify the beginning and the end of failed tracking. Since the distractor-aware training and inference enable high-quality detection score, it can be adopted to indicate the quality of tracking results. Fig.~\ref{out-of-view} shows the detection scores and according tracking overlaps in SiamRPN and DaSiamRPN. The detection scores of SiamRPN are not indicative, which can be still high even in out-of-view and full occlusion. That is to say, SiamRPN tends to find an arbitrary objectness in these challenges which causes drift in tracking. In DaSiamRPN, detection scores successfully indicate status of the tracking phase.

During failure cases, we gradually increase the search region by local-to-global strategy. Specifically, the size of search region is iteratively growing with a constant step when failed tracking is indicated. As shown in Fig.~\ref{out-of-view}, the local-to-global search region covers the target to recover the normal tracking. It is worth noting that our tracker employs bounding box regression to detect the target, so the time-consuming image pyramids strategy can be discarded. In experiments, the proposed DaSiamRPN can perform at 110 FPS on long-term tracking benchmark.

%\begin{figure}[t]
 % \centering
 % \includegraphics[width=0.9\linewidth]{person14.pdf}
 % \caption{The tracking snapshots of video \emph{person14} in full occlusion challenge. First and second row indicate results of SiamRPN and DaSiamRPN, respectively. \color{red} Red: ground truth. \color{green} Green: tracking box.  \color{blue} Blue: Search region box.}
  %\label{full occlusion}

%\end{figure}

\section{Experiments}

Experiments are performed on extensive challenging tracking datasets,
including VOT2015~\cite{VOT2015}, VOT2016~\cite{VOT2016} and VOT2017~\cite{VOT2017}, each with 60 videos, UAV20L~\cite{UAV} with 20 long-term videos, UAV123~\cite{UAV} with 123 videos and OTB2015~\cite{OTB2015} with 100 videos. All the tracking results are provided by official implementations to ensure a fair comparison.

\subsection{Experimental Details}

The modified AlexNet~\cite{AlexNet} pretrained using ImageNet \cite{ILSVRC15} is used as described in SiamRPN~\cite{SiamRPN}.
The parameters of the first three convolution layers are fixed and only the last two convolution
layers are fine-tuned. There are totally 50 epoches performed and the learning rate is decreased in log space from $10^{-2}$ to $10^{-4}$.
We extract image pairs from VID~\cite{ILSVRC15} and Youtube-BB \cite{YouTubeBB} by choosing frames with interval less than 100 and performing crop procedure as described in Section~\ref{sect:train}. In ImageNet Detection \cite{ILSVRC15} and COCO Detection \cite{COCO} datasets, image pairs are generated for training by augmenting still images.
To handle the gray videos in benchmarks, 25\% of the pairs are converted to grayscale during training. The translation is randomly performed within 12 pixels, and the range of random resize varies from 0.85 to 1.15. %There are totally 500,000

During inference phase, the distractor factor $\hat{\alpha}$ in Equation~(\ref{eq:datobj}) is set to 0.5, $\alpha_i$ is set to 1 for each distractor, and the incremental learning factor $\beta_t$ in Equation~(\ref{eq:dat_update}) is set to $\sum_{i=0}^{t-1}(\frac{\eta}{1-\eta})^i$, where $\eta=0.01$. In the long-term tracking, we find that one step iteration of local-to-global is sufficient. Specifically, the sizes of the search region in short-term phase and defined failure cases are set to 255 and 767, respectively. The thresholds to enter and leave failure cases are set to 0.8 and 0.95.
Our experiments are implemented using PyTorch on a PC with an Intel i7, 48G RAM, NVIDIA TITAN X. The proposed tracker can perform at 160 FPS on short-term benchmarks and 110 FPS on long-term benchmarks. The code and experimental results are available at \url{https://github.com/foolwood/DaSiamRPN}.

\begin{figure}[t]
\centering
\subfloat[EAO on VOT2016]{\includegraphics[width=0.48\linewidth]{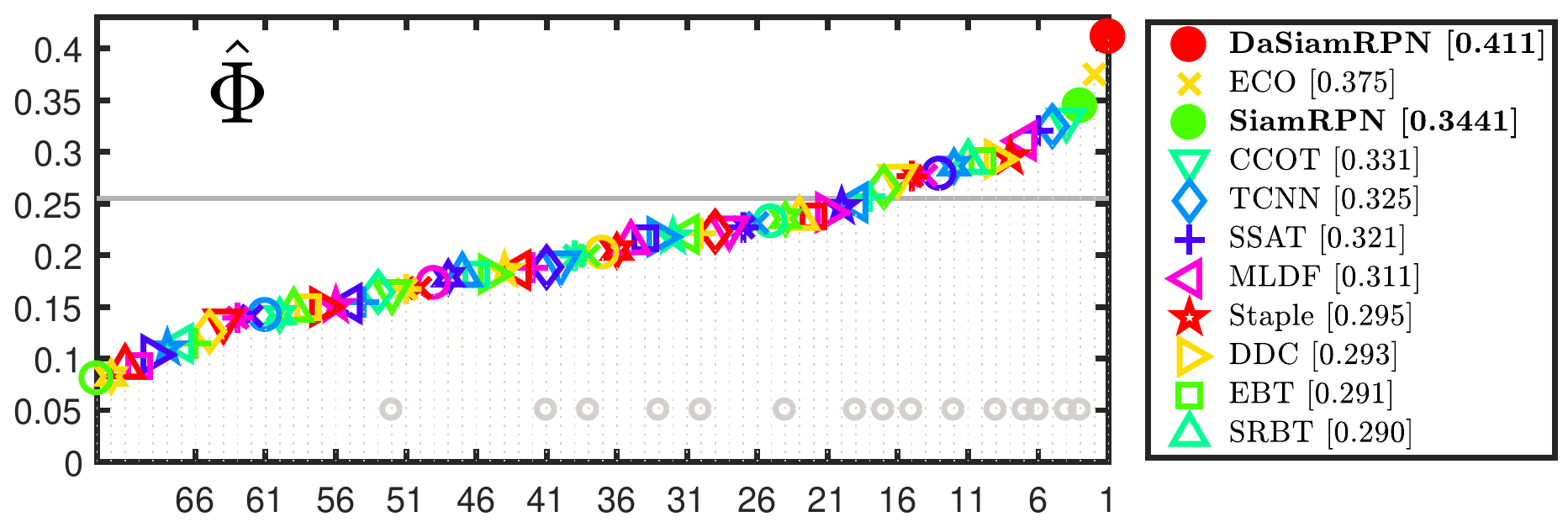}%
\label{fig:vot16}}
\subfloat[EAO on VOT2017]{\includegraphics[width=0.48\linewidth]{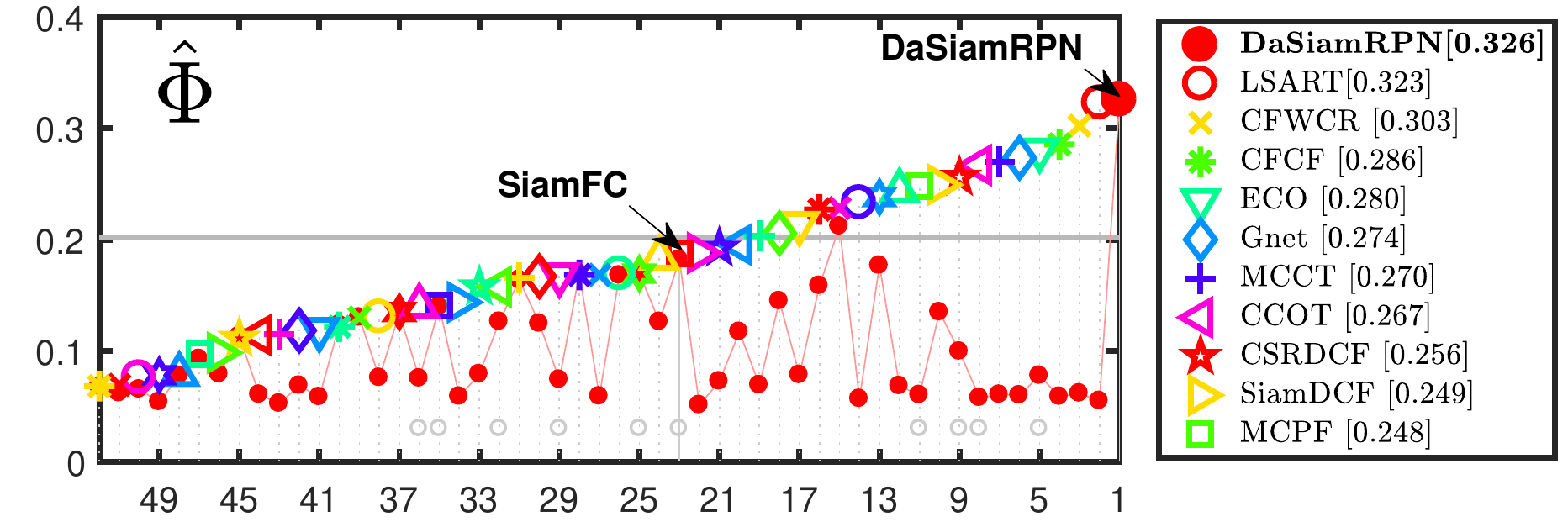}%
\label{fig:vot17}}
\caption{Expected average overlap plot for VOT2016 (a) and VOT2017 (b).}
\label{fig_vot}

\end{figure}

\subsection{State-of-the-art Comparisons on VOT Datasets}

\label{sect:vot}

In this section the latest version of the Visual Object Tracking toolkit (\emph{vot2017-challenge}) is used. The toolkit applies a reset-based methodology. Whenever a failure (zero overlap with the ground truth) is detected, the tracker is re-initialized five frames after the failure. The performance is measured in terms of accuracy (A), robustness (R), and expected average overlap (EAO). In addition, VOT2017 also introduces a real-time experiment. We report all these metrics compared with a number of the latest state-of-the-art trackers on VOT2015, VOT2016 and VOT2017.

The EAO curve evaluated on VOT2016 is presented in Fig.~\ref{fig:vot16} and 70 other state-of-the-art trackers are compared. The EAO of our baseline tracker SiamRPN on VOT2016 is 0.3441, which already outperforms most of state-of-the-arts. However, there is still a gap compared with the top-ranked tracker ECO (0.375), which improves continuous convolution operators on multi-level feature maps. Most remarkably, the proposed DaSiamRPN obtains a EAO of 0.411, outperforming state-of-the-arts by relative 9.6\%. Furthermore, our tracker runs at state-of-the-art speed with 160FPS, which is $500\times$ faster than C-COT and $20\times$ faster than ECO.

For the evaluation on VOT2017, Fig.~\ref{fig:vot17} reports the results of ours against 51 other state-of-the-art trackers with respect to the EAO score. DaSiamRPN ranks first with an EAO score of 0.326. Among the top 5 trackers, CFWCR, CFCF, ECO, and Gnet apply continuous convolution operator as the baseline approach. The top performer LSART~\cite{LSART} decomposes the target into patches and applies a weighted combination of patch-wise similarities into a kernelized ridge regression. While our method is conceptually much simpler, powerful and is also easy to follow.

Fig.~\ref{fig:vot17} also reveals the EAO values in the real-time experiment denoted by red points. Our tracker obviously is the top-performer with a real-time EAO of 0.326 and outperforms the latest state-of-the-art real-time tracker CSRDCF++ by relative 53.8\%.

Table~\ref{tab:component} shows accuracy (A) and robustness (R), as well as expected average overlap (EAO) on VOT2015, VOT2016 and VOT2017.
The baseline approach SiamRPN can process an astounding 200 frames per second while still getting an comparable performance with the state-of-the-arts.
We find the performance gains of SiamRPN are mainly due to their accurate multi-anchors regression mechanism. We propose the distractor-aware module to improve the robustness, which can make our tracker much more harmonious.
As a result, our approach, with the EAO of 0.446, 0.411 and 0.326 on three benchmarks, outperforms all the existing trackers by a large margin.
We believe that the consistent improvements demonstrate that our approach makes real contributions by both the training process and online inference.

\begin{table}[t]

\centering
\caption{Performance comparisons on public short-term benchmarks. OP: mean overlap precision at the threshold of 0.5; DP: mean distance precision of 20 pixels; EAO: expected average overlap, and mean speed (FPS). The {\color{red}\textbf{red bold}} fonts and {\color{blue}\textit{blue italic}} fonts indicate the best and the second best performance.}
\label{tab:component}
\begin{tabular}{|c|cc|ccc|ccc|ccc|c|}
\hline
\multirow{2}{*}{Trackers} & \multicolumn{2}{c|}{OTB-2015} & \multicolumn{3}{c|}{VOT2015} & \multicolumn{3}{c|}{VOT2016} & \multicolumn{3}{c|}{VOT2017} & \multicolumn{1}{c|}{\multirow{2}{*}{FPS}} \TBstrut\\
\cline{2-12} & OP & DP & A & R & EAO & A & R & EAO & A & R & EAO&  \Tstrut\\ \hline\hline
SiamFC  & 73.0 & 77.0 & 0.533 & 0.88 & 0.289 & 0.53 & 0.46 & 0.235 & 0.50 & 0.59 & 0.188 & 86\Tstrut\\
CFNet   & 69.9 & 74.7 & - & - & - & - & - & - & - & - & - & 75\Tstrut\\
Staple  & 70.9 & 78.4 & 0.57 & 1.39 & 0.300 & 0.54 & 0.38 & 0.295 & \color{blue}\textit{0.52} & 0.69 & 0.169 & 80\Tstrut\\
CSRDCF  & 70.7 & 78.7 & 0.56 & 0.86 & 0.320 & 0.51 & 0.24 & 0.338 & 0.49 & 0.36 & 0.256 & 13\Tstrut\\
%PTAV   & 81.2 & 87.9 & 76.8 & 84.1 & - & - & - & - & - & - & - & - & - & 15\Tstrut\\
BACF    & 76.7 & 81.5 & 0.59 & 1.56 & - & - & - & - & - & - & - & 35\Tstrut\\
ECO-HC  & 78.4 & 85.6 & - & - & - & 0.54 & 0.30 & 0.322 & 0.49 & 0.44 & 0.238 & 60\Tstrut\\
\hline\hline
%SINT  & 81.6 & 88.1 & 71.9 & 78.8 & - & - & - & - & - & - & - & - & - & 5\Tstrut\\
CREST  & 77.5 & 83.7 & - & - & - & 0.51 & 0.25 & 0.283 & - & - & - & 1\Tstrut\\
MDNet  & \color{blue}\textit{85.4} & \color{blue}\textit{90.9} & \color{blue}\textit{0.60} & \color{blue}\textit{0.69} & \color{blue}\textit{0.378} & 0.54 & 0.34 & 0.257 & - & - & - & 1\Tstrut\\
C-COT  & 82.0 & 89.8 & 0.54 & 0.82 & 0.303 & 0.54 & 0.24 & 0.331 & 0.49 & \color{blue}\textit{0.32} & 0.267 & 0.3\Tstrut\\
ECO    & 84.9 & \color{red}\textbf{91.0} & - & - & - & 0.55 & \color{red}\textbf{0.20} & \color{blue}\textit{0.375} & 0.48 & \color{red}\textbf{0.27} & \color{blue}\textit{0.280} & 8\Tstrut\\
\hline\hline
SiamRPN    & 81.9 & 85.0 & 0.58 & 1.13 & 0.349 & \color{blue}\textit{0.56} & 0.26 & 0.344 & 0.49 & 0.46 & 0.244 & {\color{red}\textbf{200}}\Tstrut\\
\textbf{Ours} & {\color{red}\textbf{86.5}} & 88.0 & \color{red}\textbf{0.63} & \color{red}\textbf{0.66} & \color{red}\textbf{0.446} & \color{red}\textbf{0.61} & \color{blue}\textit{0.22} & \color{red}\textbf{0.411} & \color{red}\textbf{0.56} & 0.34 & \color{red}\textbf{0.326} & {\color{blue}\textit{160}}\Tstrut\\
\hline
\end{tabular}

\end{table}

\subsection{State-of-the-art Comparisons on UAV Datasets}

The UAV~\cite{UAV} videos are captured from low-altitude unmanned aerial vehicles. The dataset contains a long-term evaluation subset UAV20L and a short-term evaluation subset UAV123. The evaluation is based on two metrics: precision plot and success plot.
%The precision plot shows the percentage of frames that the tracking results are within certain distance determined by given threshold to the ground truth. The value when threshold is 20 pixels is always taken as the representative precision score. The success plot shows the ratios of successful frames when the threshold varies from 0 to 1, where a successful frame means its overlap is larger than this given threshold. The area under curve (AUC) of each success plot is used to rank the tracking algorithm.

\begin{figure*}[!tp]
 \centering
\begin{minipage}[c]{3cm}
\includegraphics[width=3cm]{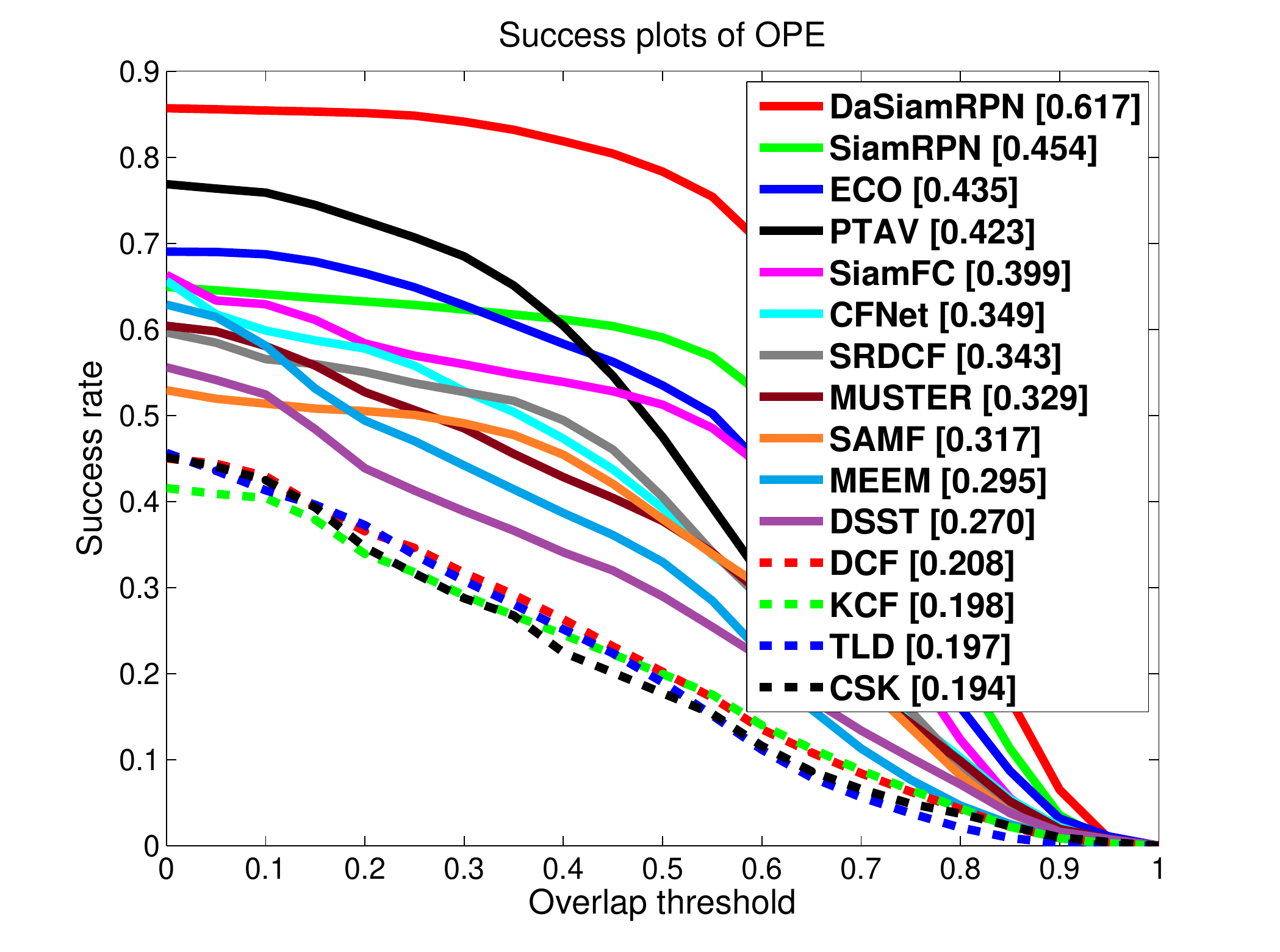}
\end{minipage}%
\begin{minipage}[c]{3cm}
\includegraphics[width=3cm]{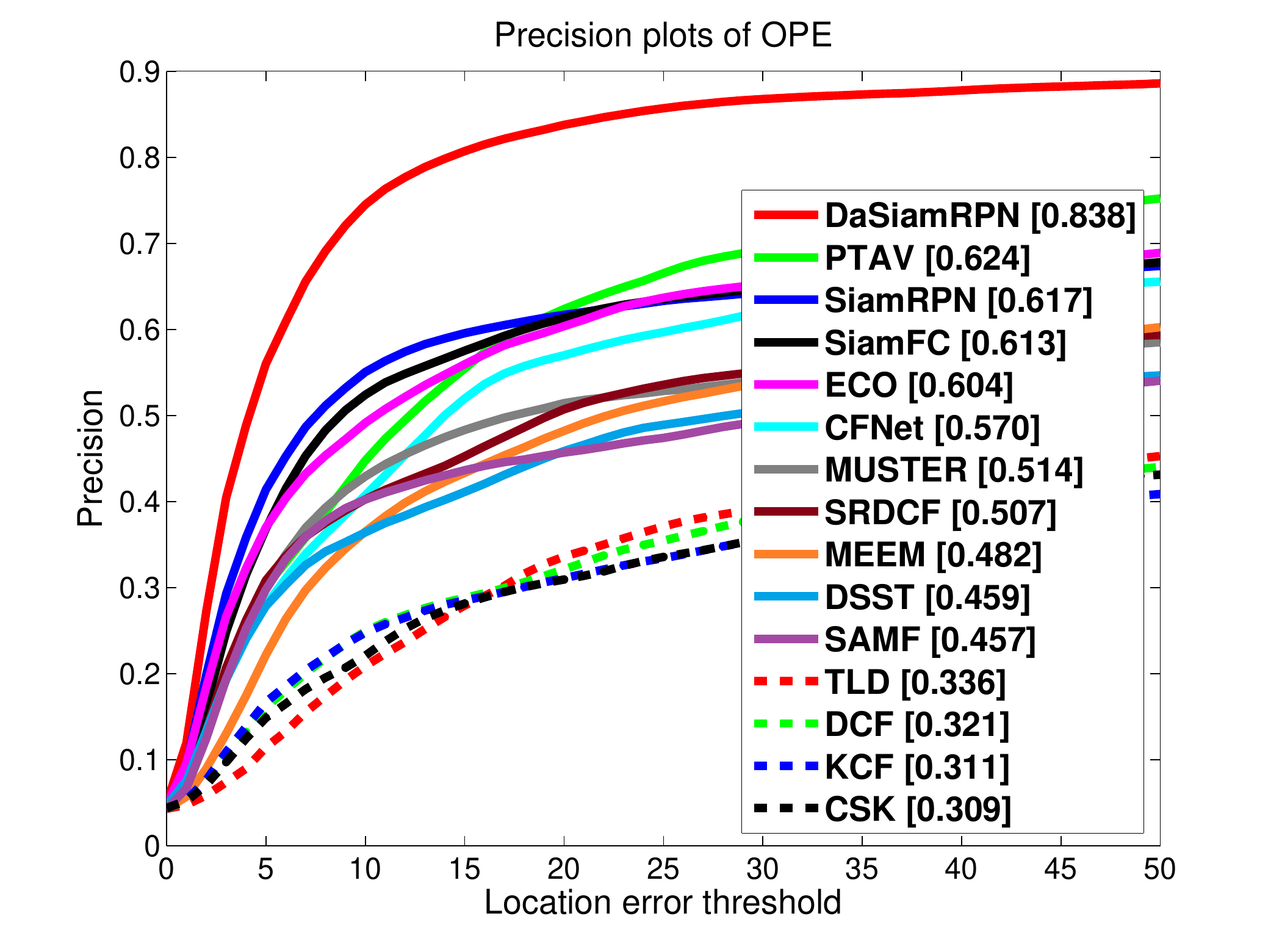}
\end{minipage}%
 \begin{minipage}[c]{3cm}
\includegraphics[width=3cm]{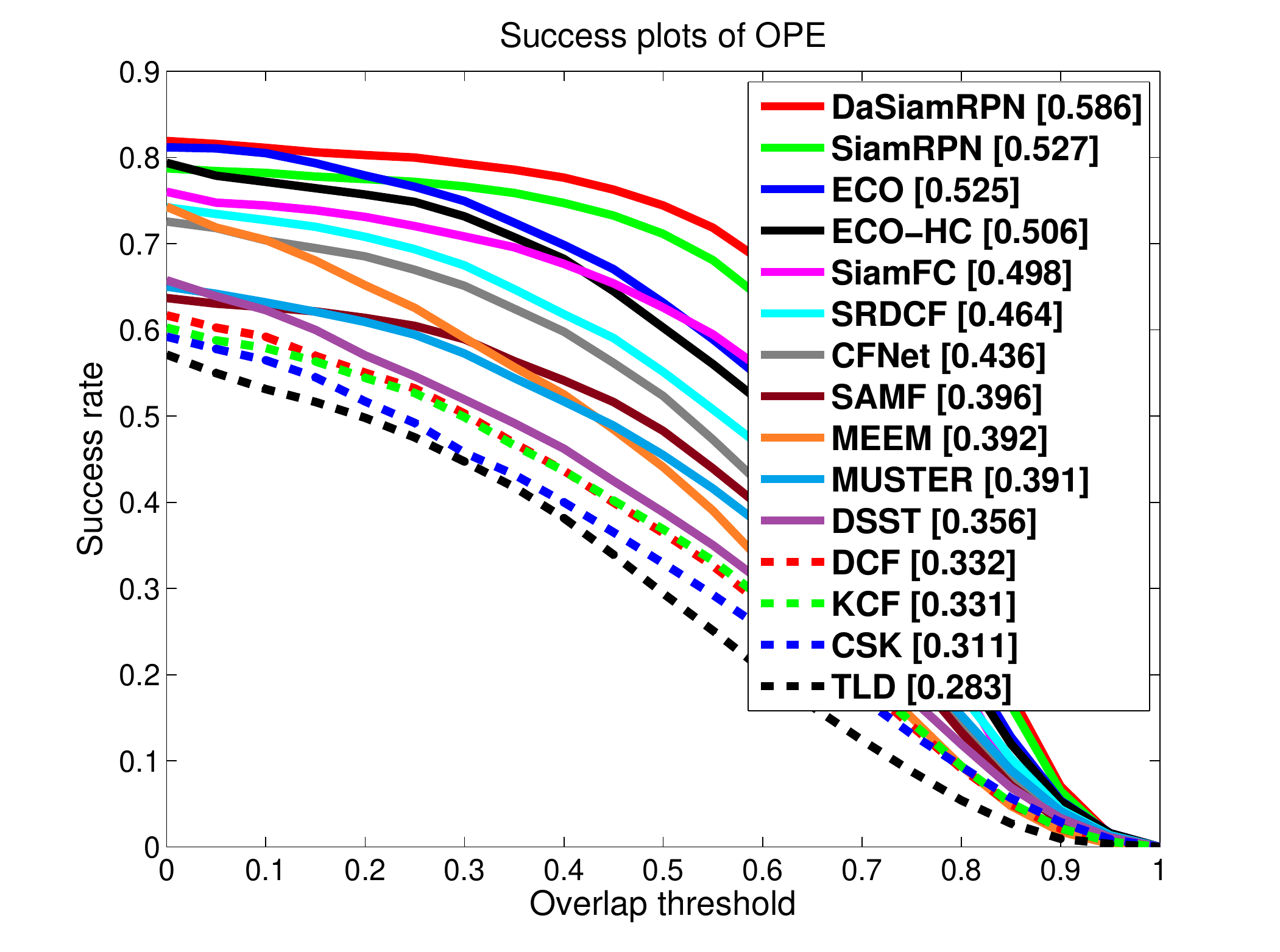}
\end{minipage}%
\begin{minipage}[c]{3cm}
\includegraphics[width=3cm]{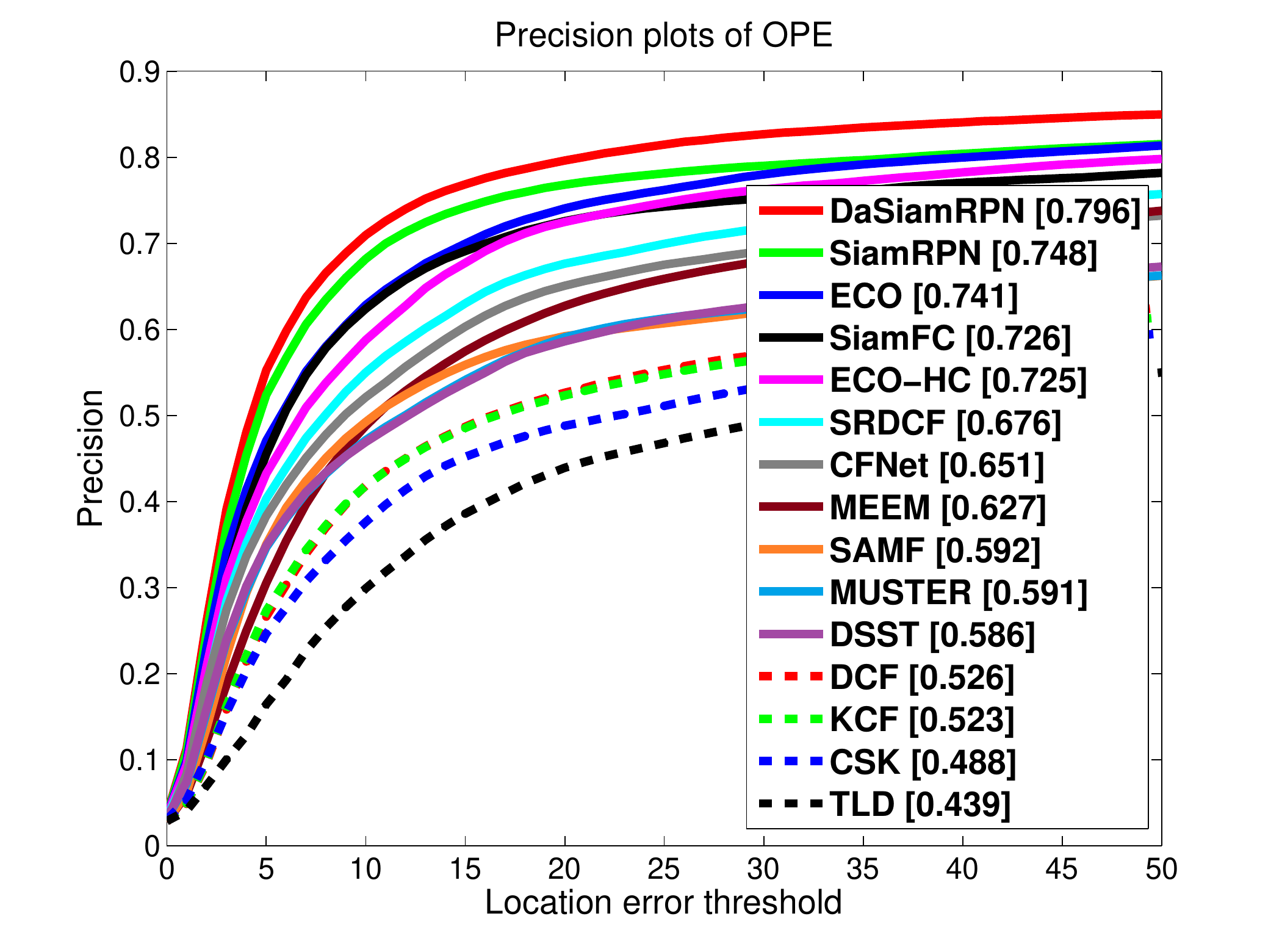}
\end{minipage}%
\caption{Success and precision plots on UAV~\cite{UAV} dataset. First and second sub-figures are results of UAV20L, third and last sub-figures are results of UAV123.}
\label{UAV_ALL}

\end{figure*}

\subsubsection{Results on UAV20L}
UAV20L is a long-term tracking benchmark that contains 20 sequences with average sequence length 2934 frames. Besides the challenging situations in short-term tracking, severe out-of-view and full occlusion introduce extra challenges. In this experiment, the proposed method is compared against recent trackers in \cite{UAV}. Besides, ECO~\cite{ECO} (state-of-the-art short-term tracker), PTAV~\cite{PTAV} (state-of-the-art long-term tracker), SiamRPN~\cite{SiamRPN} (the baseline), SiamFC~\cite{SiamFC} and CFNet~\cite{CFNet} (representative Siamese trackers) are added for comparison.

The results including success plots and precision plots are illustrated in Fig.~\ref{UAV_ALL}. It clearly illustrates that our algorithm, denoted by DaSiamRPN, outperforms the state-of-the-art trackers significantly in both measures. In the success plot, our approach obtains an AUC score of 0.617, significantly outperforming state-of-the-art short-term trackers SiamRPN~\cite{SiamRPN} and ECO~\cite{ECO}. The improvement ranges are relative 35.9\% and 41.8\%, respectively. Compared with PTAV~\cite{PTAV}, MUSTer~\cite{muster} and TLD~\cite{TLD} which are qualified to perform long-term tracking, the proposed DaSiamRPN outperforms these trackers by relative 45.8\%, 87.5\% and 213.2\%.
In the precision plot, our approach obtains a score of 0.838, outperforming state-of-the-art long-term tracker (PTAV~\cite{PTAV}) and short-term tracker (SiamRPN~\cite{SiamRPN}) by relative 34.3\% and 35.8\%, respectively. The excellent performance of DaSiamRPN in this long-term tracking dataset can be attributed to the distractor-aware features and local-to-global search strategy.

For detailed performance analysis, we also report the results on various challenge attributes in UAV20L, i.e. full occlusion, out-of-view, background clutter and partial occlusion. Fig.~\ref{UAV20L_attributes} demonstrates that our tracker effectively handles these challenging situations while other trackers obtain lower scores. Specially, in full occlusion and background clutter attributes, the proposed DaSiamRPN outperforms SiamRPN~\cite{SiamRPN} by relative 153.1\% and 393.2\%.

\begin{figure*}[!tp]
 \centering
\begin{minipage}[c]{3cm}
\includegraphics[width=3cm]{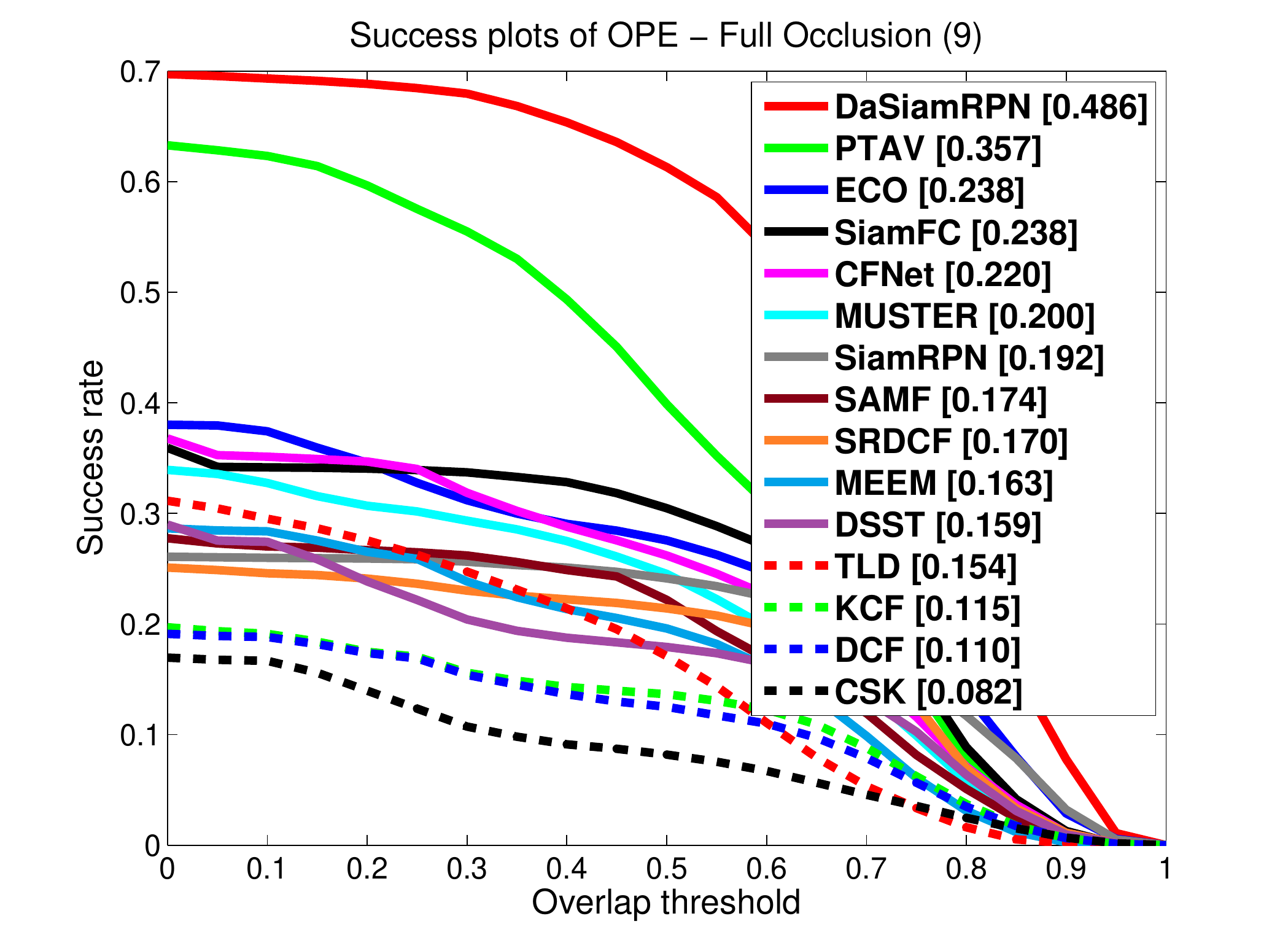}
\end{minipage}%
\begin{minipage}[c]{3cm}
\includegraphics[width=3cm]{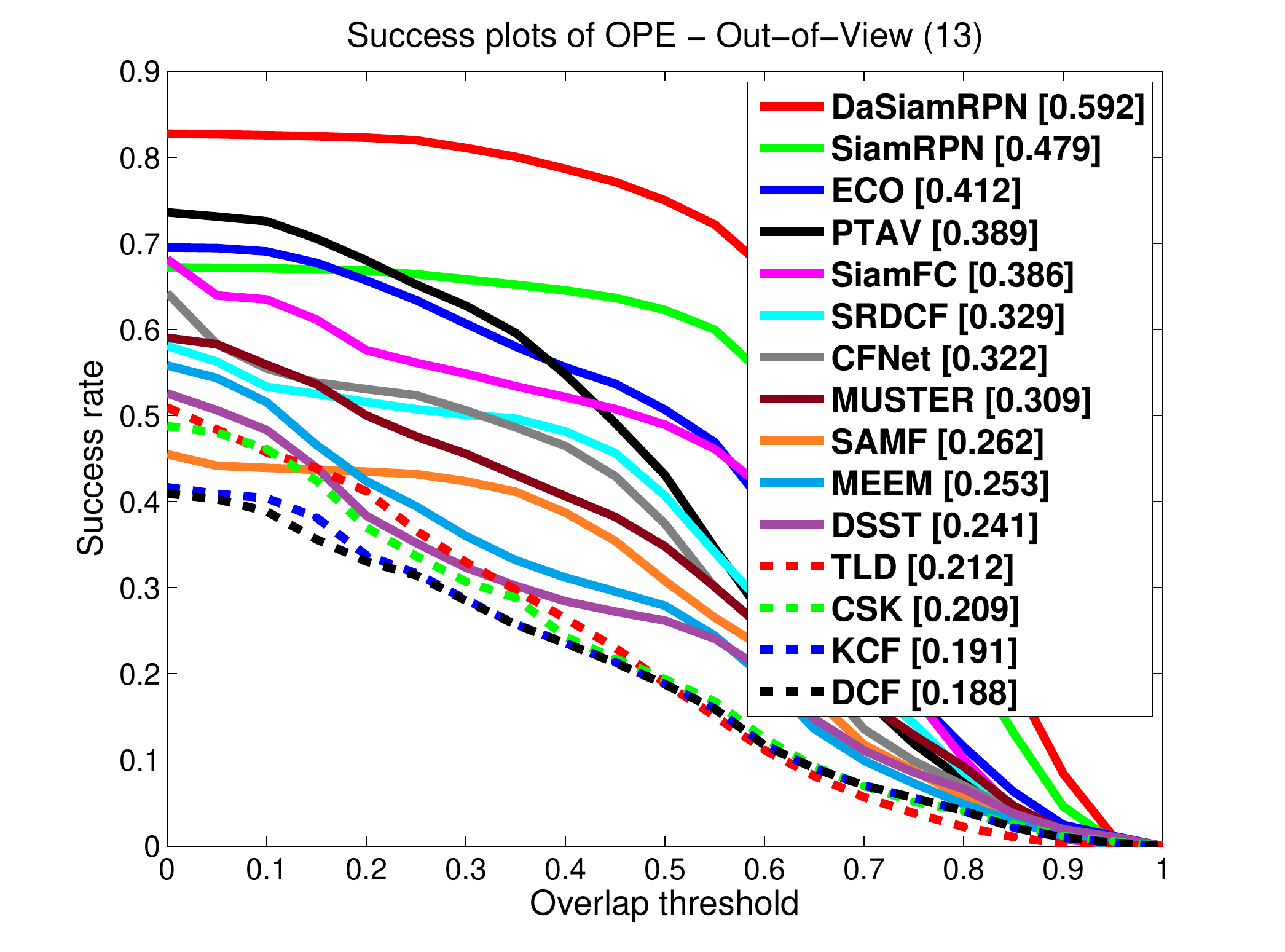}
\end{minipage}%
\begin{minipage}[c]{3cm}
\includegraphics[width=3cm]{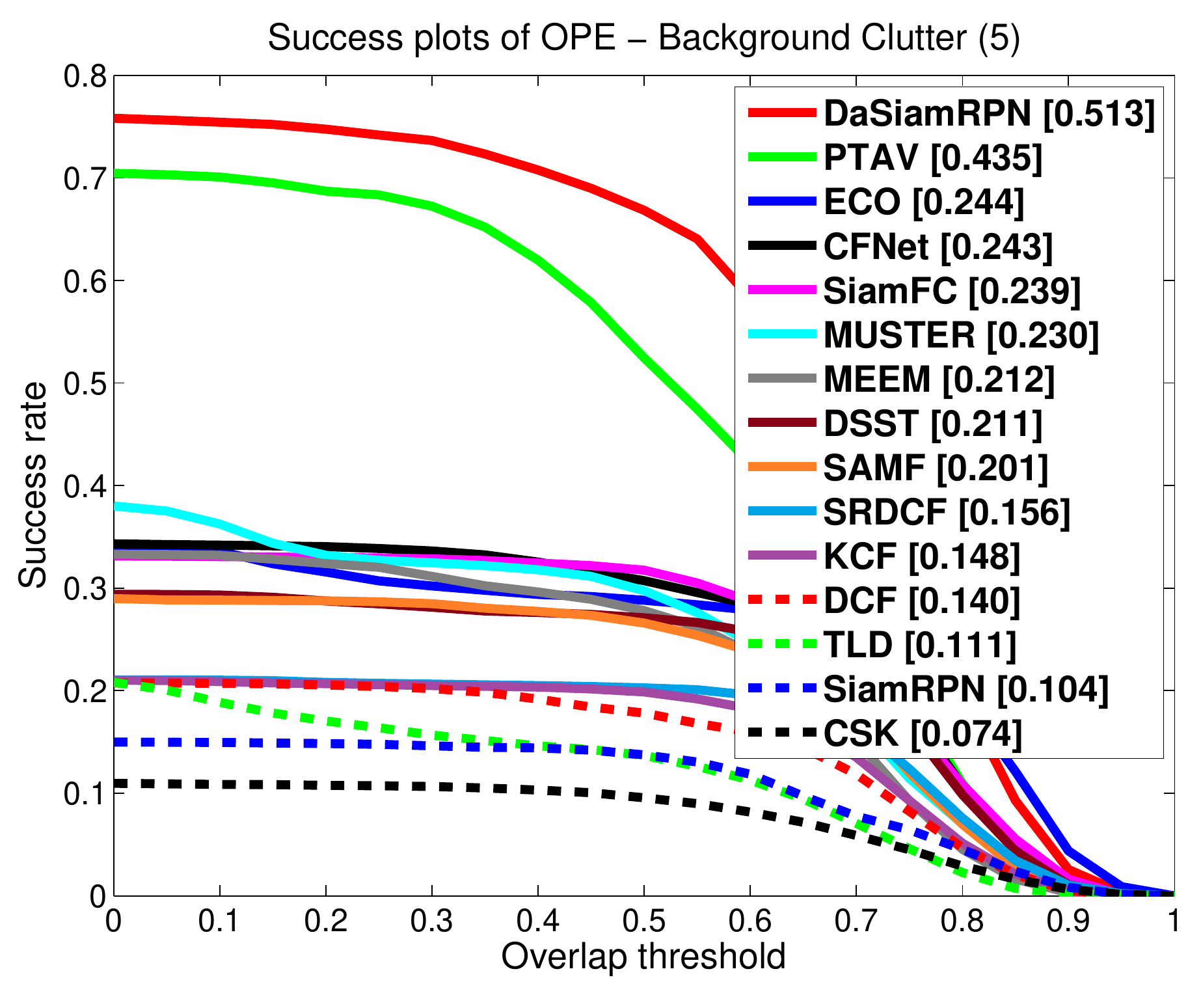}
\end{minipage}%
\begin{minipage}[c]{3cm}
\includegraphics[width=3cm]{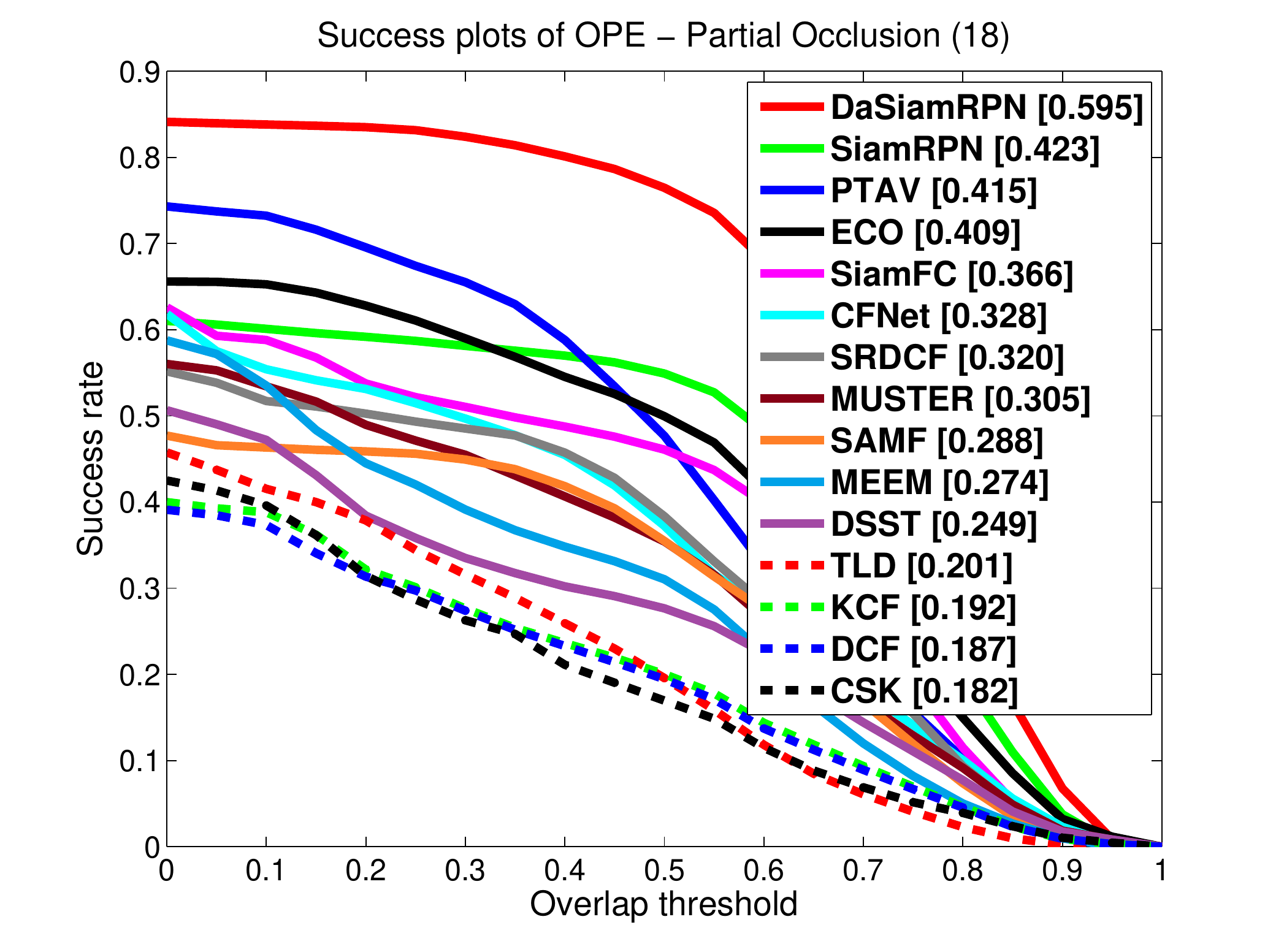}
\end{minipage}%
\caption{Success plots with attributes on UAV20L. Best viewed on color display.}
\label{UAV20L_attributes}

\end{figure*}

\subsubsection{Results on UAV123}
UAV123 dataset includes 123 sequences with average sequence length of 915 frames. Besides the recent trackers in~\cite{UAV}, ECO~\cite{ECO}, PTAV~\cite{PTAV}, SiamRPN~\cite{SiamRPN}, SiamFC~\cite{SiamFC}, CFNet~\cite{CFNet} are added for comparison. Fig.~\ref{UAV_ALL} illustrates the precision and success plots of the compared trackers. The proposed DaSiamRPN approach outperforms all the other trackers in terms of success and precision scores. Specifically, our method achieves a success score of 0.586, which outperforms the SiamRPN (0.527) and ECO (0.525) method with a large margin.

\subsection{State-of-the-Art Comparisons on OTB Datasets}

\label{sect:otb}

We evaluate the proposed algorithms with numerous fast and state-of-the-art trackers including SiamFC~\cite{SiamFC}, CFNet~\cite{CFNet}, Staple~\cite{Staple}, CSRDCF~\cite{CSRDCF}, BACF~\cite{BACF}, ECO-HC~\cite{ECO}, CREST~\cite{CREST}, MDNet~\cite{MDNet}, CCOT~\cite{CCOT}, ECO~\cite{ECO}, and the baseline tracker SiamRPN~\cite{SiamRPN}. All the trackers are initialized with the ground-truth object state in the first frame. Mean overlap precision (OP) and mean distance precision (DP) are reported in Table~\ref{tab:component}.

Among the real-time trackers, SiamFC and CFNet are latest Siamese network based trackers while the accuracies is still left far behind the state-of-the-art BACF and ECO-HC with HOG features. The proposed DaSiamRPN tracker outperforms all these trackers by a large margin on both the accuracy and speed.

For state-of-the-art comparisons on OTB, MDNet, trained on visual tracking datasets, performs the best against the other trackers at a speed of 1 FPS. C-COT and ECO achieve state-of-the-art performance, but their tracking speeds are not fast enough for real-time applications. The baseline tracker SiamRPN obtains an OP score of $81.9\%$, which is slightly less accurate than CCOT. The bottleneck of SiamRPN is its inferior robust performance. Since the distractor-aware mechanisms in both training and inference focus on improving the robustness, the proposed DaSiamRPN tracker achieves $3.0\%$ improvement on DP and performs best OP score of $86.5\%$ on OTB2015.

\subsection{Ablation Analyses}

To verify the contributions of each component in our algorithm, we implement and evaluate four variations of our approach. Analyses results include EAO on VOT2016~\cite{VOT2016} and AUC on UAV20L~\cite{UAV}.

As shown in Table~\ref{tab:ablation}, SiamRPN is our baseline algorithm. In VOT2016, the EAO criterion increases to 0.368 from 0.344 when detection data is added in training. Similarly, when negative pairs and distractor-aware learning are adopted in training and inference, both the  performance increases by near 2\%. In UAV20L, detection data, negative pairs in training and distractor-aware inference gain the performance by 1\%-2\%. The AUC criterion increases to 61.7\% from 49.8\% when long-term tracking module is adopted.

\begin{table}[t]
\centering
\caption{Ablation analyses of our algorithm on VOT2016~\cite{VOT2016} and UAV20L~\cite{UAV}}

\setlength{\tabcolsep}{6.0pt}
\begin{tabular}{p{5cm}<{\centering}|p{0.7cm}<{\centering}|p{0.7cm}<{\centering}p{0.7cm}<{\centering}p{0.7cm}<{\centering}p{0.7cm}<{\centering}}
\toprule[1.5pt]
\multicolumn{1}{c|}{Component}&\multicolumn{1}{c|}{SiamRPN}&\multicolumn{4}{c}{DaSiamRPN}\\
\hline
positive pairs in detection data?                    &  & \Checkmark & \Checkmark & \Checkmark & \Checkmark \\
semantic negative pairs?            &  &            & \Checkmark & \Checkmark & \Checkmark \\
distractor-aware updating?   &  &            &            & \Checkmark & \Checkmark \\
long-term tracking module?         &  &            &            &            & \Checkmark \\
\hline
EAO in VOT2016                     &  ~~~~~0.344 &   0.368   & 0.389    &     0.411  & --\\
AUC in UAV20L(\%)                  &  ~~~~~~45.4  &    47.2   & 48.6    &  49.8 & 61.7\\
\bottomrule[1.5pt]
\end{tabular}
\label{tab:ablation}

\end{table}

%\subsection{Qualitative Results}
%To visualize the superiority of the proposed framework, we show examples of the DaSiamRPN results compared to recent trackers (SiamRPN, SiamFC and PTAV) on challenging sample videos. As shown in Figure.~\ref{vis}, the proposed DaSiamRPN can handle the challenges while SiamRPN and SiamFC tend to drift to distractor. PTAV also adopts long-term component, but fails to track in the second and last videos. The superiority performance of our algorithm can be attributed to the design of distractor-aware Siamese networks and local-to-global search strategy.
%\begin{figure}[t]
%\centering
%\includegraphics[width=1\linewidth]{./VIS/vis.pdf}
%\caption{The qualitative results of the DaSiamRPN and compared trackers.}
%\label{vis}

%\end{figure}

\section{Conclusions}

In this paper, we propose a distractor-aware Siamese framework for accurate and long-term tracking.
During offline training, a distractor-aware feature learning scheme is proposed, which can significantly boost the discriminative power of the networks.
During inference, a novel distractor-aware module is designed, effectively transferring the general embedding to the current video domain.
In addition, we extend the proposed approach for long-term tracking by introducing a simple yet effective local-to-global search strategy.
The proposed tracker obtains state-of-the-art accuracy in comprehensive experiments of short-term and long-term visual tracking benchmarks, while the overall system speed is still far from being real-time.

\bibliographystyle{splncs04}
\bibliography{0188}

\clearpage

\section{Supplementary Material}

\subsection{Visualization of Siamese Network tracker}

\begin{figure*}[htbp]
\captionsetup{font={small}}
\centering
\subfloat[\textit{bolt2}]{\includegraphics[width=0.9\linewidth]{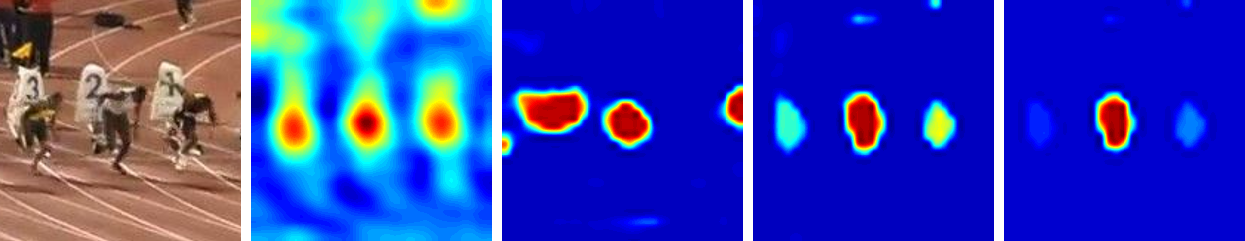}}
\hfil
\subfloat[\textit{gymnastics4}]{\includegraphics[width=0.9\linewidth]{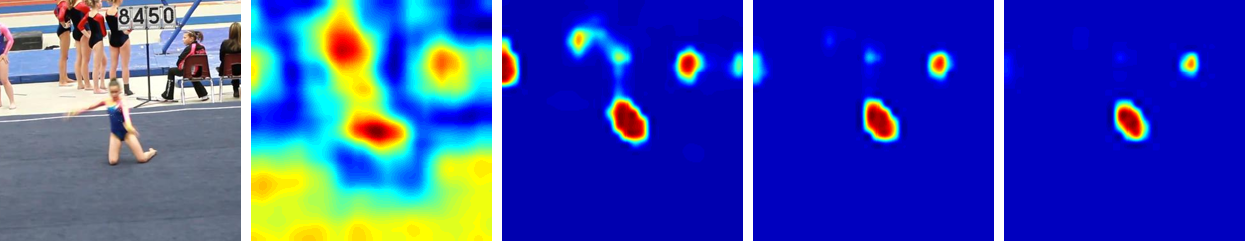}}
\hfil
\subfloat[\textit{bird2}]{\includegraphics[width=0.9\linewidth]{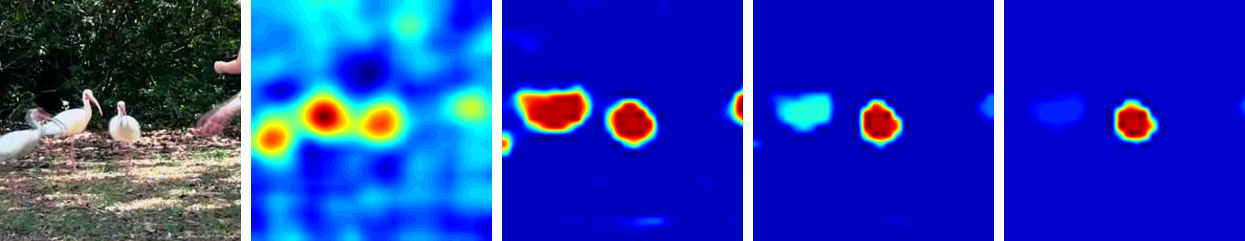}}
\caption{Comparisons with general Siamese network trackers. From left to right: original images; response maps from SiamFC; SiamRPN; SiamRPN+(our method); and DaSiamRPN(our method).}
\label{fig:heatmap}
\end{figure*}

In this section, we provide further visualization of the response maps on VOT dataset. Fig.~\ref{fig:heatmap} shows the heatmaps of SiamFC, SiamRPN, SiamRPN+, and DaSiamRPN. The proposed DaSiamRPN generates more discriminative response maps across all different videos.

\subsection{Detailed results on VOT}
\begin{figure}[htbp]
\centering
\subfloat[\footnotesize Expected overlap curves]{\includegraphics[width=0.48\linewidth,trim=5.8cm 0cm 0cm 0cm, clip]{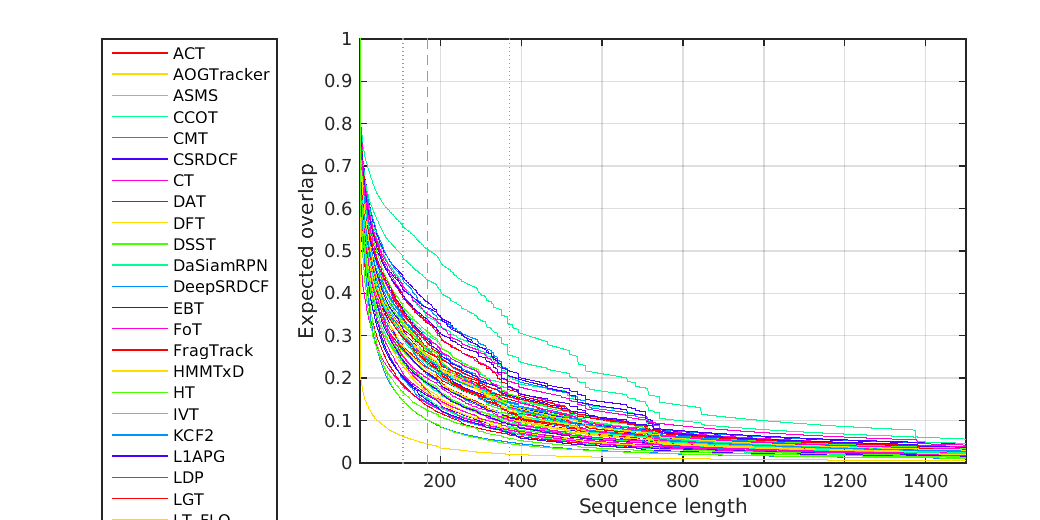}}
\subfloat[\footnotesize Ranking plot]{\includegraphics[width=0.33\linewidth]{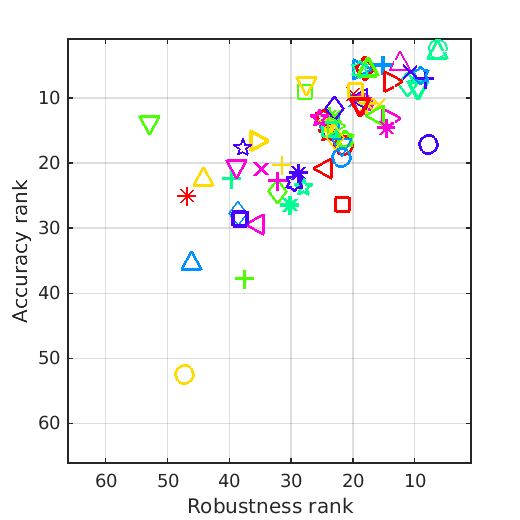}}
\hfil
\includegraphics[width=1\linewidth]{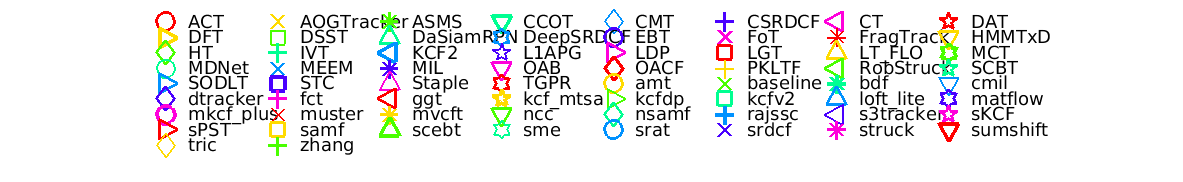}
\caption{Expected overlap curves and ranking plot for DaSiamRPN on VOT2015.}
\label{vot2015}
\end{figure}

\begin{figure}[htbp]
\centering
\subfloat[\footnotesize Expected overlap curves]{\includegraphics[width=0.48\linewidth,trim=5.8cm 0cm 0cm 0cm, clip]{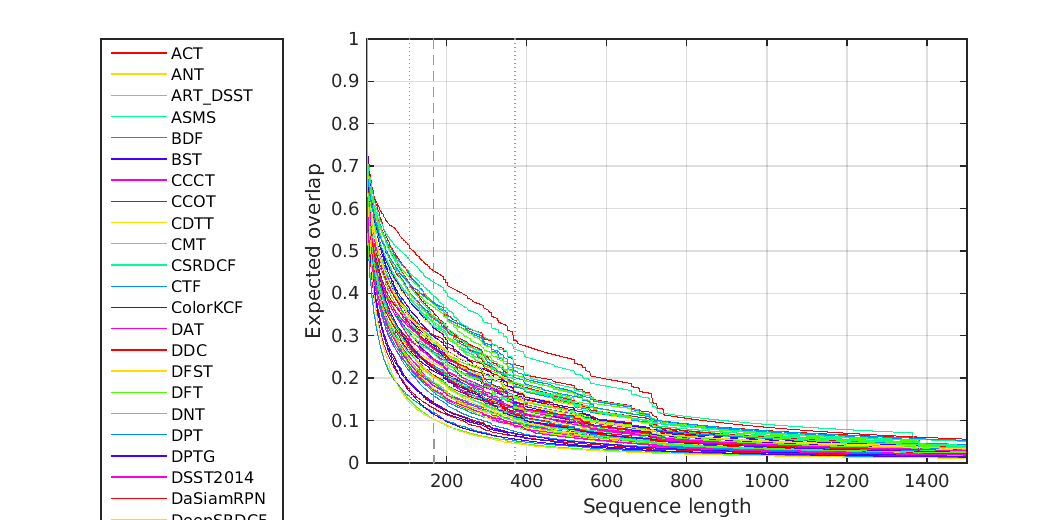}}
\subfloat[\footnotesize Ranking plot]{\includegraphics[width=0.33\linewidth]{img/rankingplot_baseline_mean2015.png}}
\hfil
\includegraphics[width=1\linewidth]{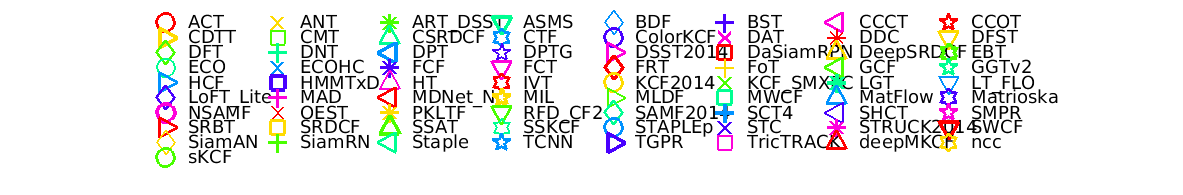}
\caption{Expected overlap curves and ranking plot for DaSiamRPN on VOT2016.}
\label{vot2016}
\end{figure}

\begin{figure}[htbp]
\centering
\subfloat[\footnotesize Expected overlap curves]{\includegraphics[width=0.48\linewidth,trim=5.8cm 0cm 0cm 0cm, clip]{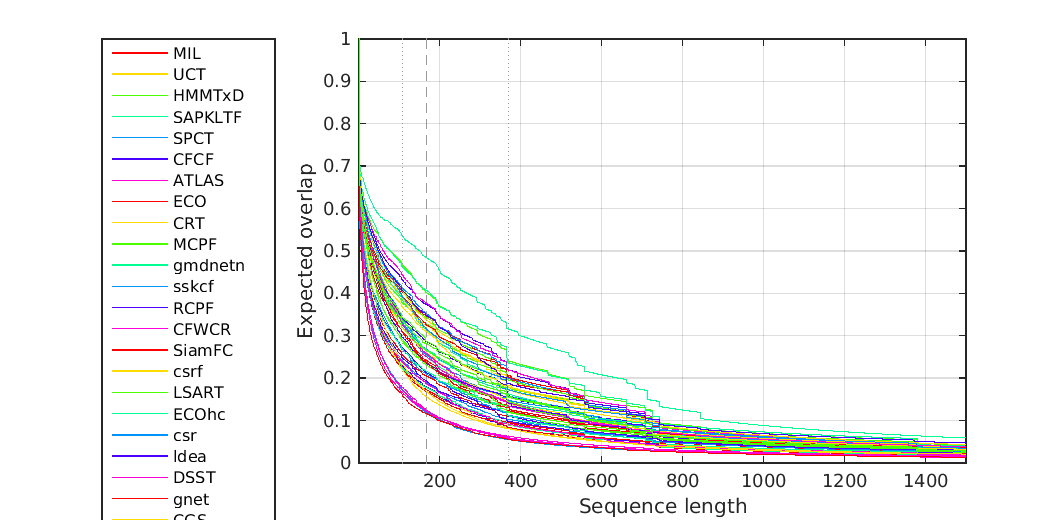}}
\subfloat[\footnotesize Ranking plot]{\includegraphics[width=0.33\linewidth]{img/rankingplot_baseline_mean2015.png}}
\hfil
\includegraphics[width=1\linewidth]{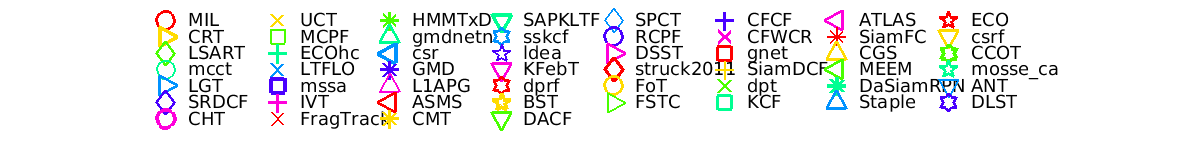}
\caption{Expected overlap curves and ranking plot for DaSiamRPN on VOT2017.}
\label{vot2017}
\end{figure}

In this section, detailed results on VOT2015, VOT2016 and VOT2017 are provided as shown in Fig.~\ref{vot2015}, Fig.~\ref{vot2016} and Fig.~\ref{vot2017}.

%\subsection{Detailed results on VOT2015}

%\subsection{Detailed results on VOT2016}

%\subsection{Detailed results on VOT2017}

\subsection{Additional results with attributes on UAV20L}
In this section, we report additional results on the UAV20L with 12 different attributes, including out-of-view, background clutter, illumination variation, viewpoint change, camera motion, similar object, scale variation, aspect ratio change, low resolution, fast motion, full occlusion, partial occlusion. The performance is ranked by area under curve (AUC) of success plot. As shown in Fig.~\ref{UAV20L_attributes}, the proposed DaSiamRPN effectively handles these challenges and achieves leading performance in all attributes.

\begin{figure*}[htbp]
 \centering
\begin{minipage}[c]{3cm}
\includegraphics[width=3cm]{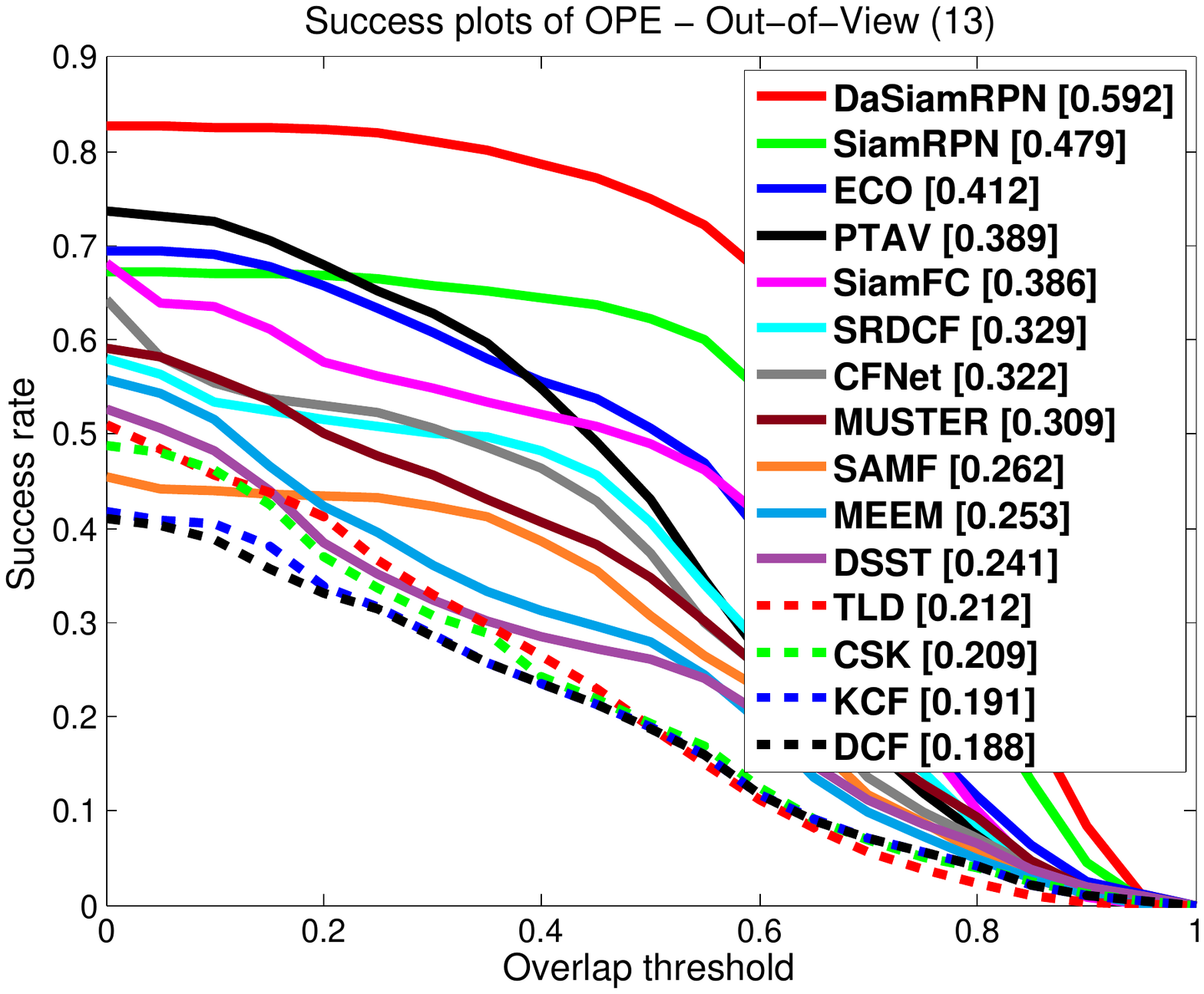}
\end{minipage}%
\begin{minipage}[c]{3cm}
\includegraphics[width=3cm]{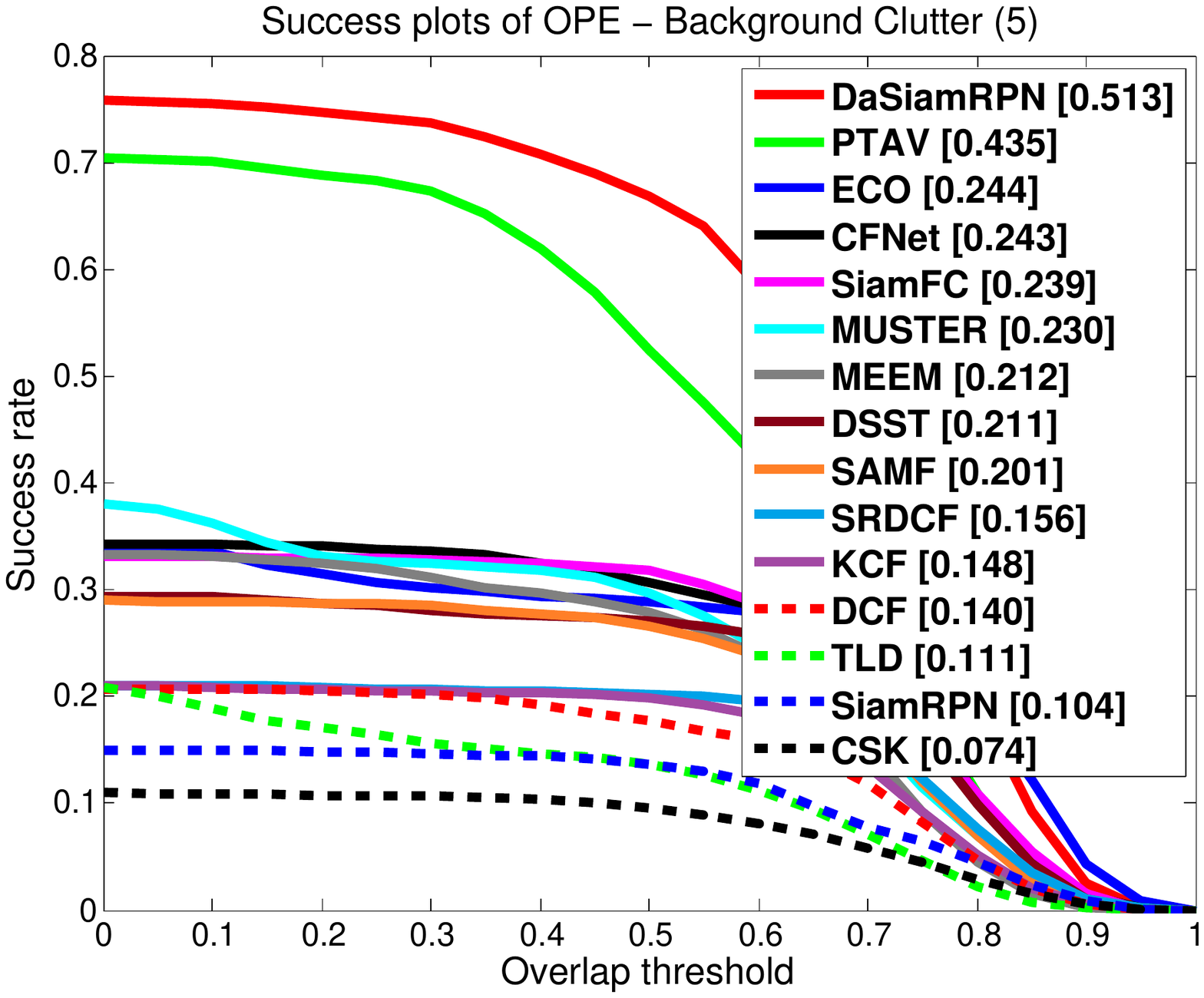}
\end{minipage}%
\begin{minipage}[c]{3cm}
\includegraphics[width=3cm]{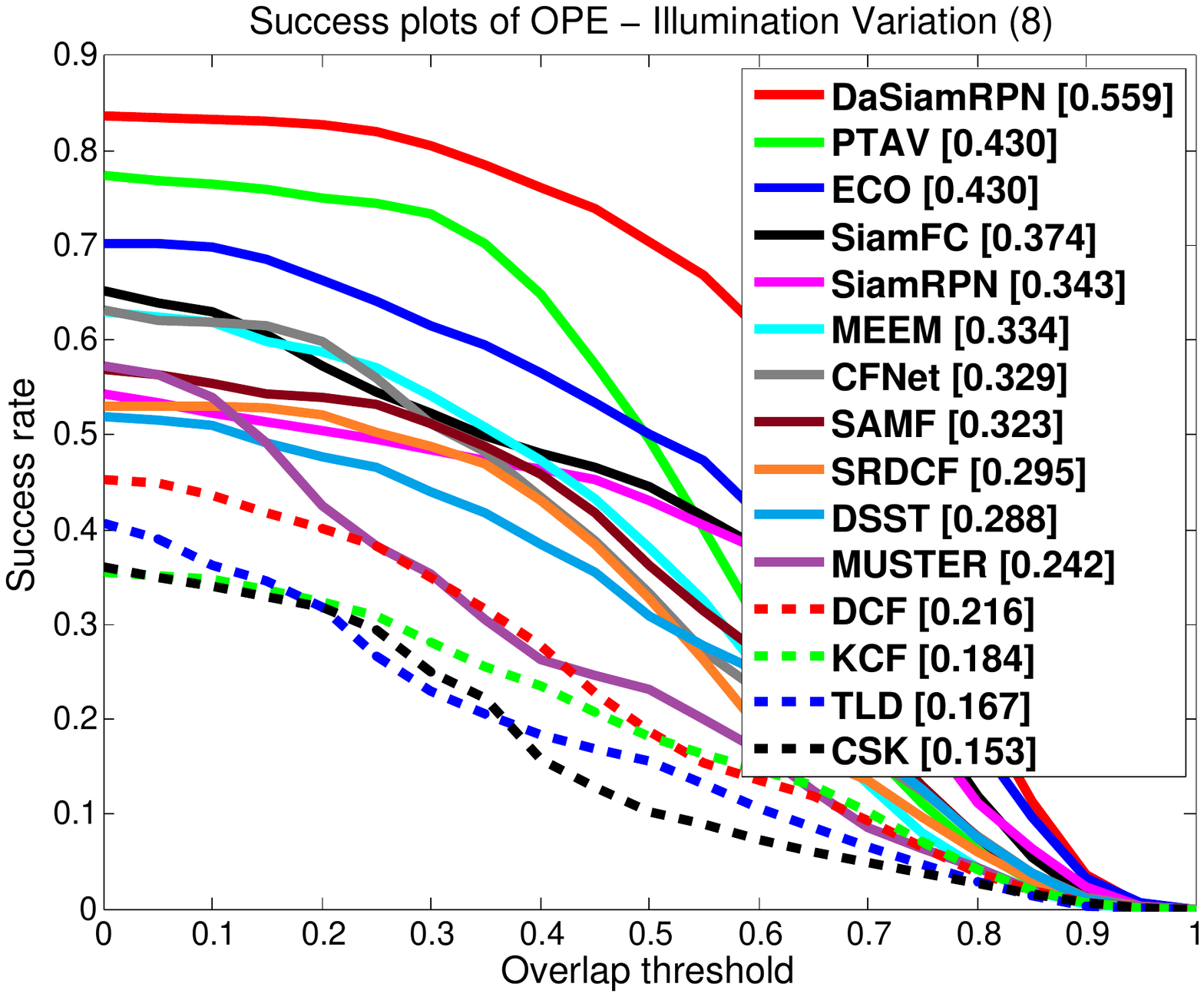}
\end{minipage}%
\begin{minipage}[c]{3cm}
\includegraphics[width=3cm]{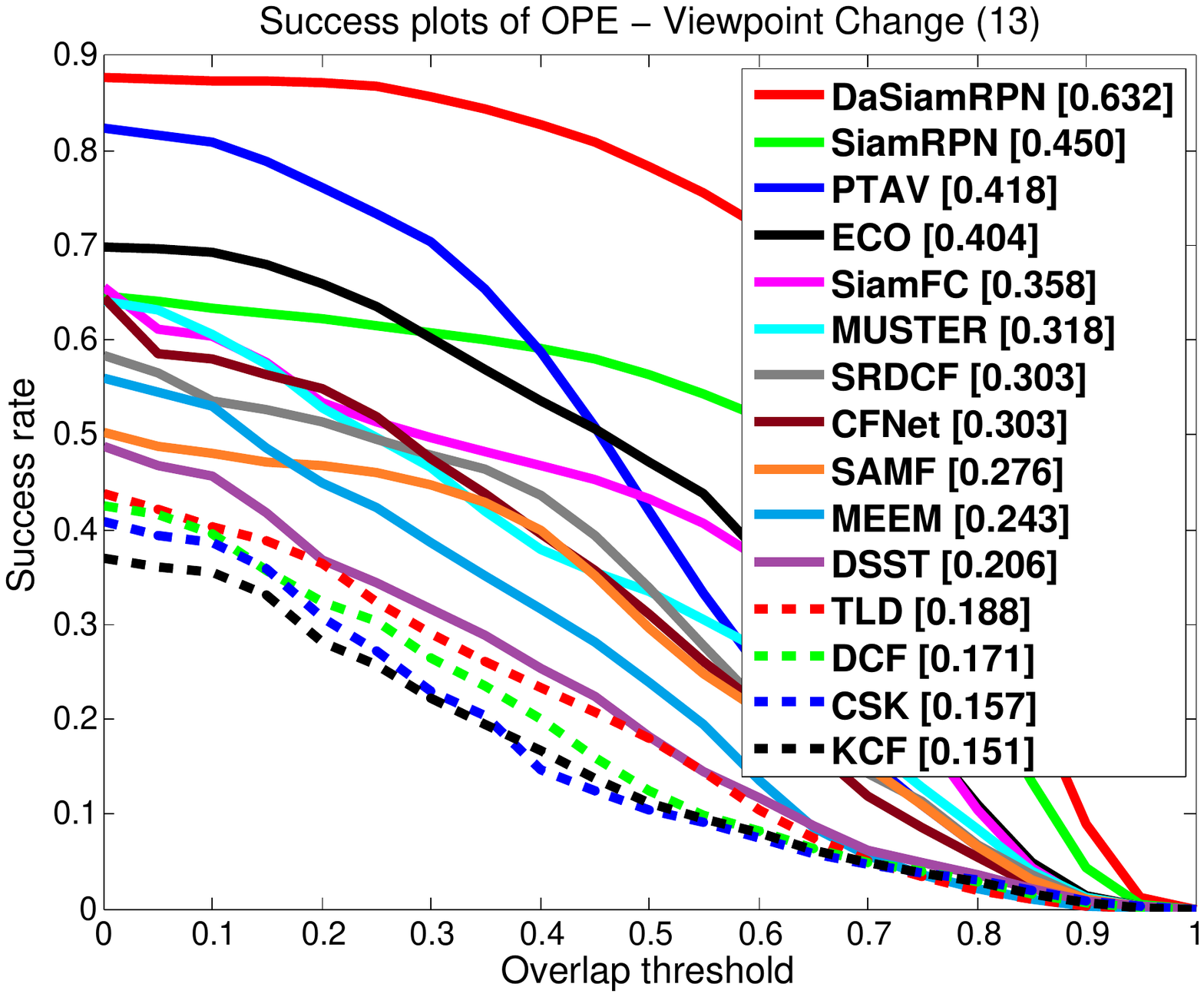}
\end{minipage}%

\begin{minipage}[c]{3cm}
\includegraphics[width=3cm]{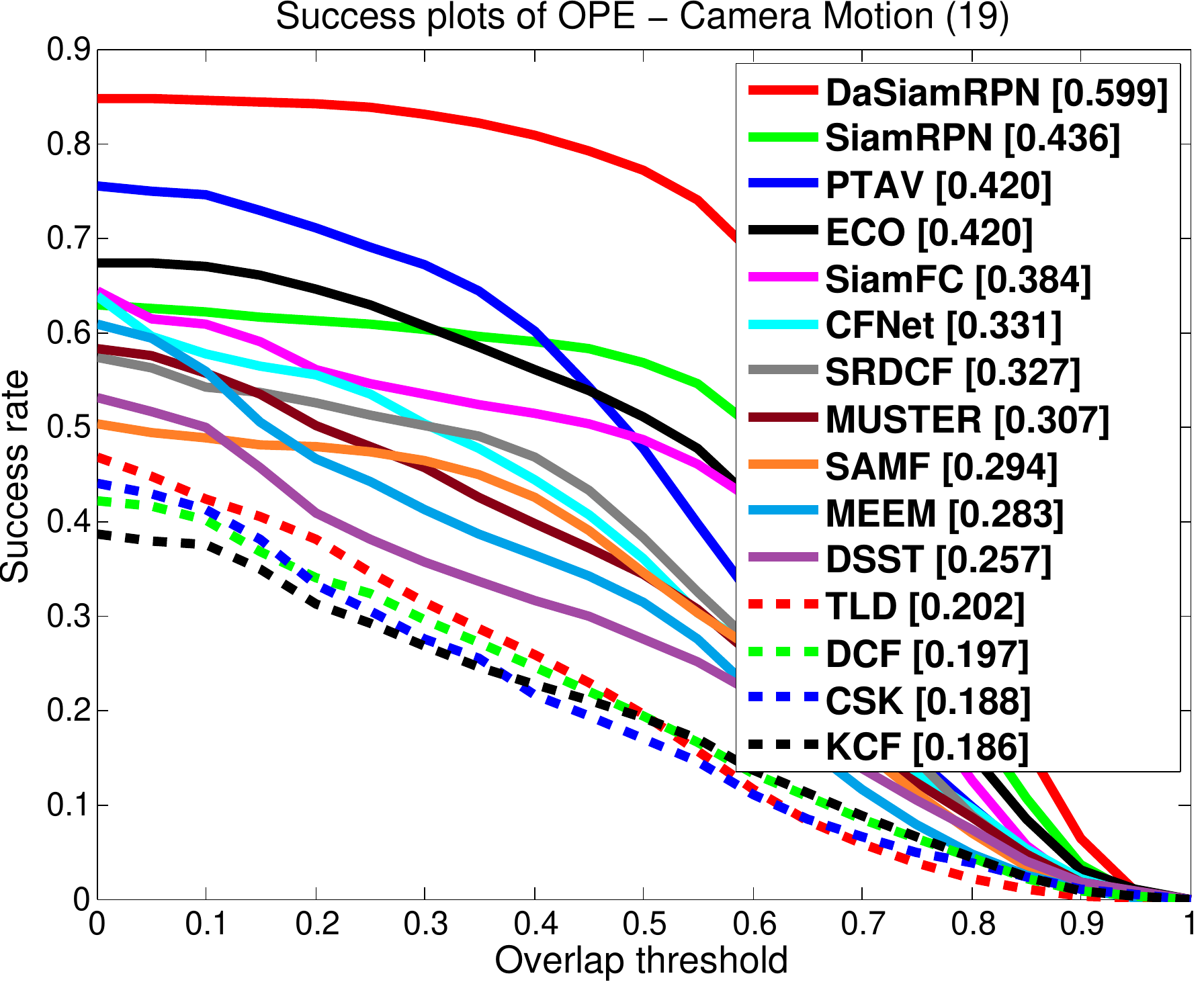}
\end{minipage}%
\begin{minipage}[c]{3cm}
\includegraphics[width=3cm]{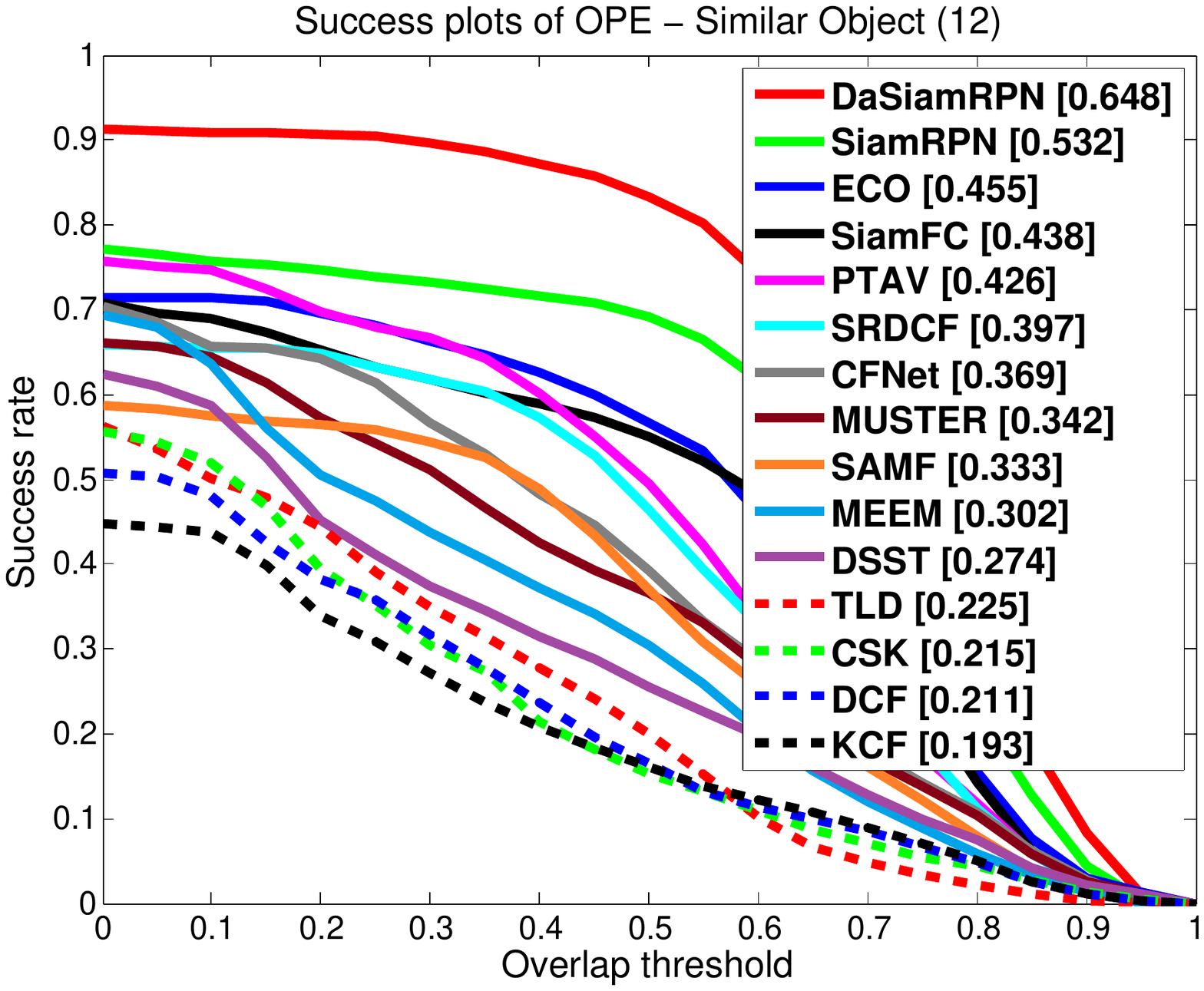}
\end{minipage}%
\begin{minipage}[c]{3cm}
\includegraphics[width=3cm]{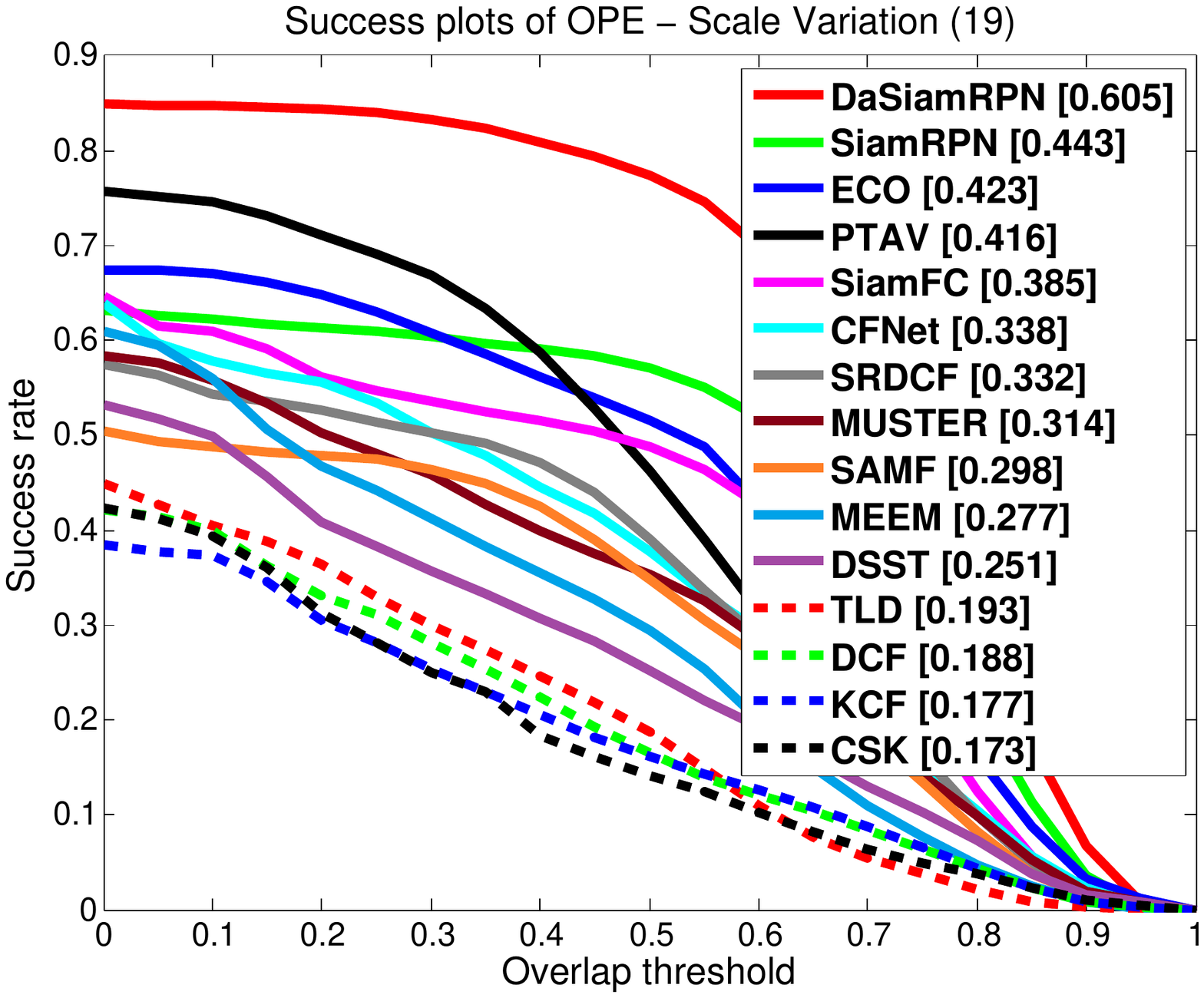}
\end{minipage}%
\begin{minipage}[c]{3cm}
\includegraphics[width=3cm]{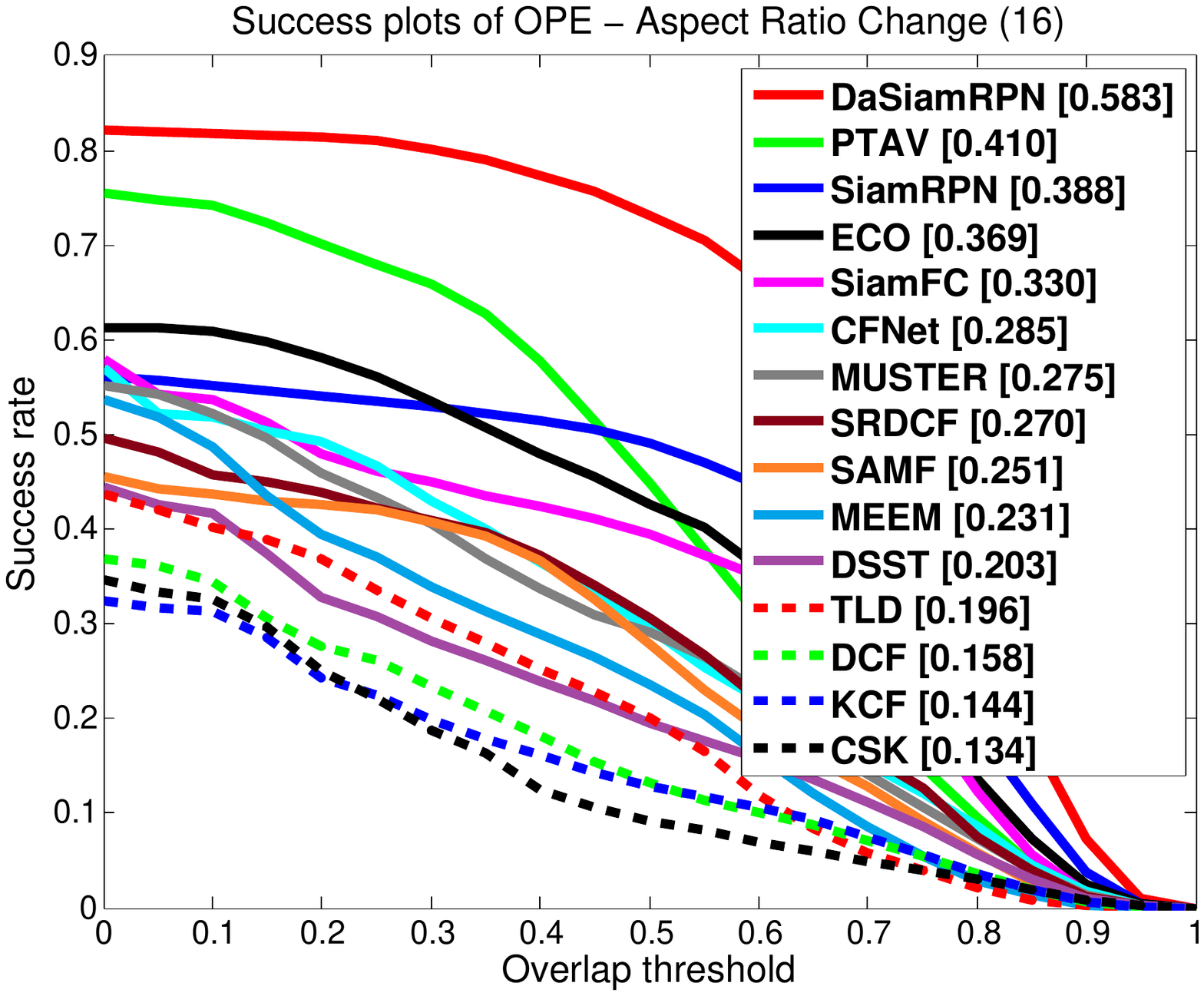}
\end{minipage}%

\begin{minipage}[c]{3cm}
\includegraphics[width=3cm]{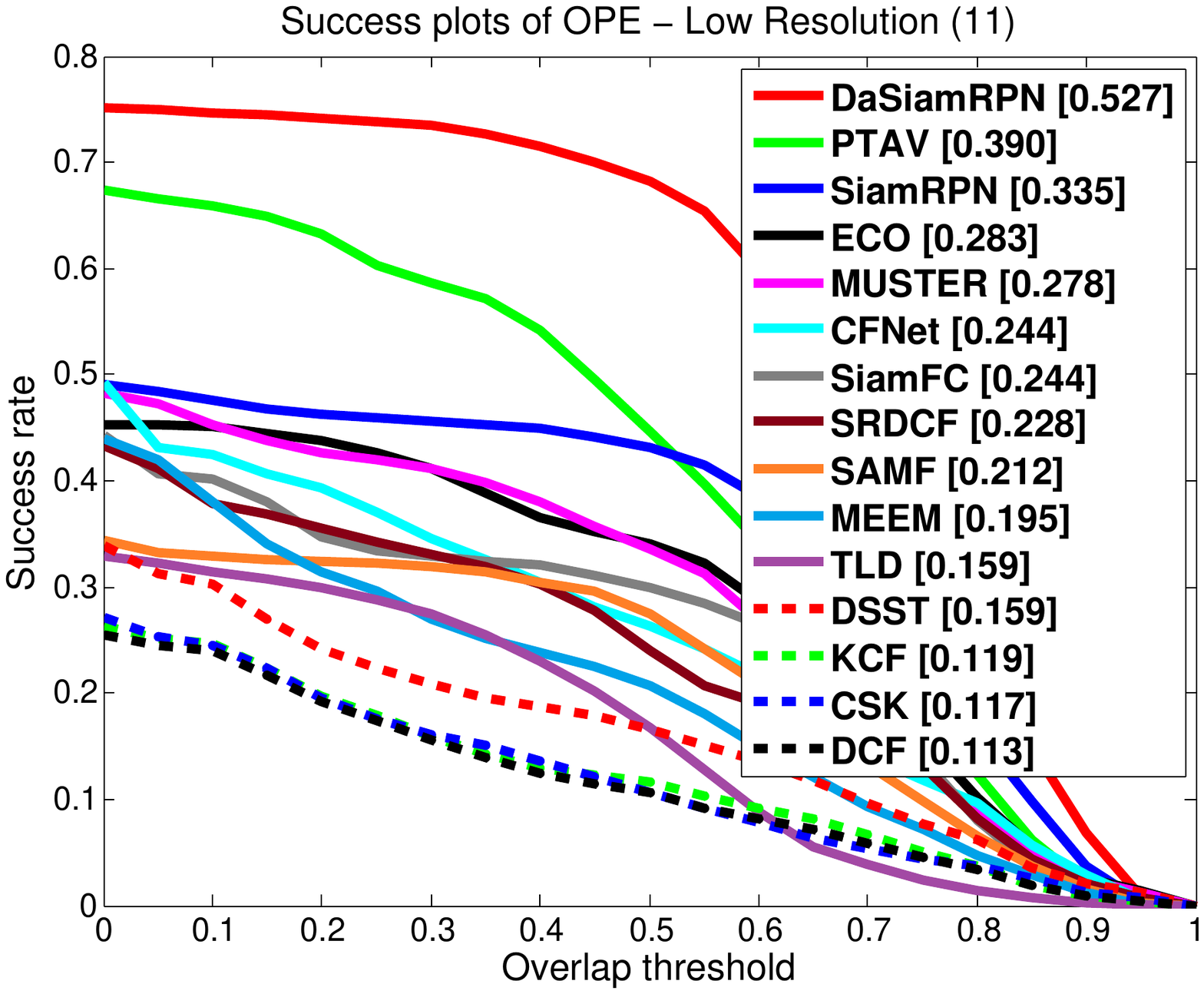}
\end{minipage}%
\begin{minipage}[c]{3cm}
\includegraphics[width=3cm]{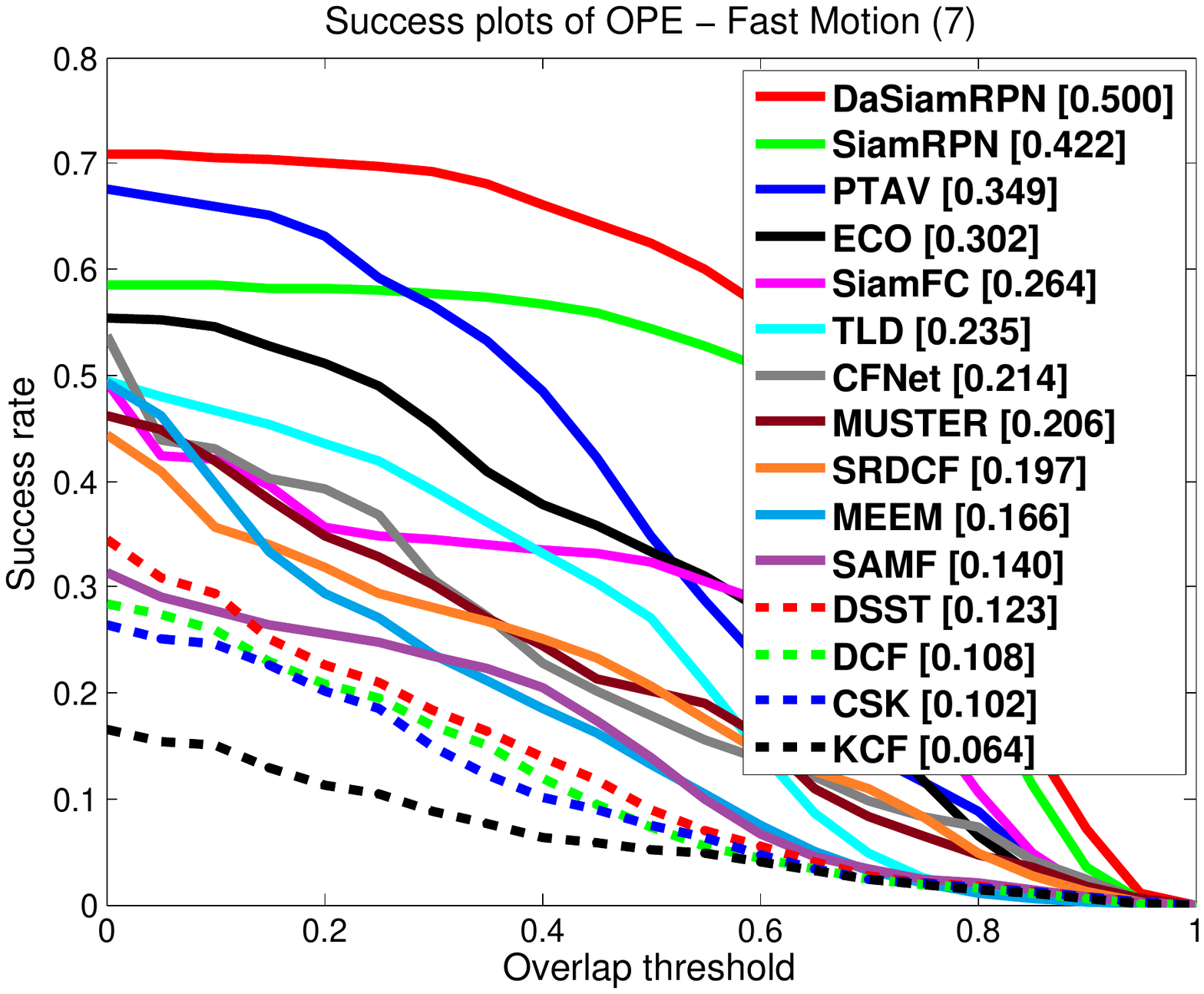}
\end{minipage}%
\begin{minipage}[c]{3cm}
\includegraphics[width=3cm]{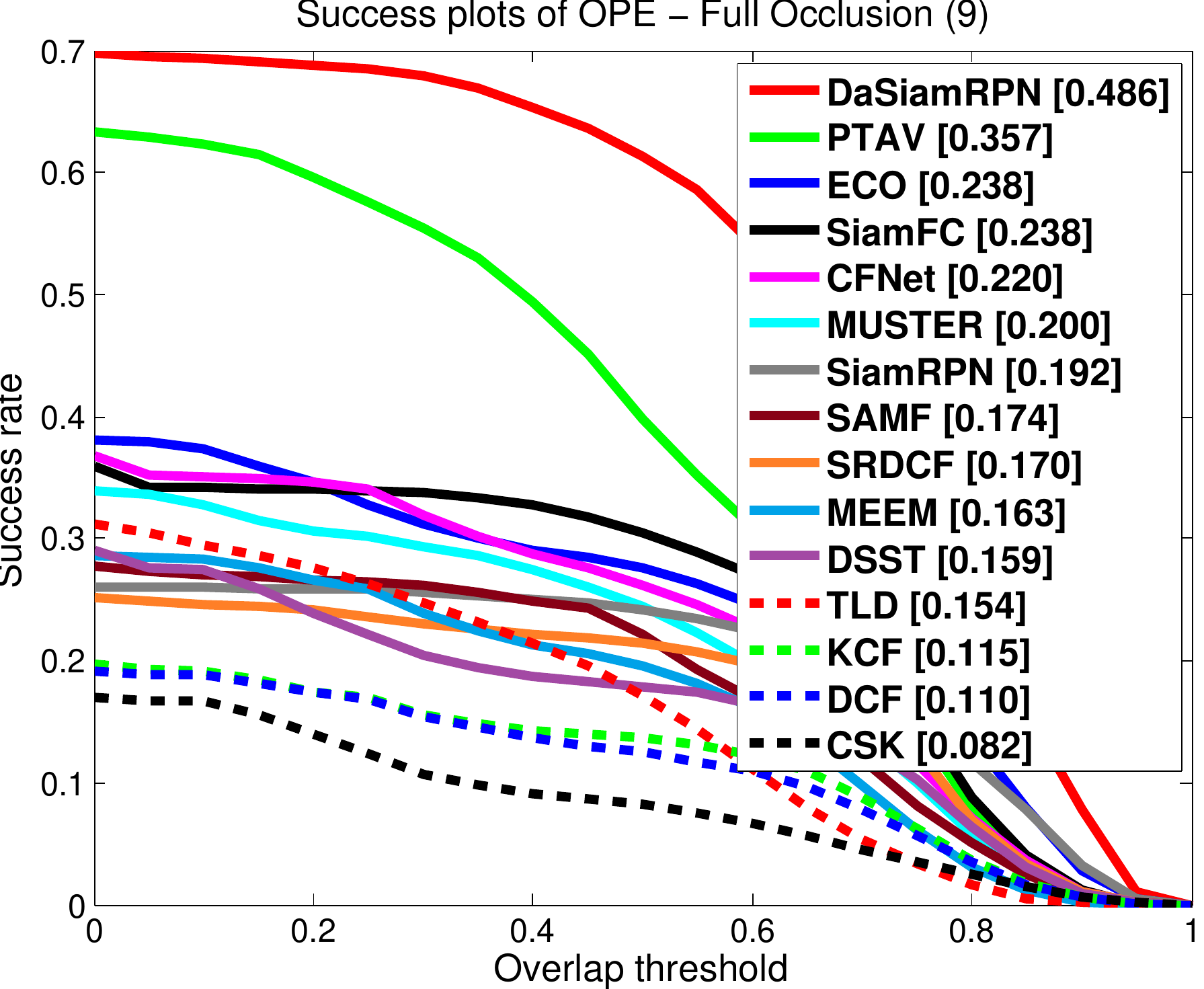}
\end{minipage}%
\begin{minipage}[c]{3cm}
\includegraphics[width=3cm]{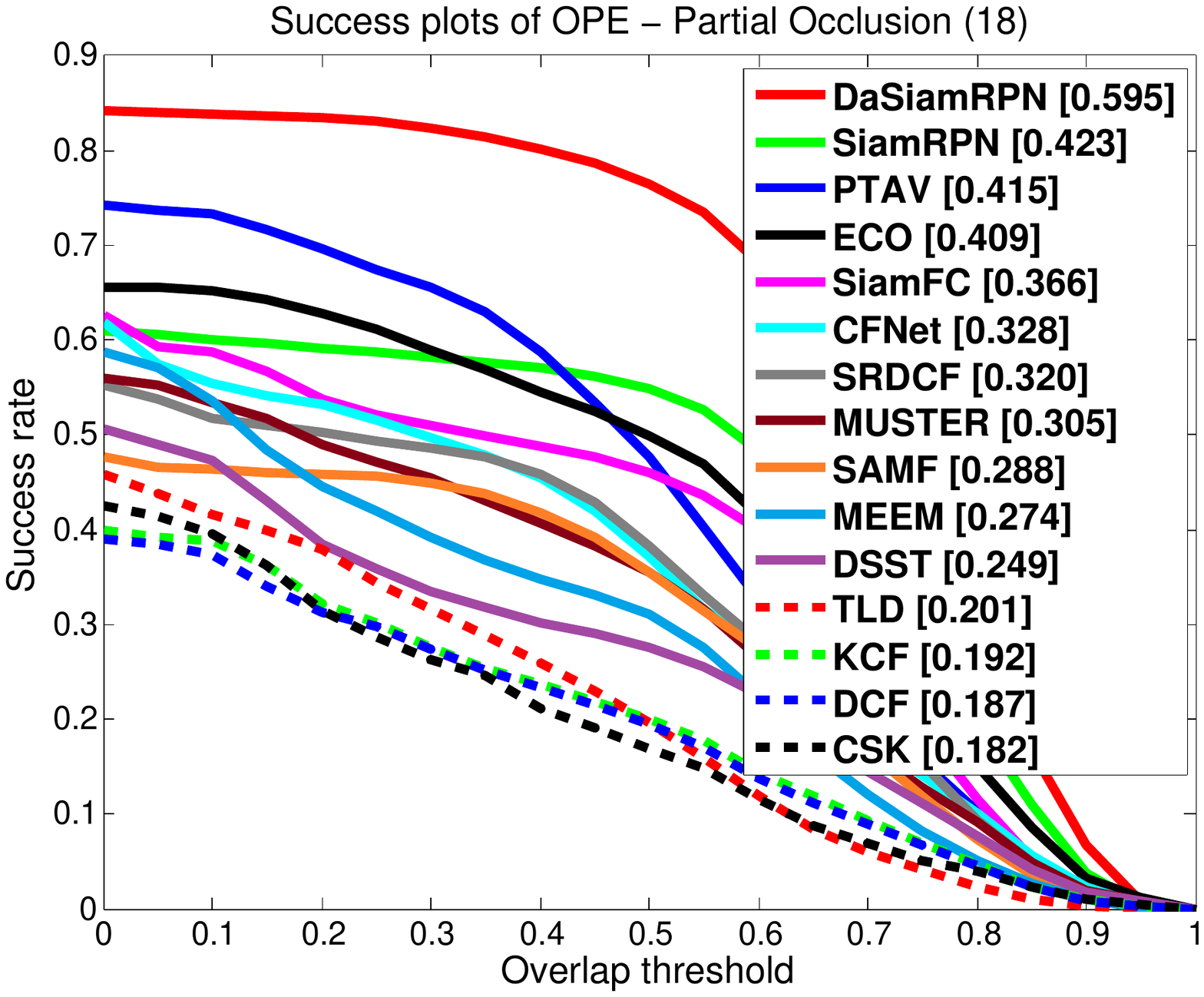}
\end{minipage}%
\caption{Success plots with attributes on UAV20L. Best viewed on color display.}
\label{UAV20L_attributes}
%\vspace{-0.5cm}
\end{figure*}

\subsection{Additional results with attributes on UAV123}
\begin{figure*}[htbp]
 \centering
\begin{minipage}[c]{3cm}
\includegraphics[width=3cm]{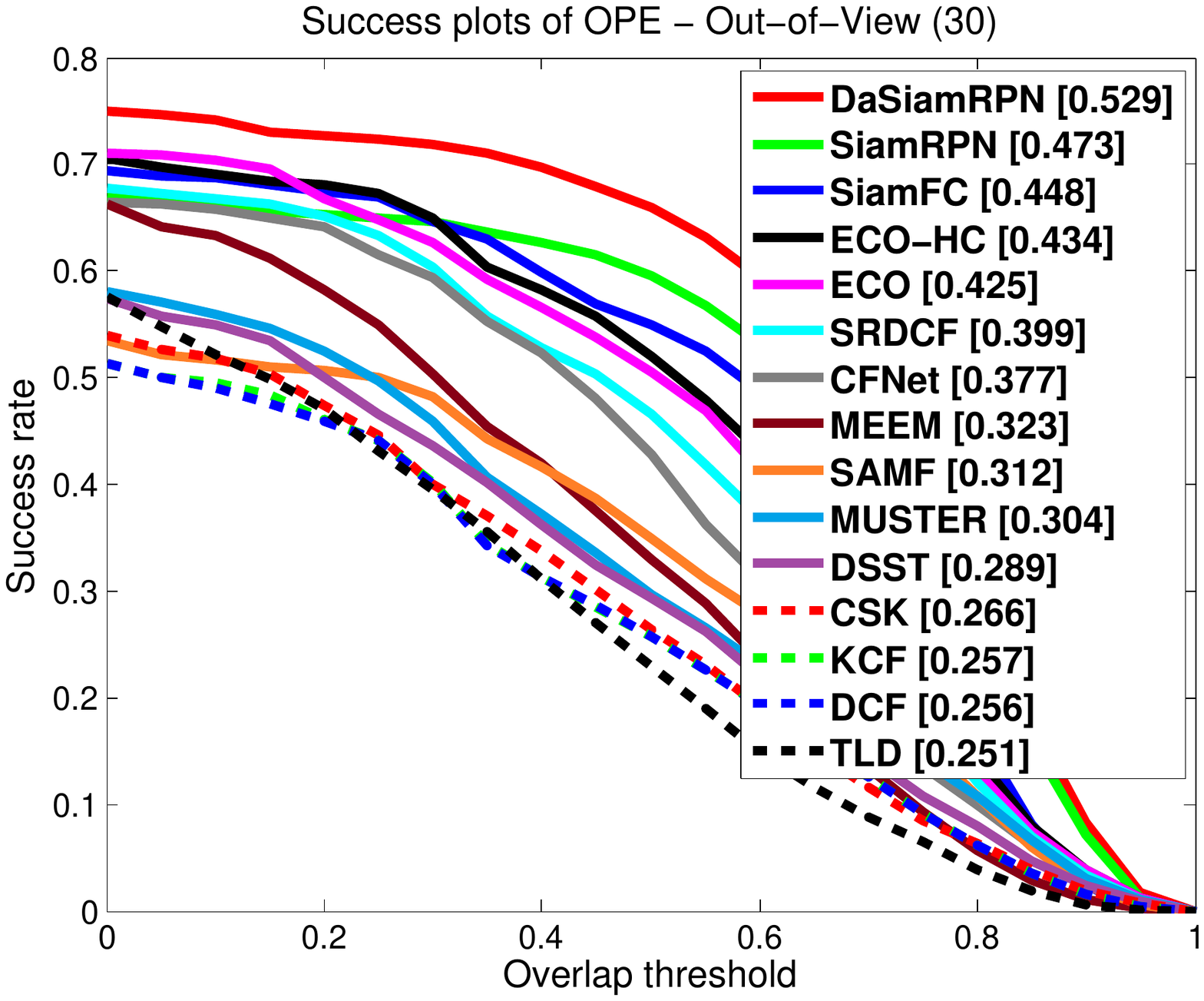}
\end{minipage}%
\begin{minipage}[c]{3cm}
\includegraphics[width=3cm]{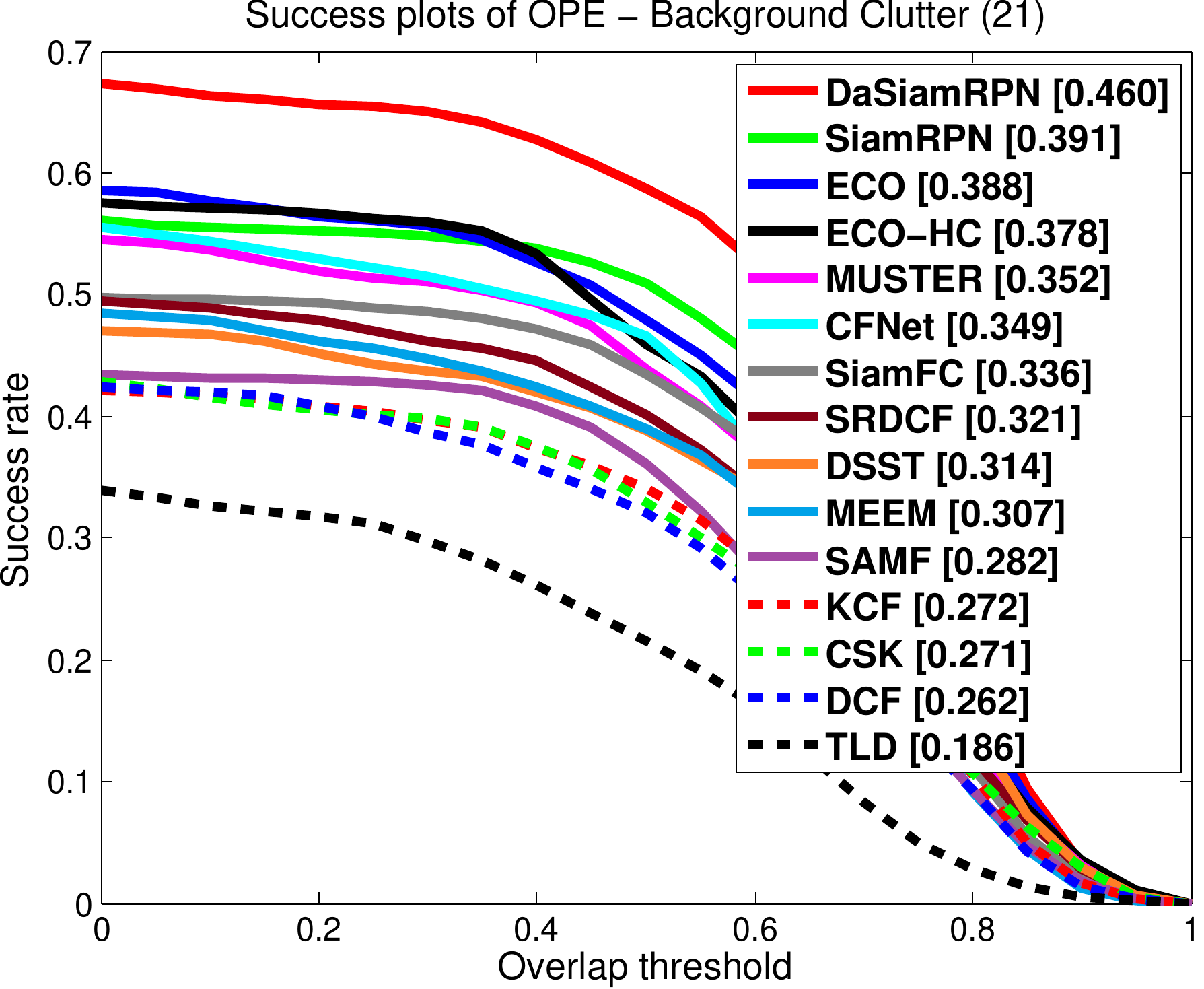}
\end{minipage}%
\begin{minipage}[c]{3cm}
\includegraphics[width=3cm]{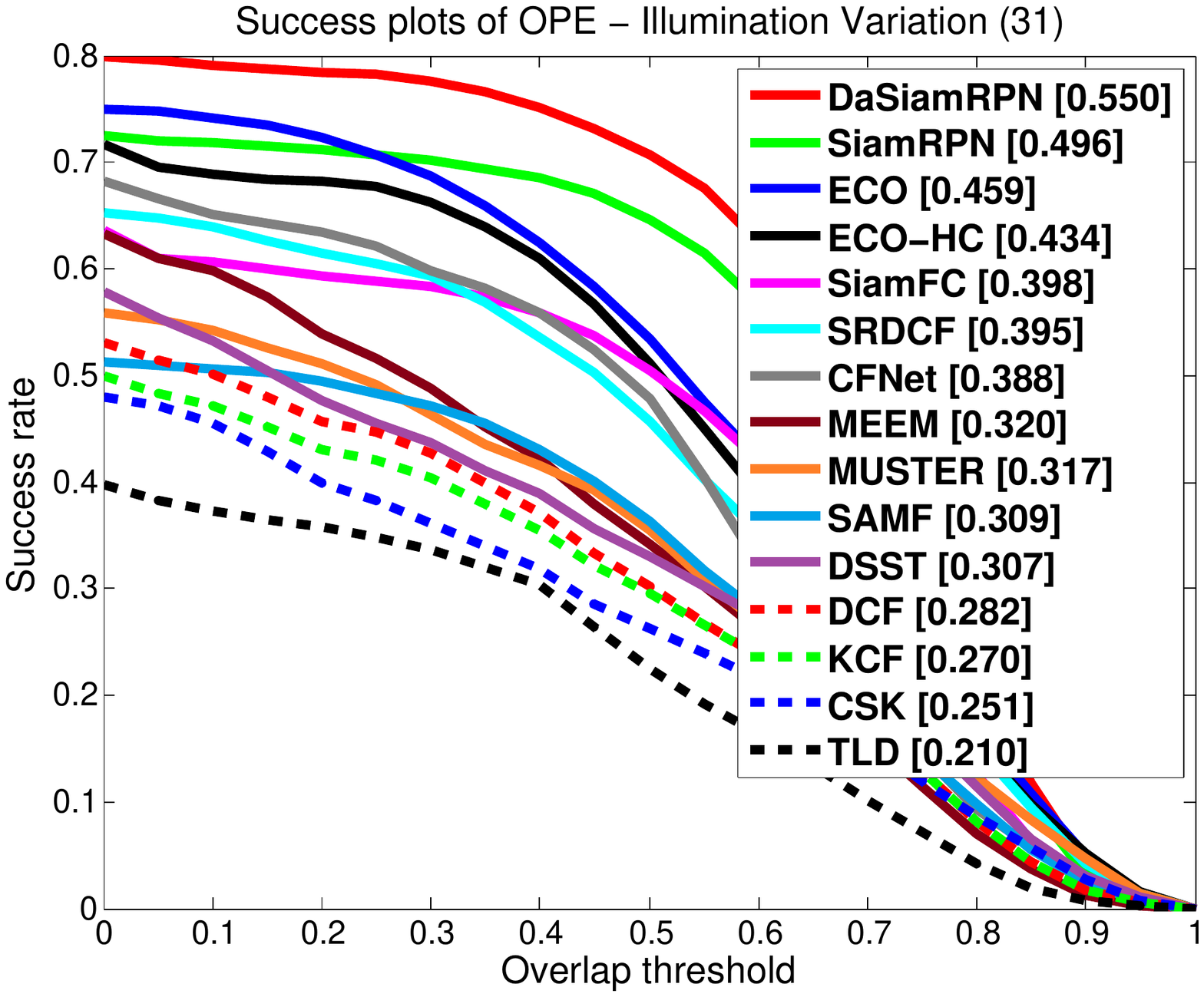}
\end{minipage}%
\begin{minipage}[c]{3cm}
\includegraphics[width=3cm]{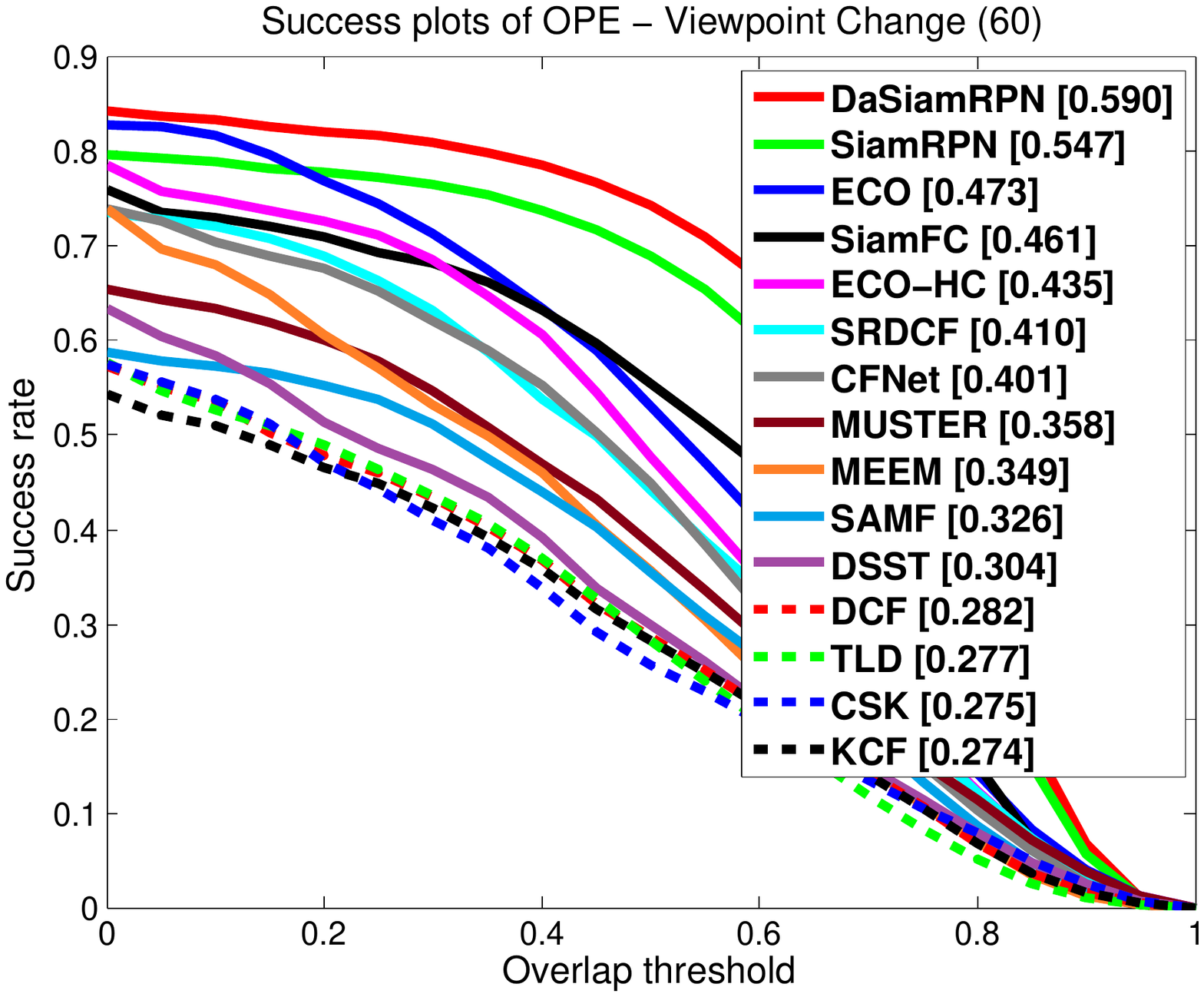}
\end{minipage}%

\begin{minipage}[c]{3cm}
\includegraphics[width=3cm]{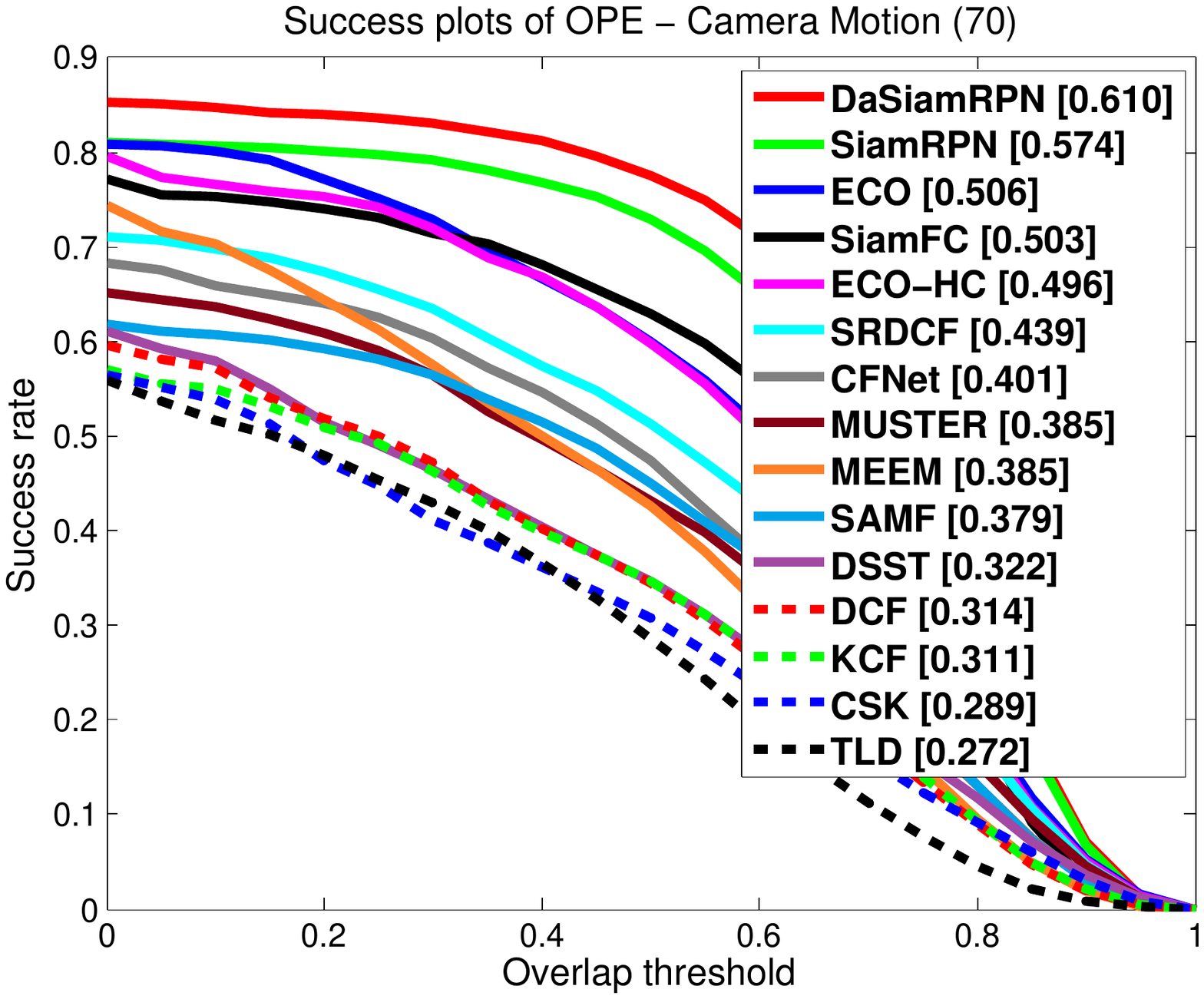}
\end{minipage}%
\begin{minipage}[c]{3cm}
\includegraphics[width=3cm]{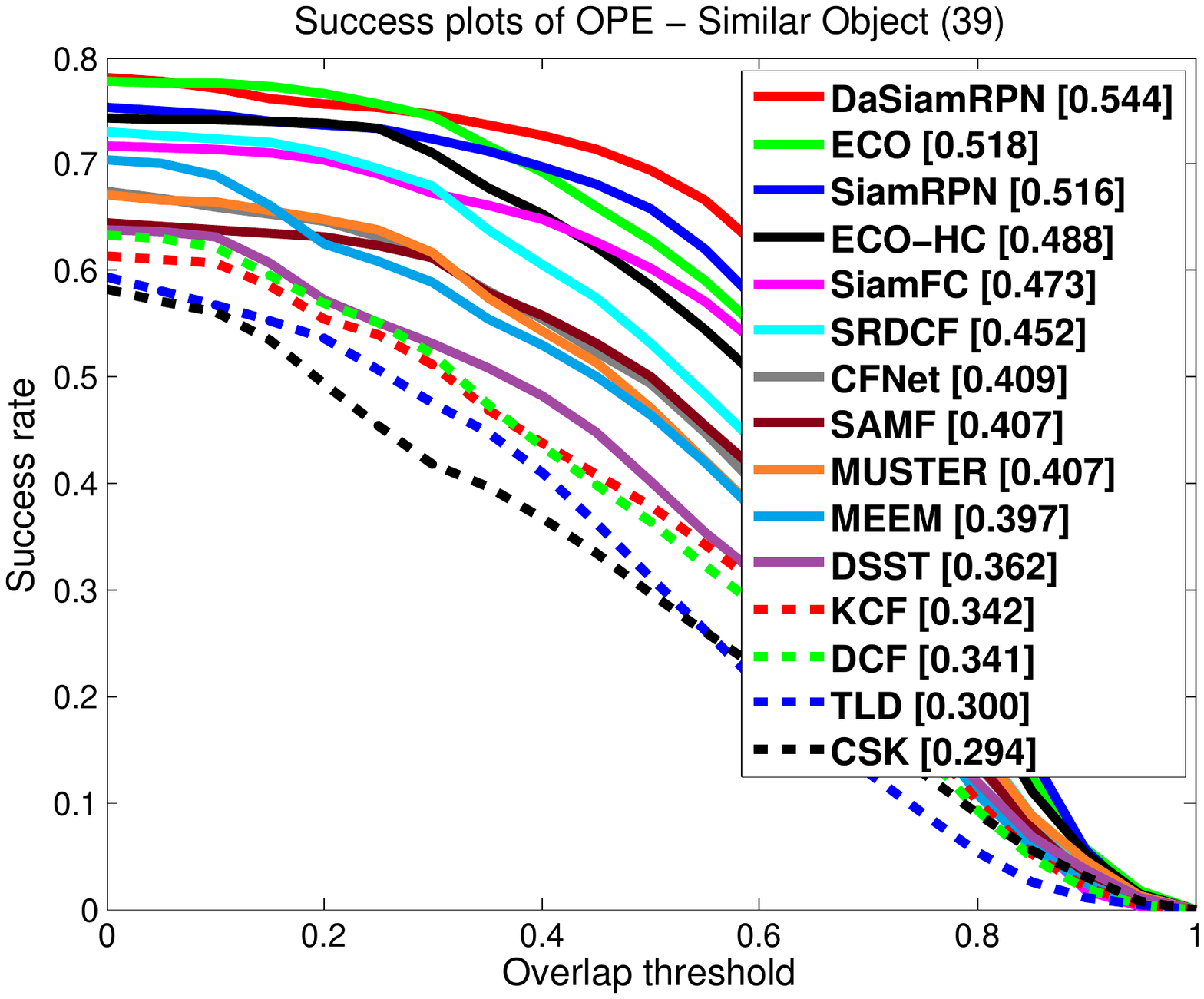}
\end{minipage}%
\begin{minipage}[c]{3cm}
\includegraphics[width=3cm]{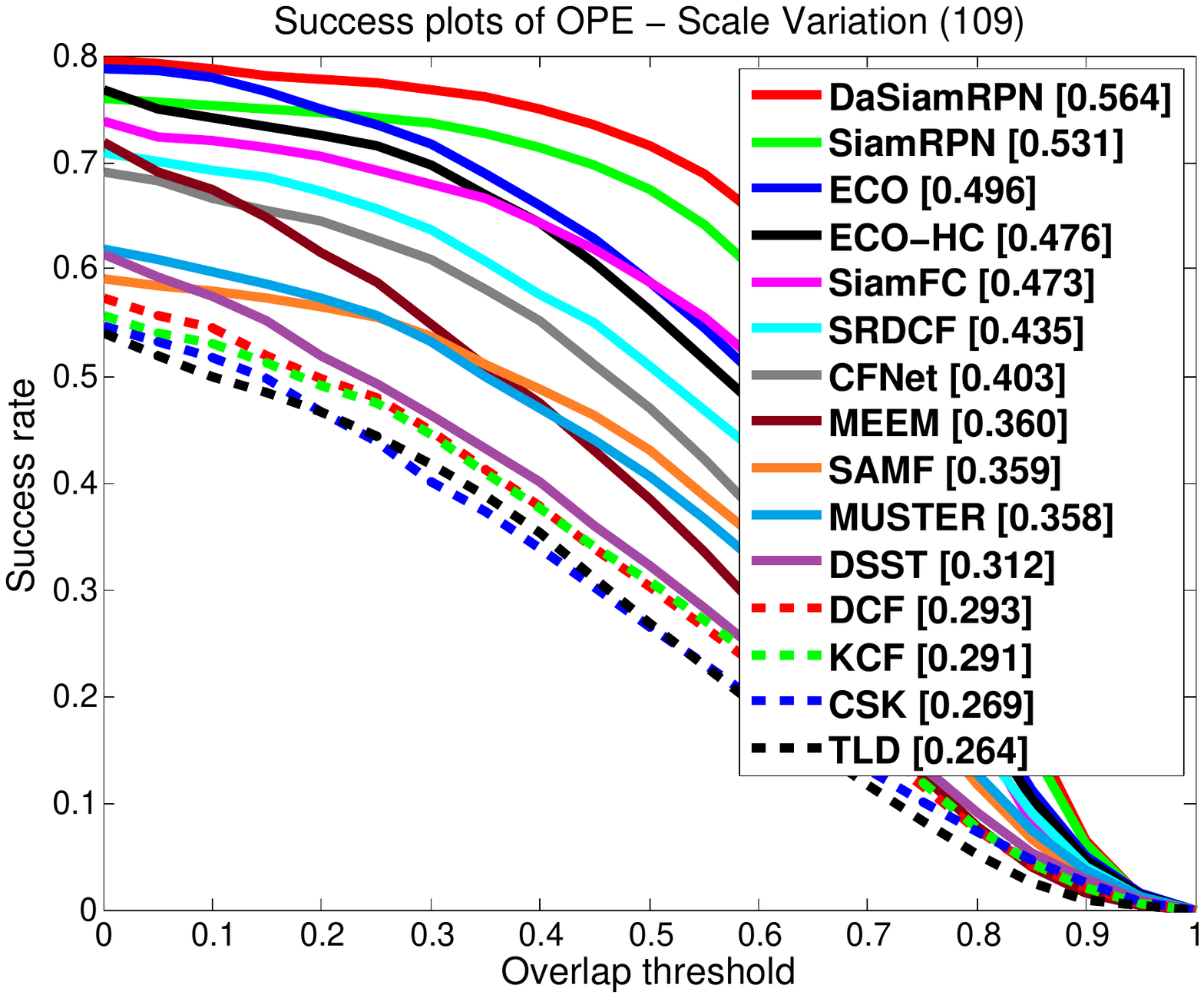}
\end{minipage}%
\begin{minipage}[c]{3cm}
\includegraphics[width=3cm]{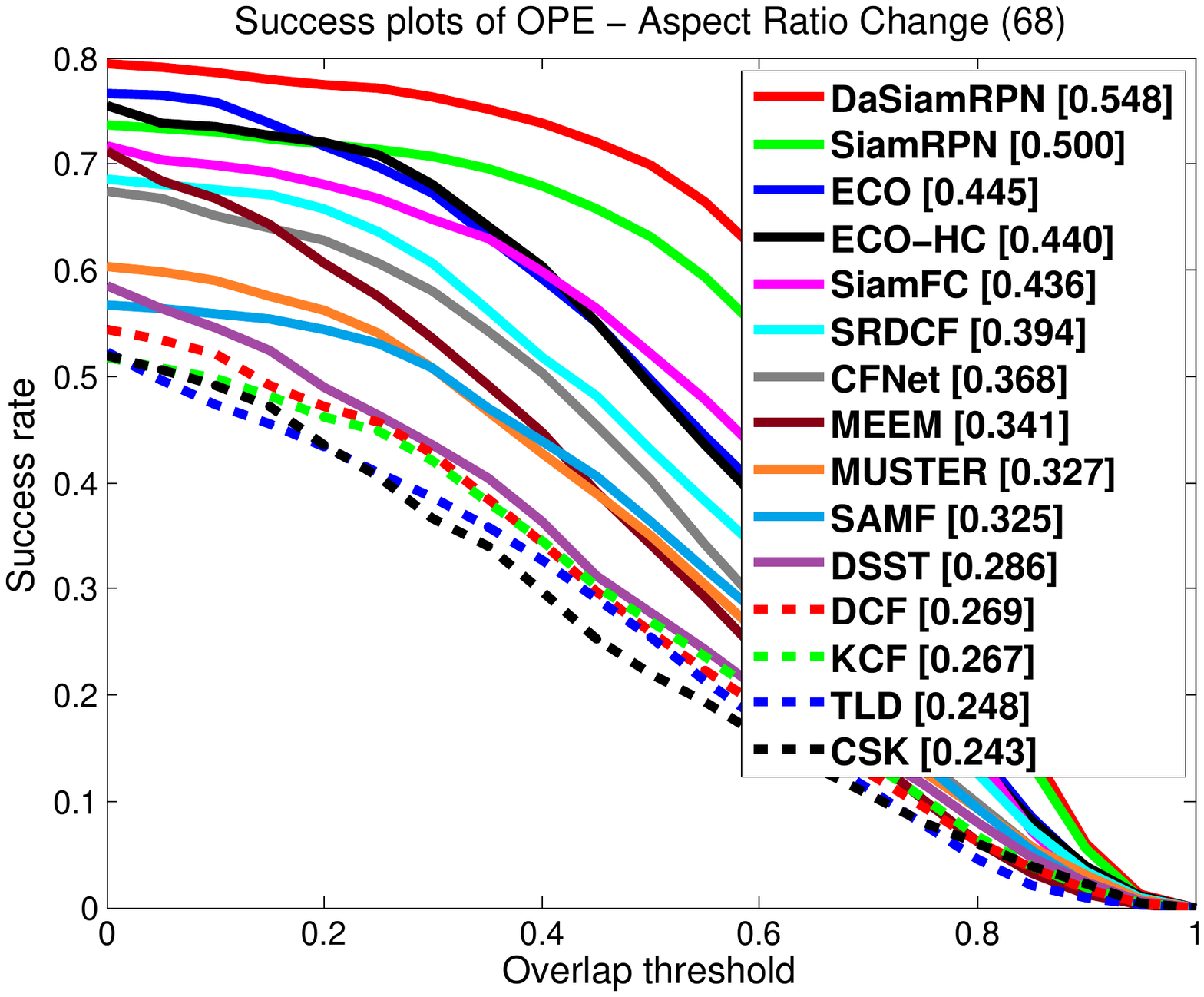}
\end{minipage}%

\begin{minipage}[c]{3cm}
\includegraphics[width=3cm]{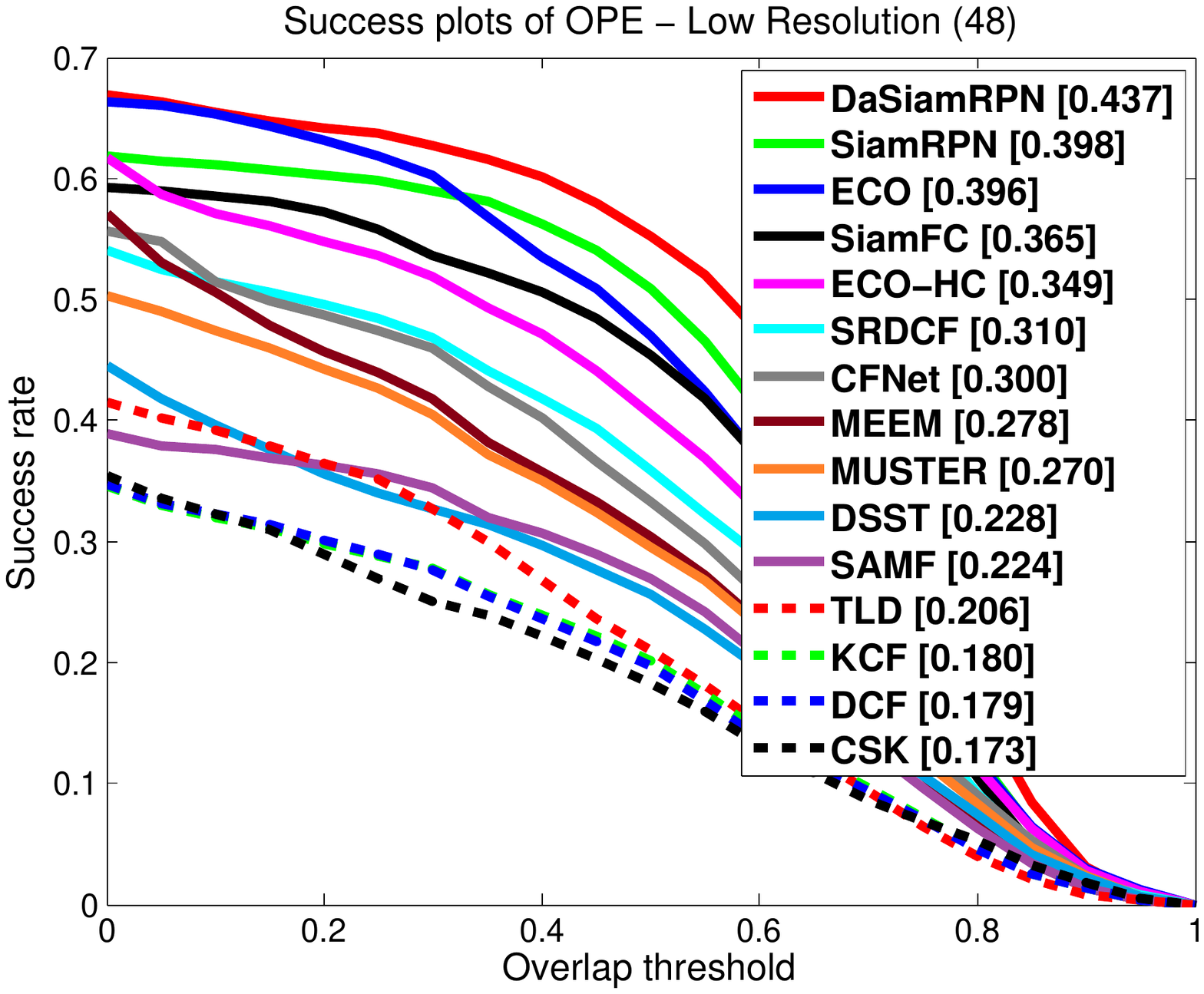}
\end{minipage}%
\begin{minipage}[c]{3cm}
\includegraphics[width=3cm]{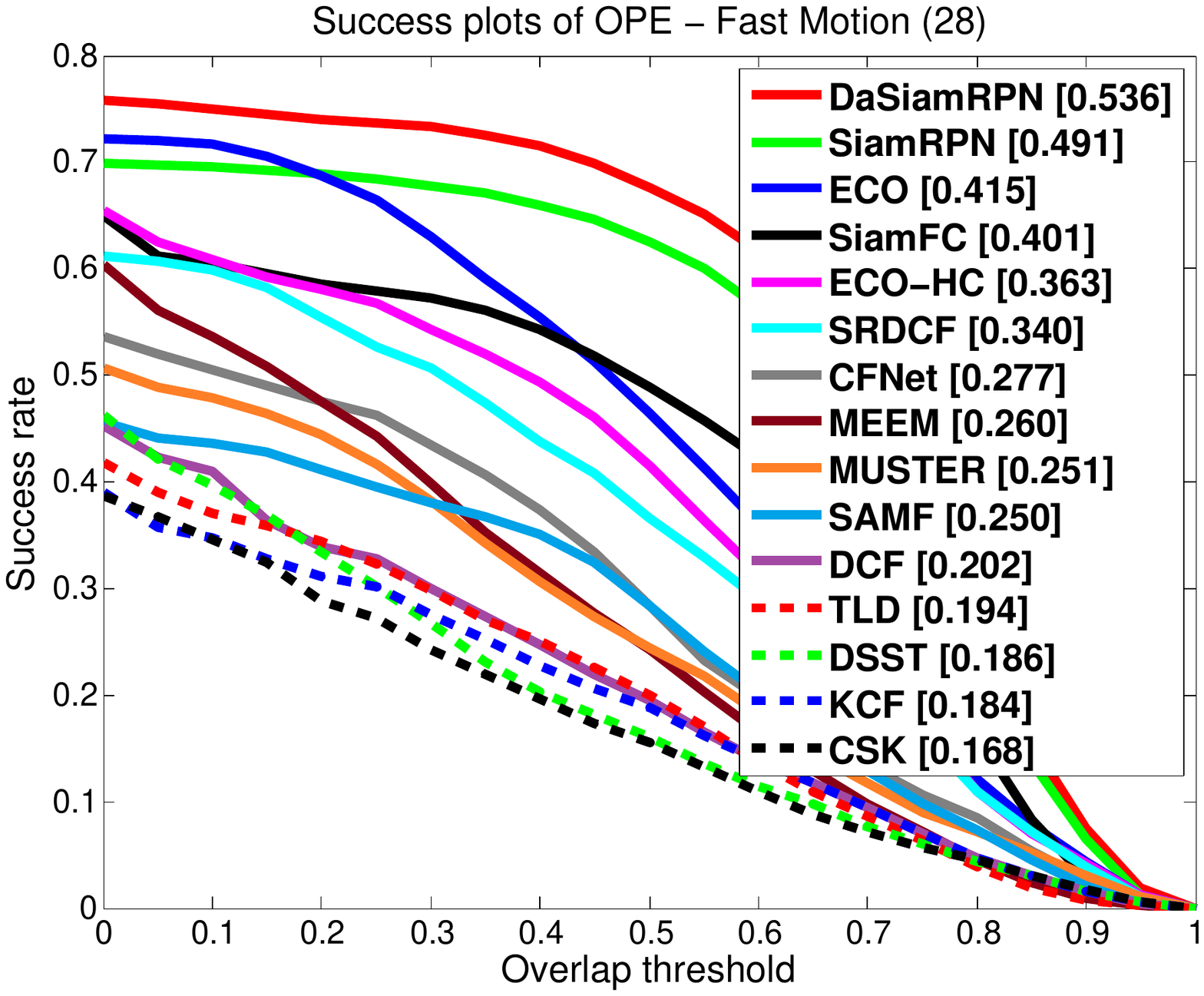}
\end{minipage}%
\begin{minipage}[c]{3cm}
\includegraphics[width=3cm]{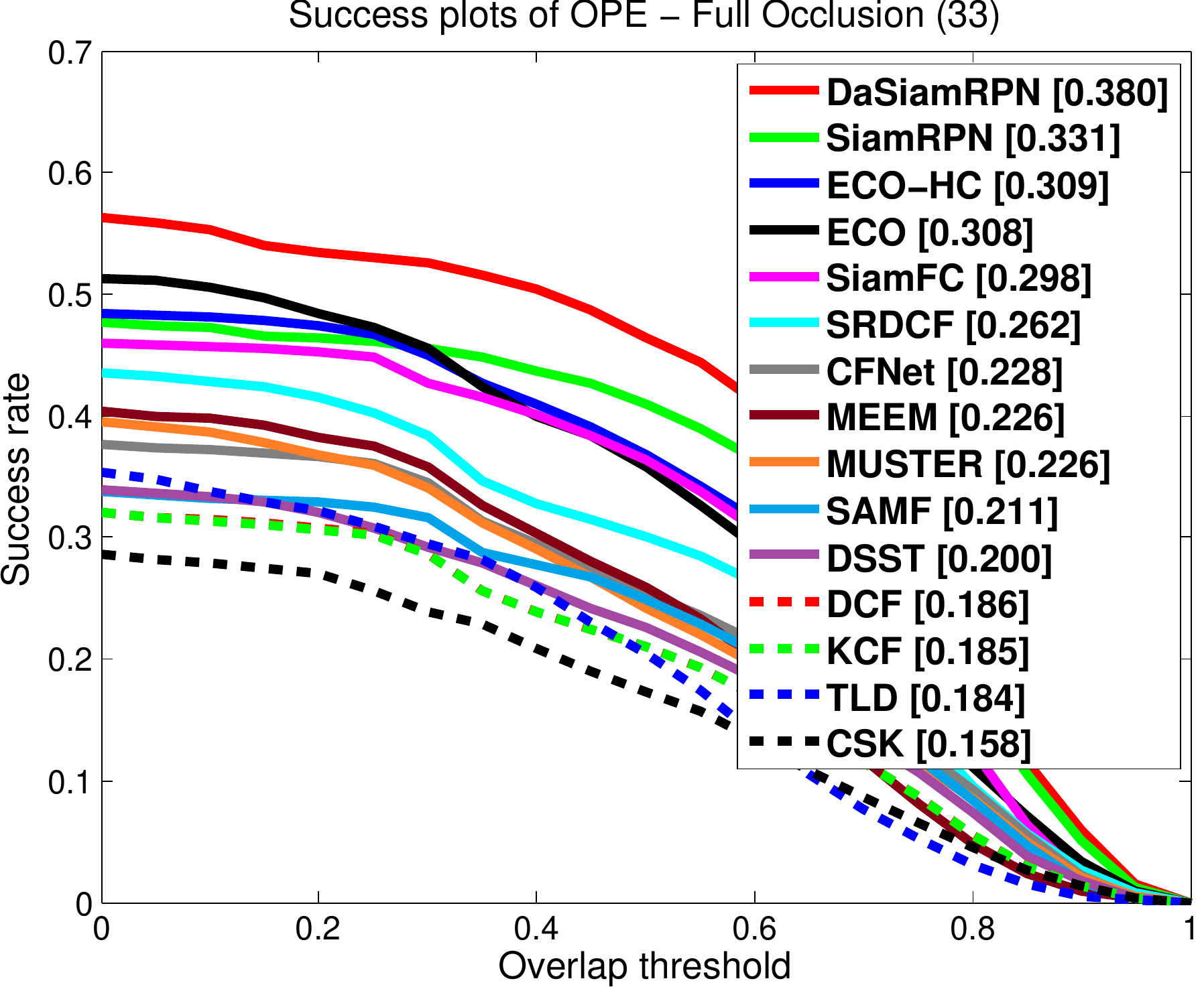}
\end{minipage}%
\begin{minipage}[c]{3cm}
\includegraphics[width=3cm]{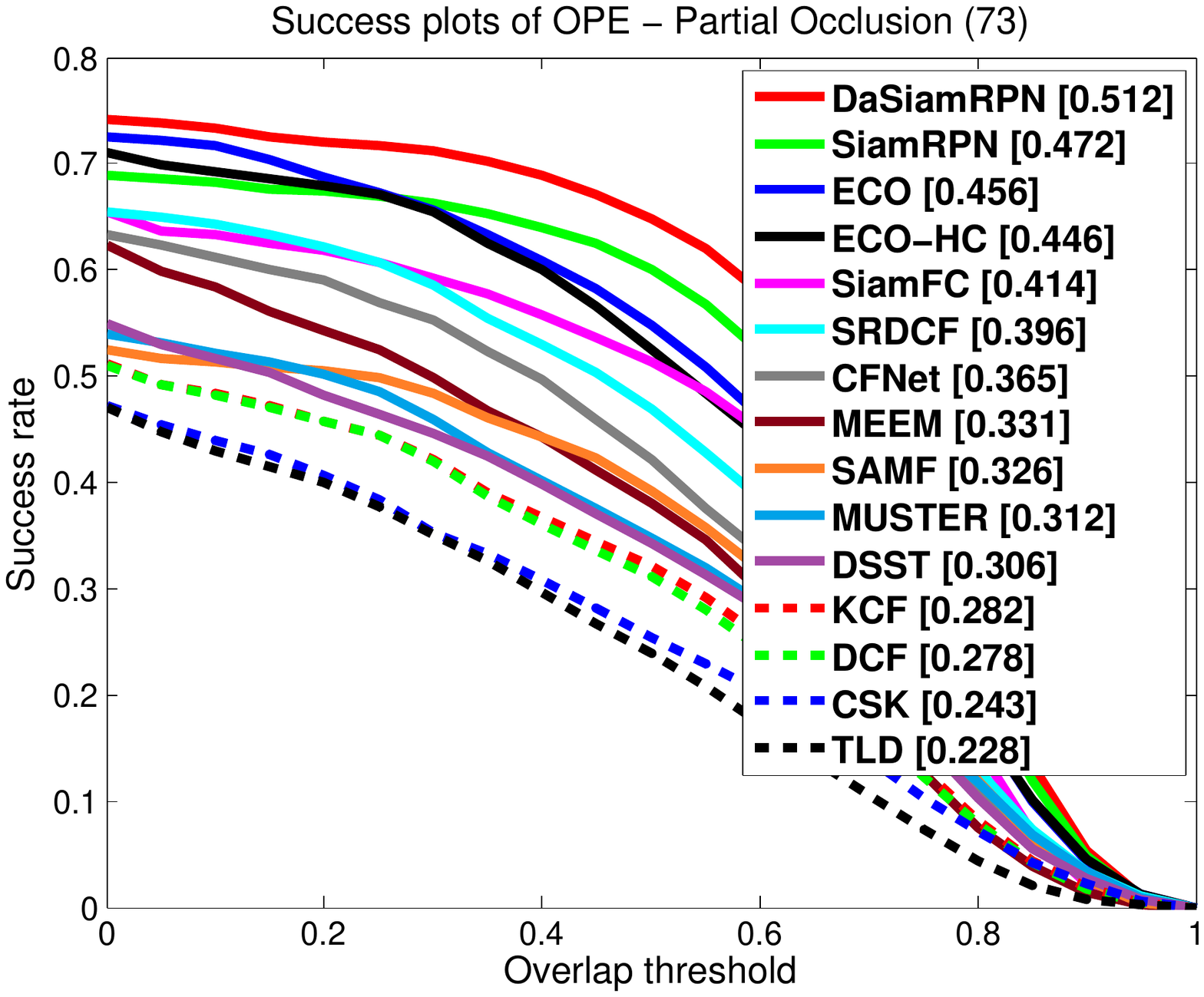}
\end{minipage}%
\caption{Success plots with attributes on UAV123. Best viewed on color display.}
\label{UAV123_attributes}
%\vspace{-0.5cm}
\end{figure*}

Additional results on the UAV123 with 12 different attributes are reported in this section. As shown in Fig.~\ref{UAV123_attributes}, the performance is ranked by area under curve (AUC) of success plot.

\subsection{Detailed results on OTB}
\begin{figure*}[htbp]
\captionsetup{font={small}}
\centering
\subfloat[IPR on OTB-2013]{\includegraphics[width=0.3\linewidth]{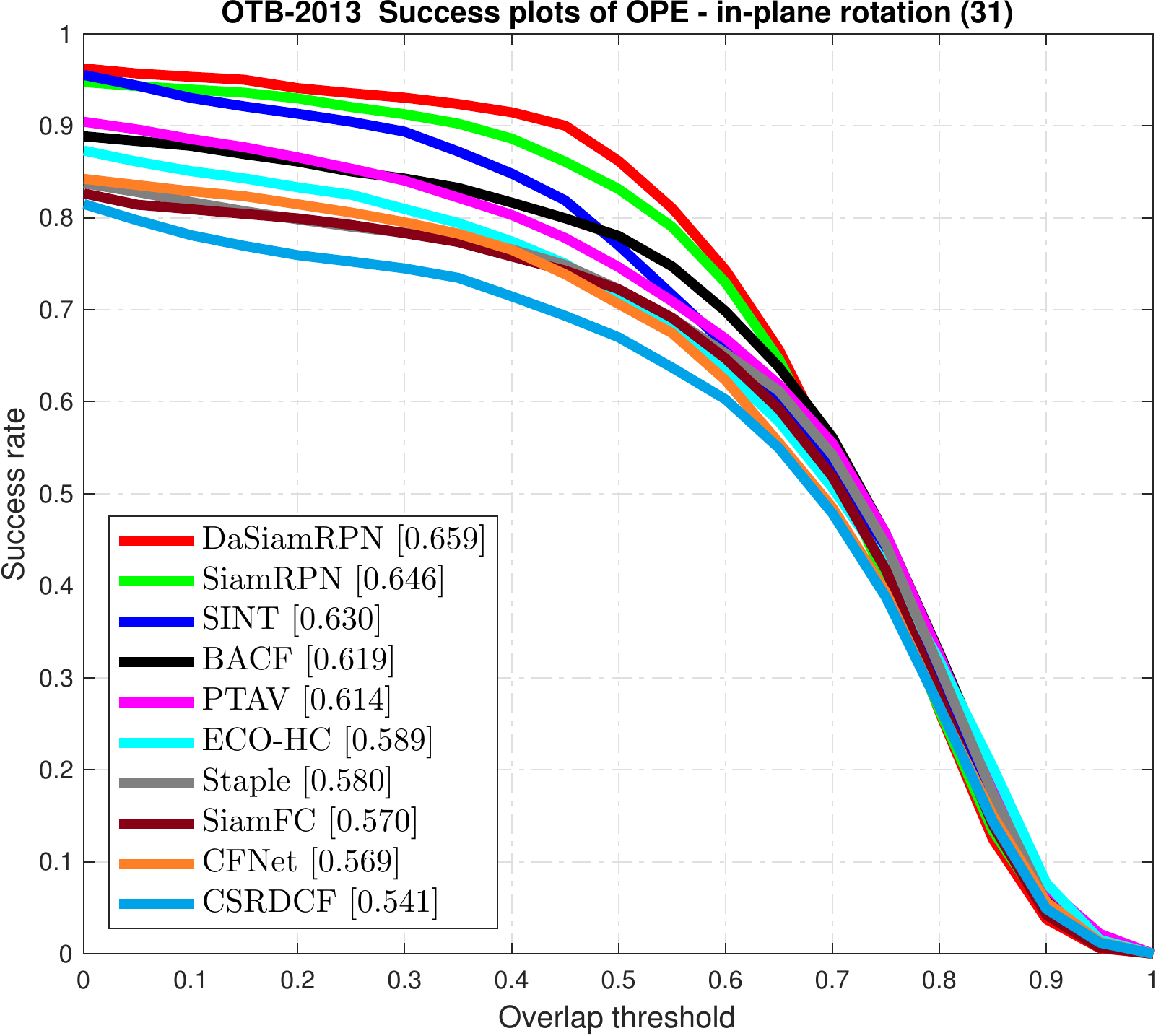}
\label{IPR on OTB2013}}
\subfloat[OPR on OTB-2013]{\includegraphics[width=0.3\linewidth]{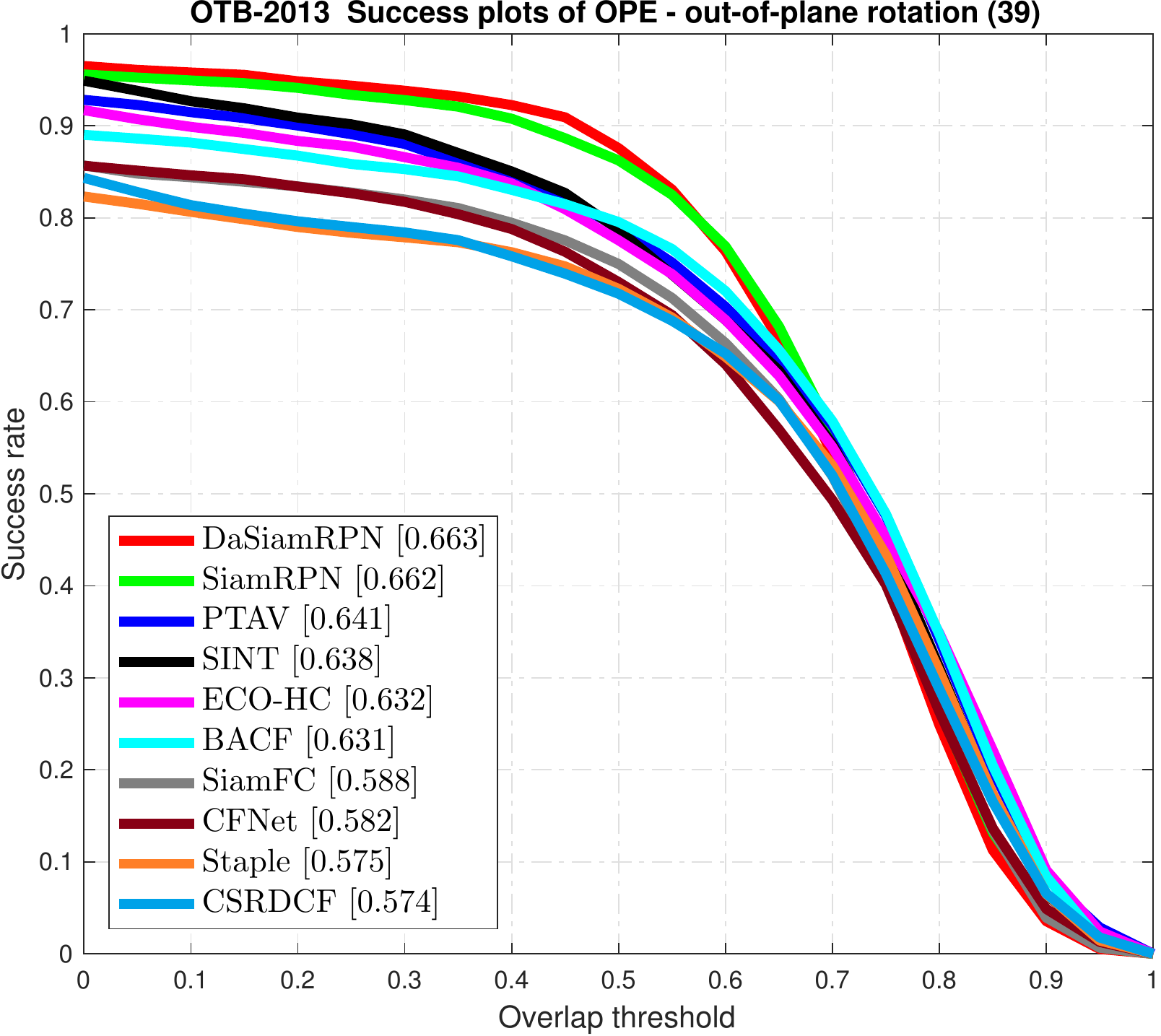}
\label{OPR on OTB2013}}
\subfloat[SV on OTB-2013]{\includegraphics[width=0.3\linewidth]{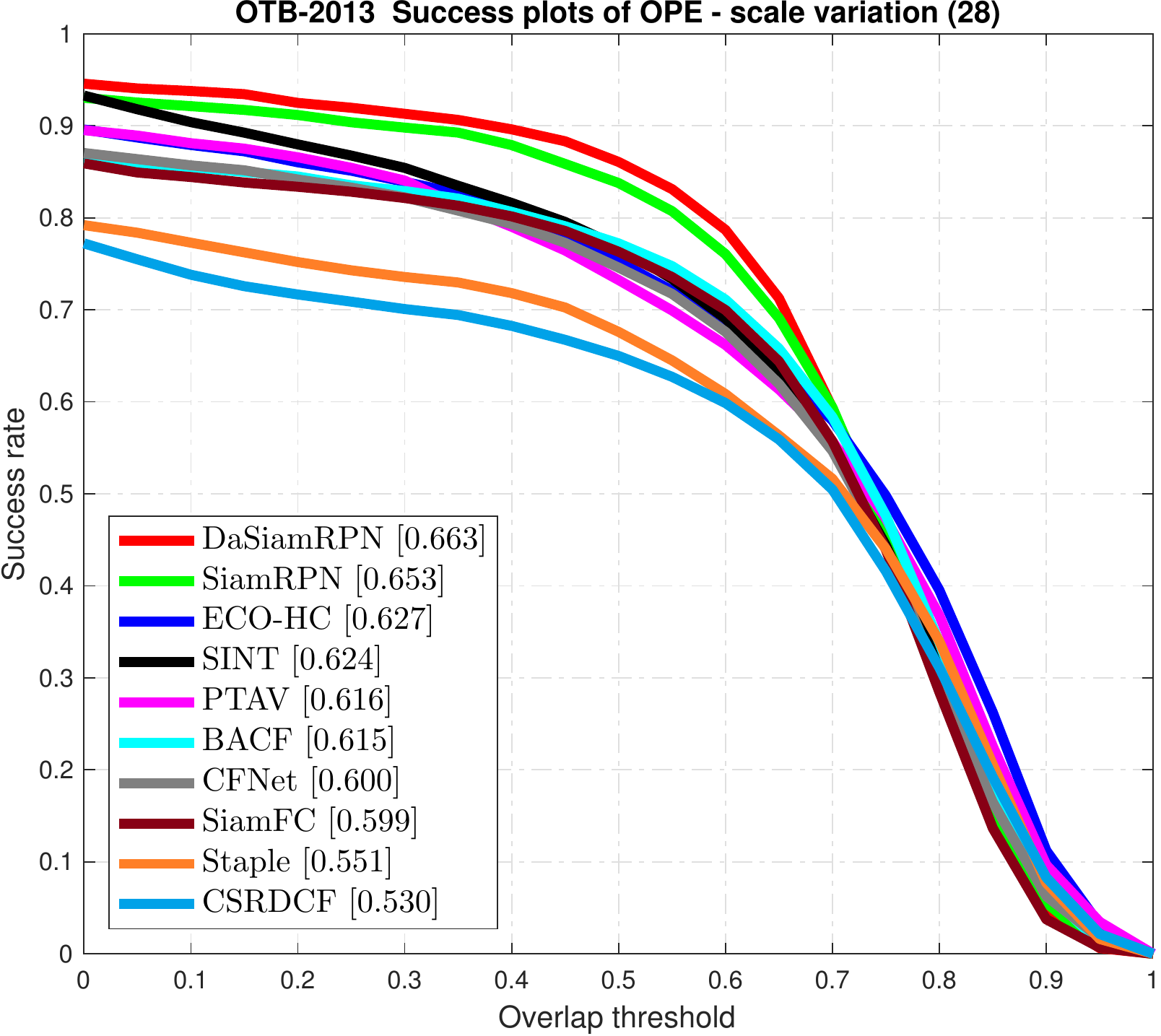}
\label{SV on OTB2013}}
\hfil
\subfloat[OV on OTB-2013]{\includegraphics[width=0.3\linewidth]{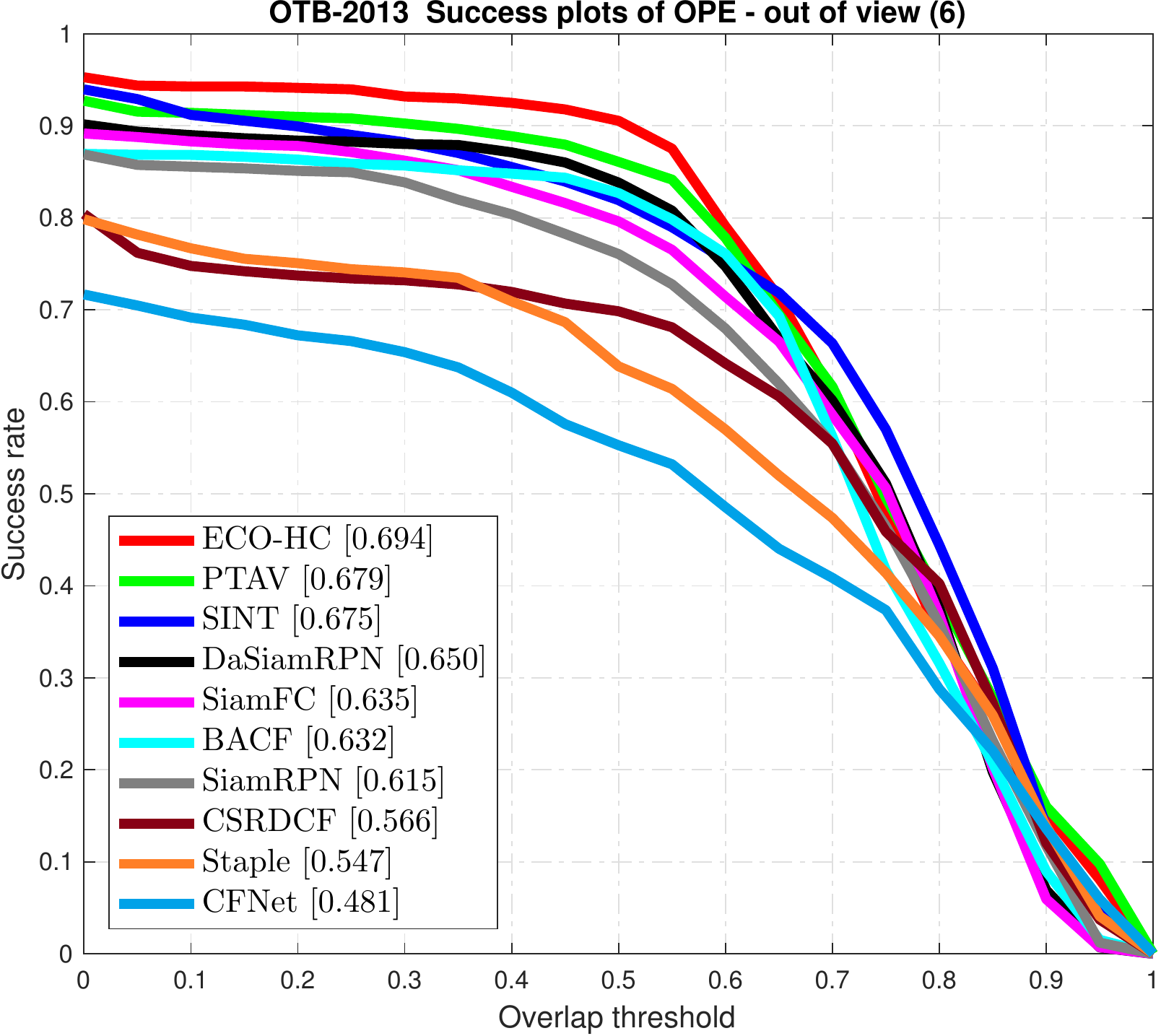}
\label{OV on OTB2013}}
\subfloat[OCC on OTB-2013]{\includegraphics[width=0.3\linewidth]{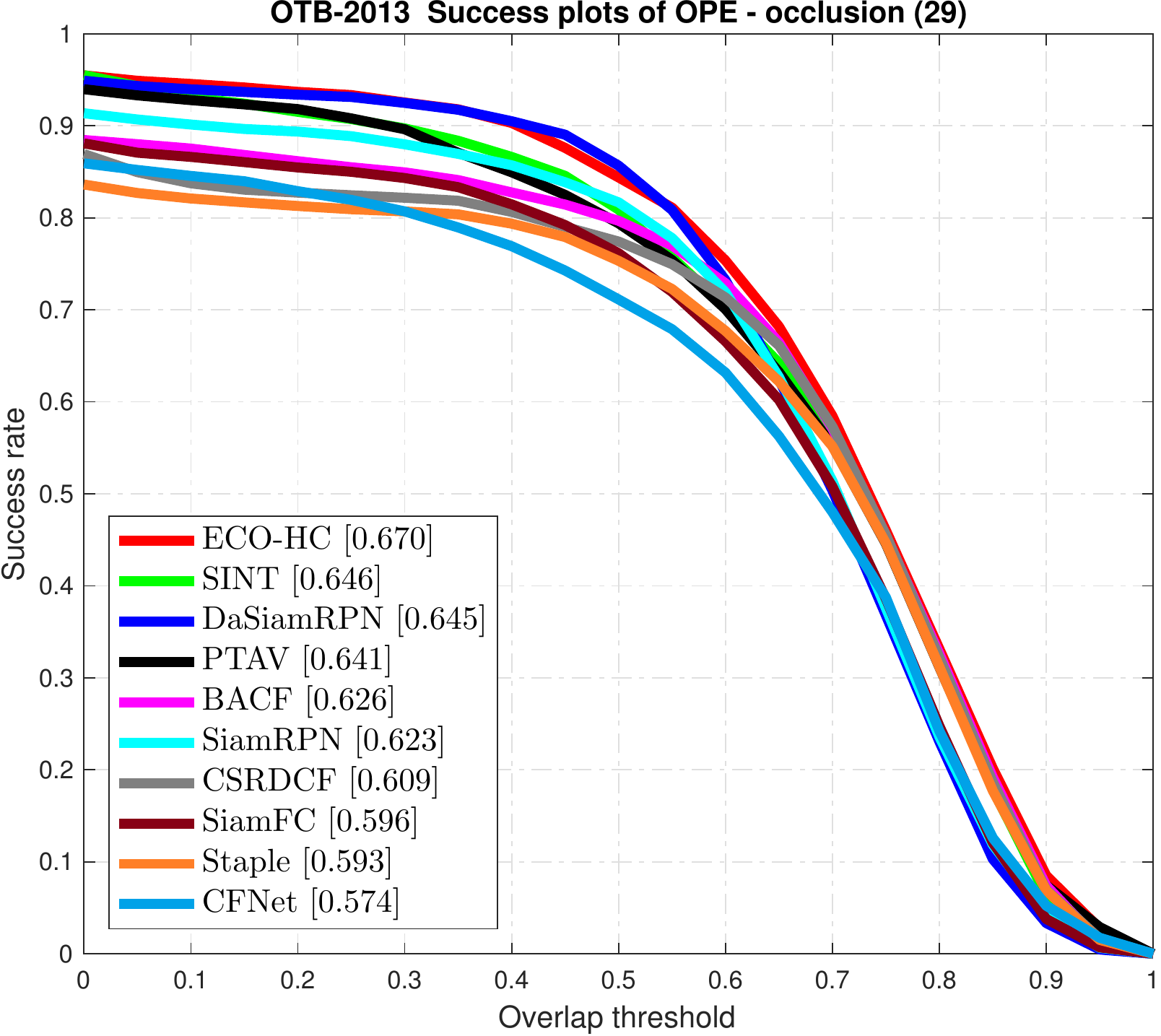}
\label{OCC on OTB2013}}
\subfloat[BC on OTB-2013]{\includegraphics[width=0.3\linewidth]{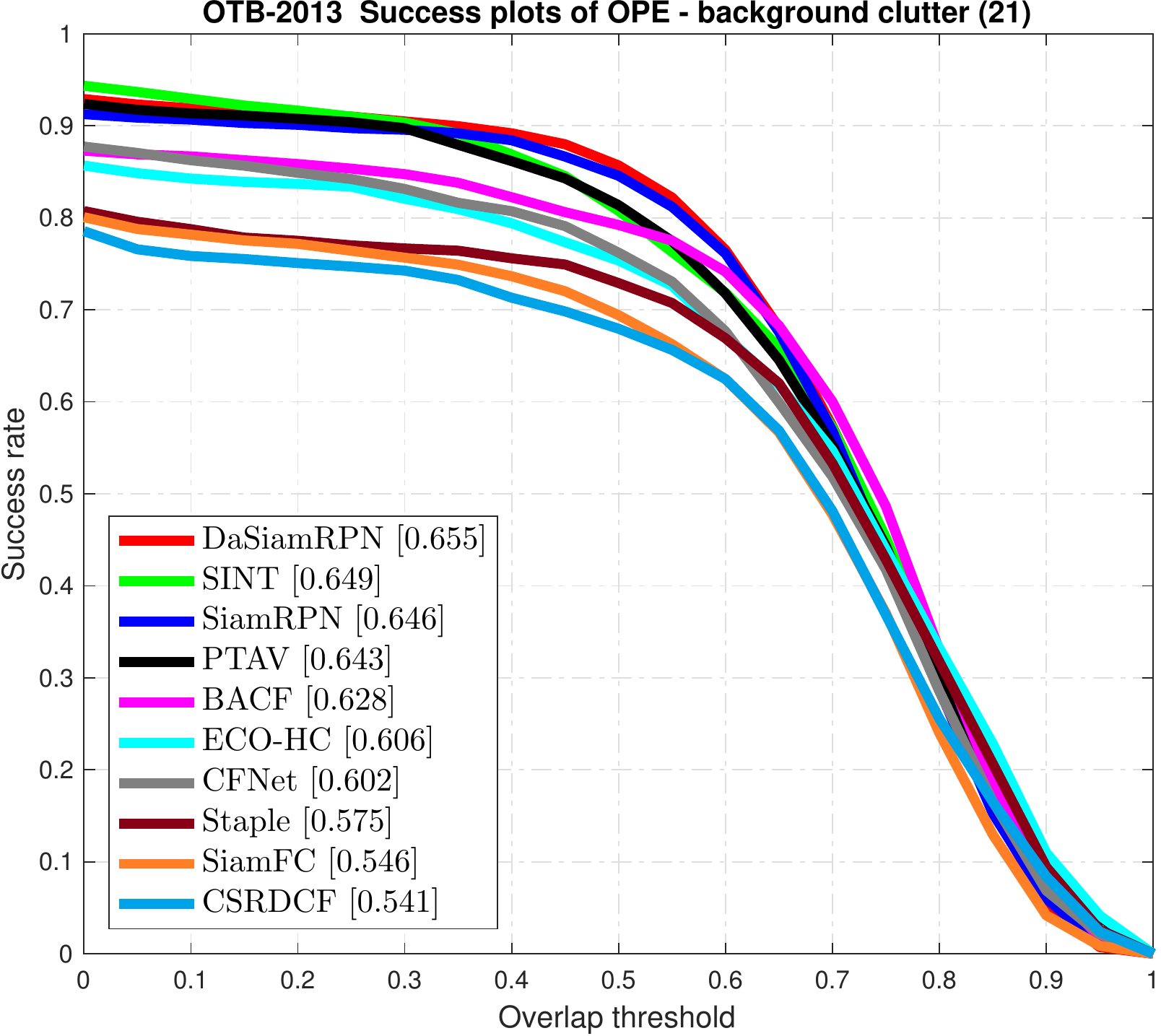}
\label{BC on OTB2013}}
\hfil
\subfloat[DEF on OTB-2013]{\includegraphics[width=0.3\linewidth]{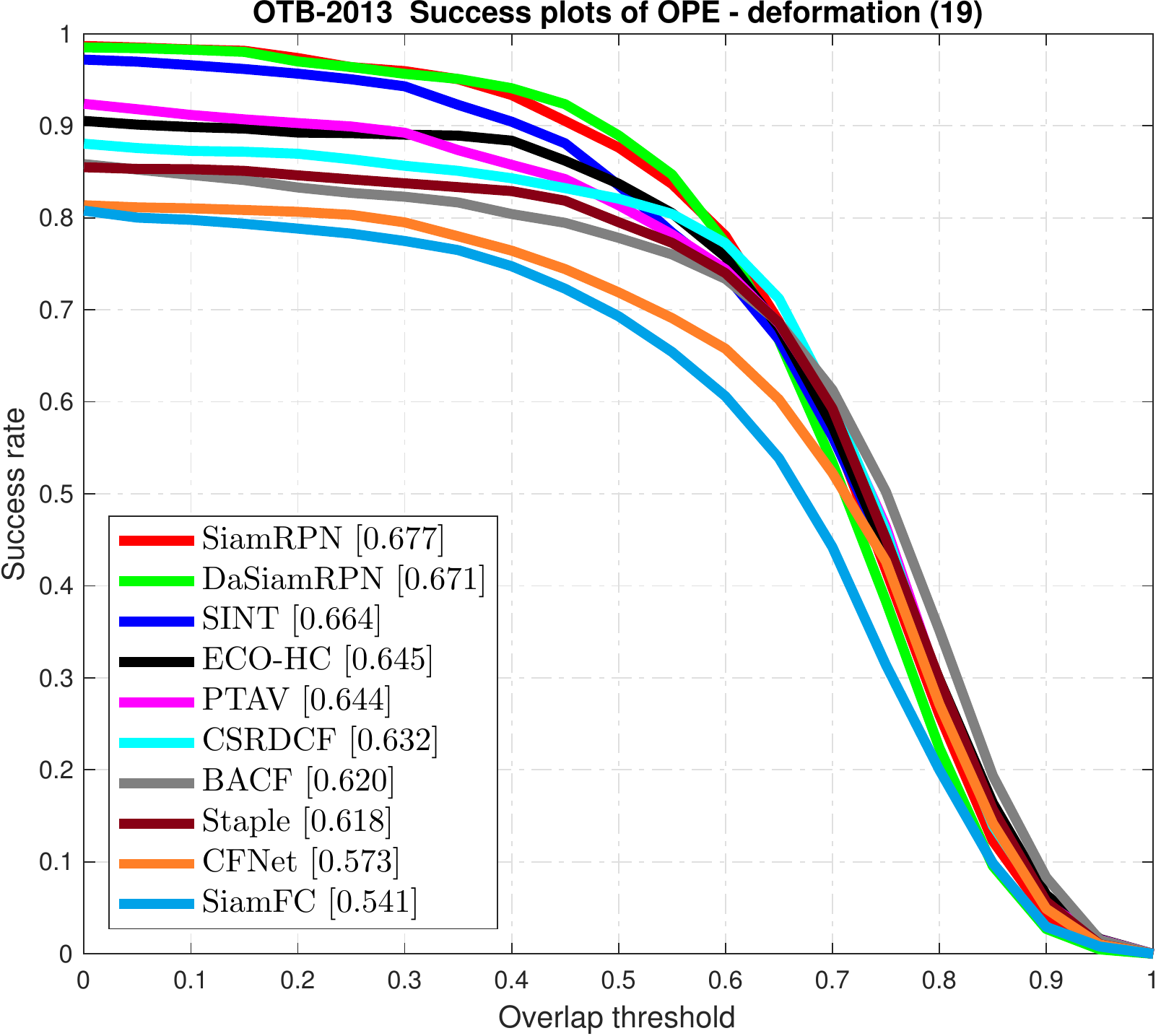}
\label{DEF on OTB2013}}
\subfloat[IV on OTB-2013]{\includegraphics[width=0.3\linewidth]{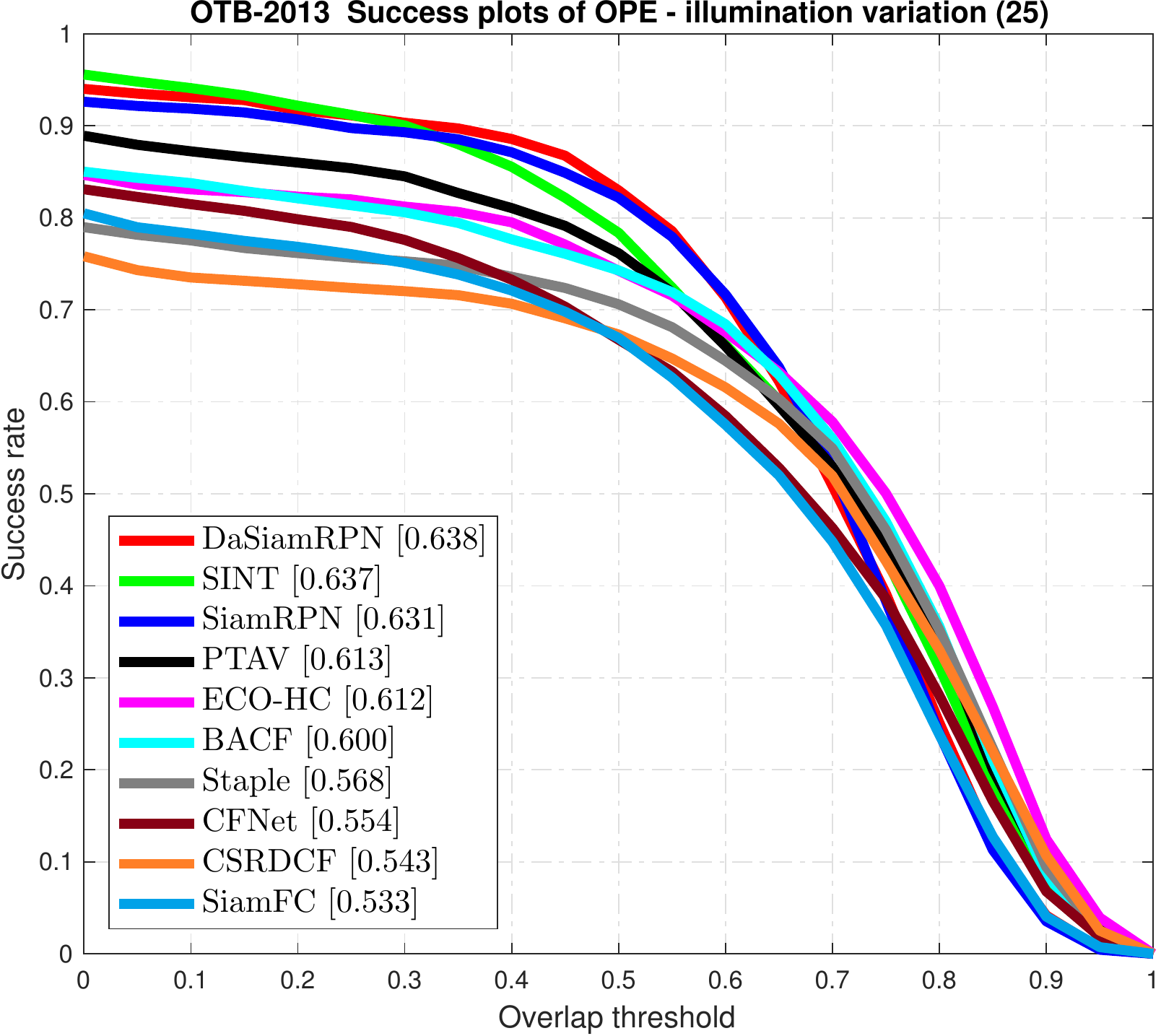}
\label{IV on OTB2013}}
\subfloat[LR on OTB-2013]{\includegraphics[width=0.3\linewidth]{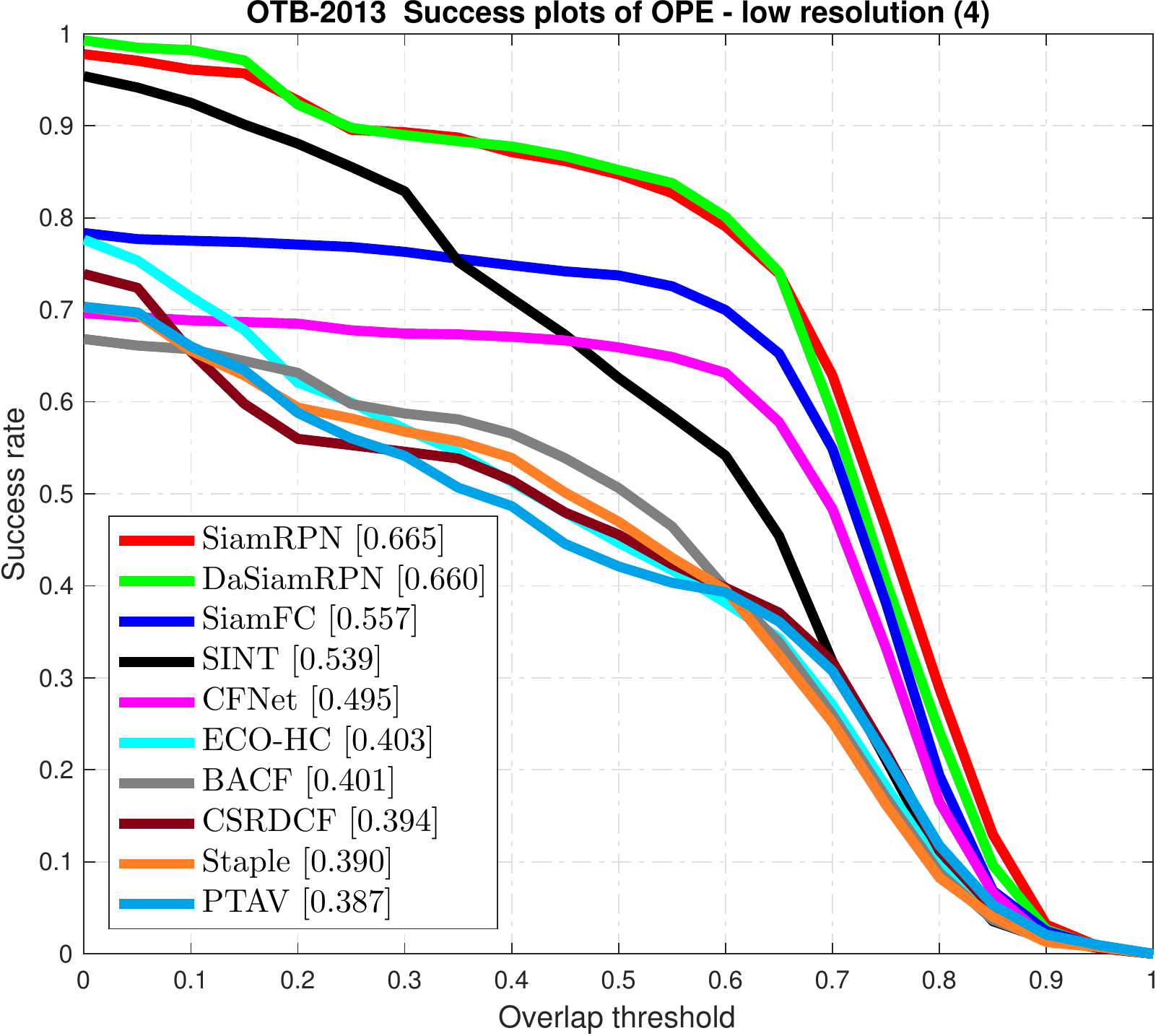}
\label{LR on OTB2013}}
\hfil
\subfloat[FM on OTB-2013]{\includegraphics[width=0.3\linewidth]{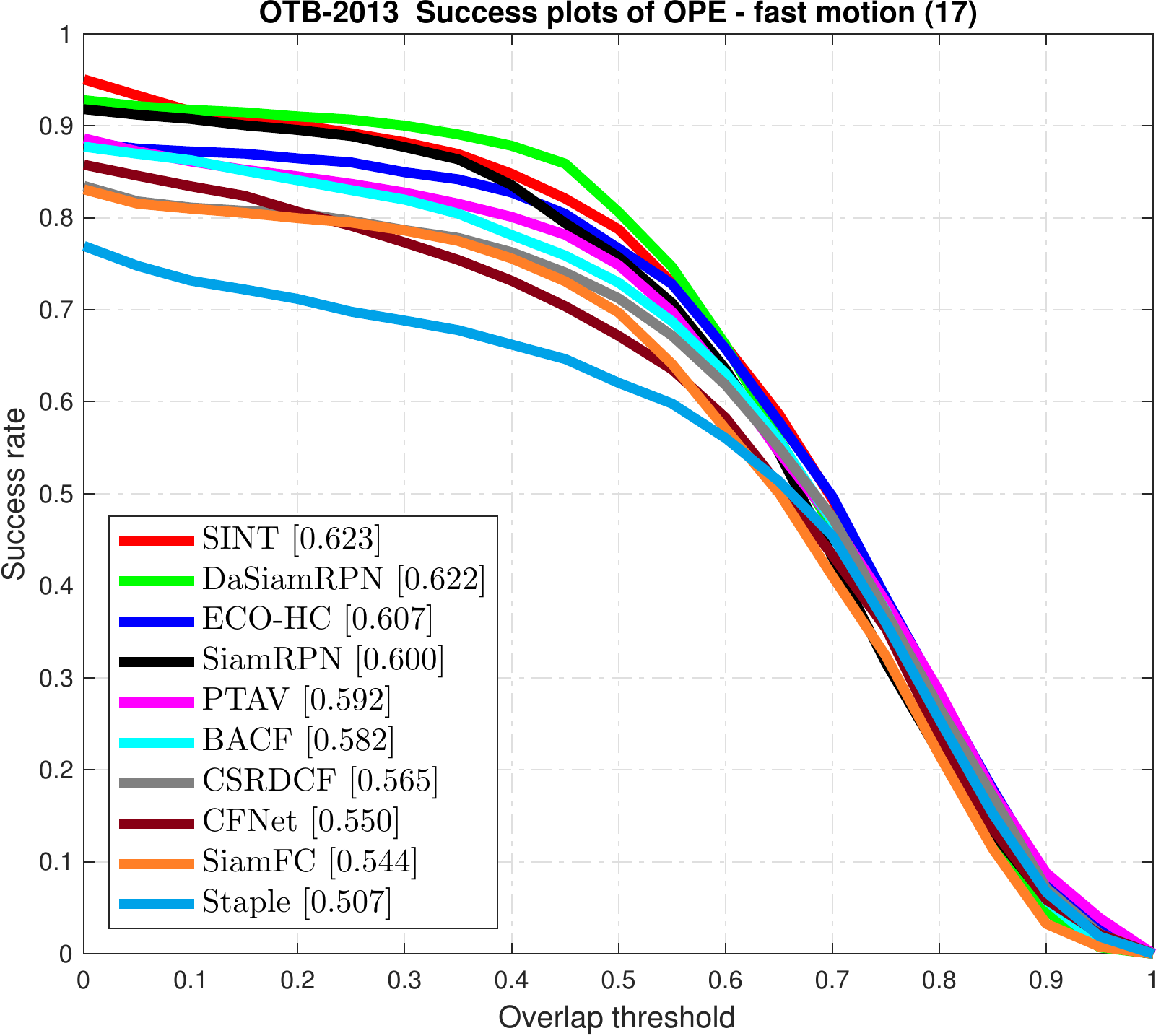}
\label{FM on OTB2013}}
\subfloat[MB on OTB-2013]{\includegraphics[width=0.3\linewidth]{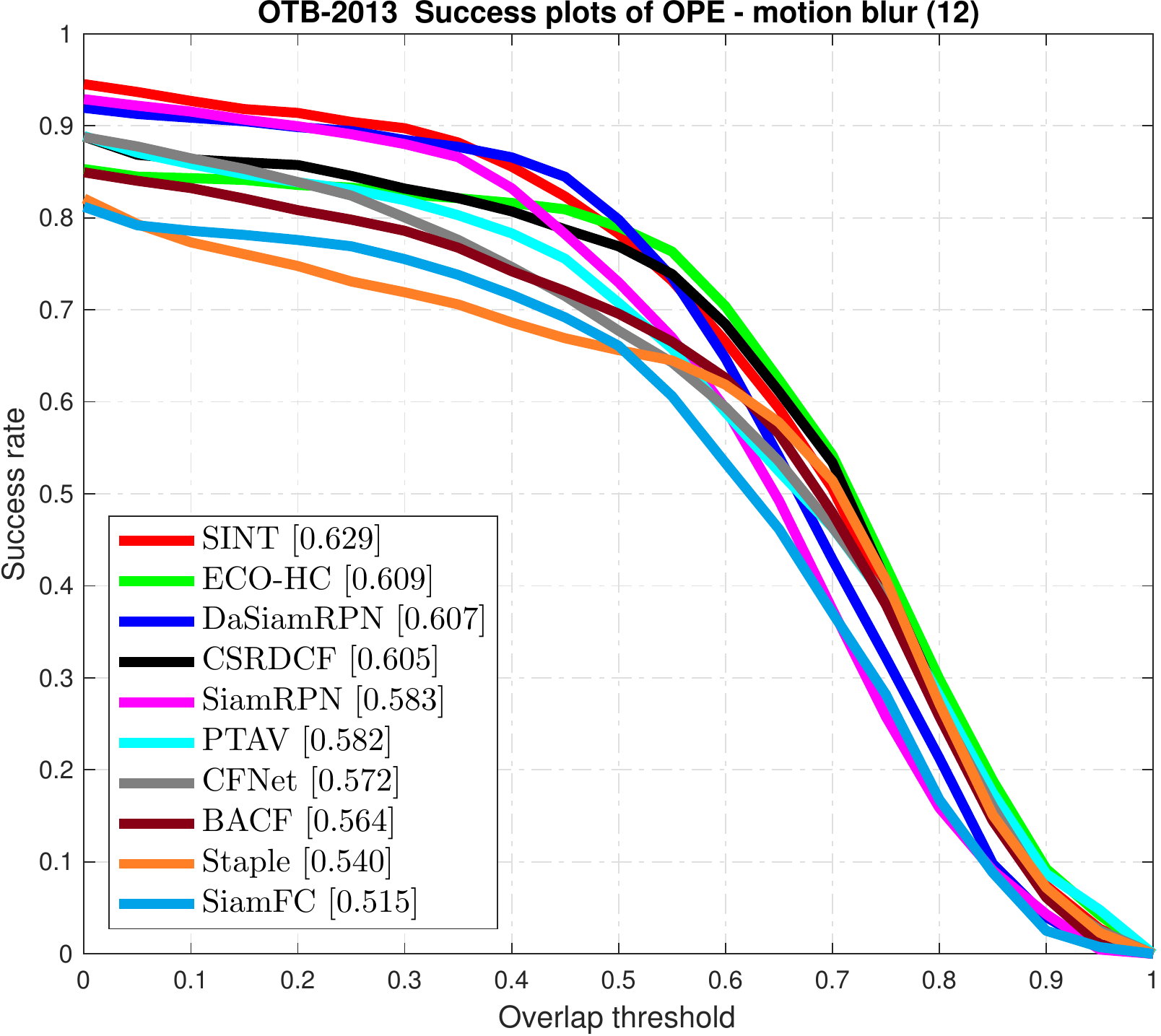}
\label{MB on OTB2013}}
\caption{The success plots on OTB-2013 for eleven challenge attributes: in-plain rotation, out-of-plane rotation, scale variation, out of view, occlusion, background clutter,  deformation, illumination variation, low resolution, fast motion and motion blur.}
\label{fig:result1OTB}
\end{figure*}

\begin{figure*}[htbp]
\captionsetup{font={small}}
\centering
\subfloat[IPR on OTB-2015]{\includegraphics[width=0.3\linewidth]{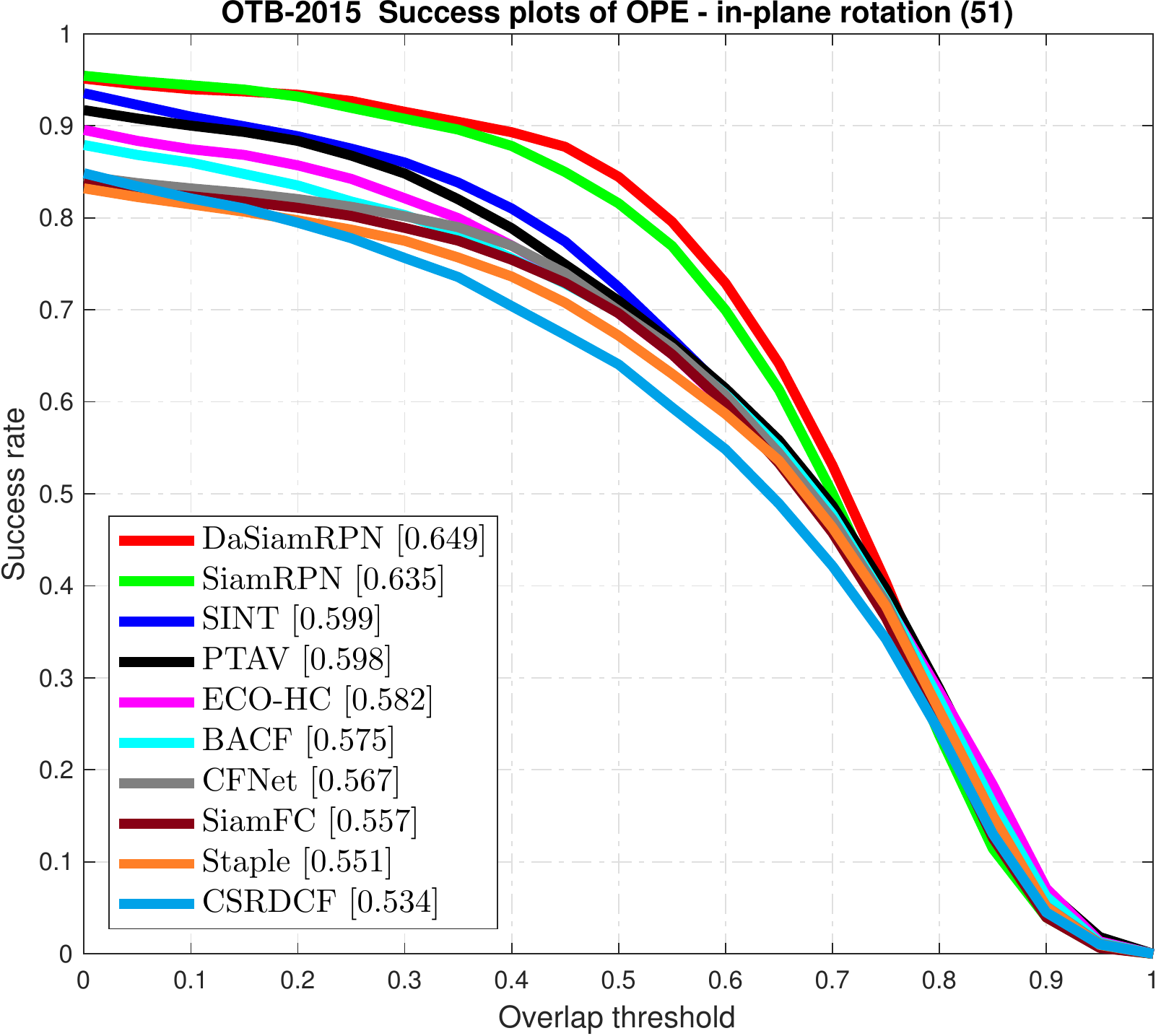}
\label{IPR on OTB2015}}
\subfloat[OPR on OTB-2015]{\includegraphics[width=0.3\linewidth]{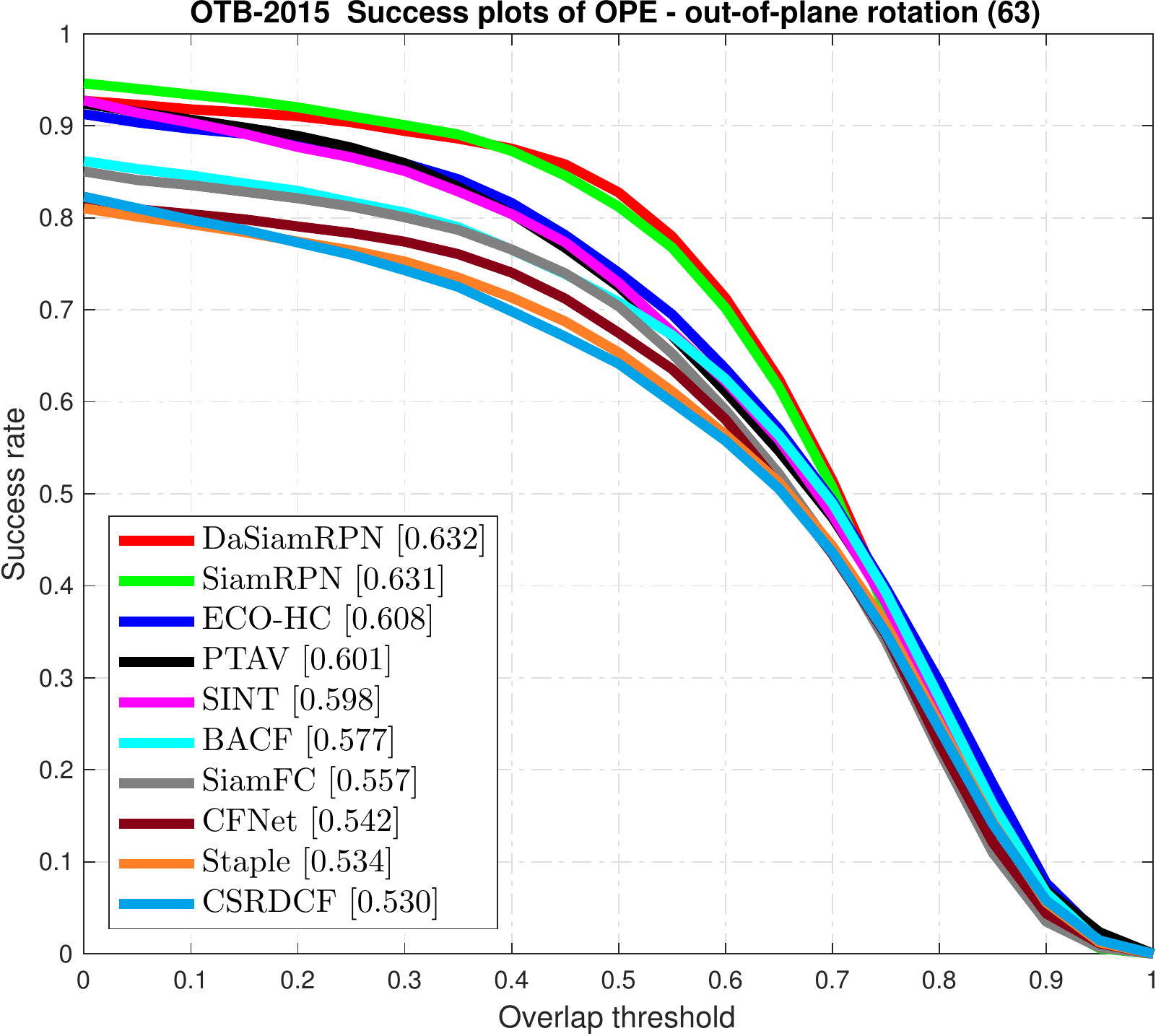}
\label{OPR on OTB2015}}
\subfloat[SV on OTB-2015]{\includegraphics[width=0.3\linewidth]{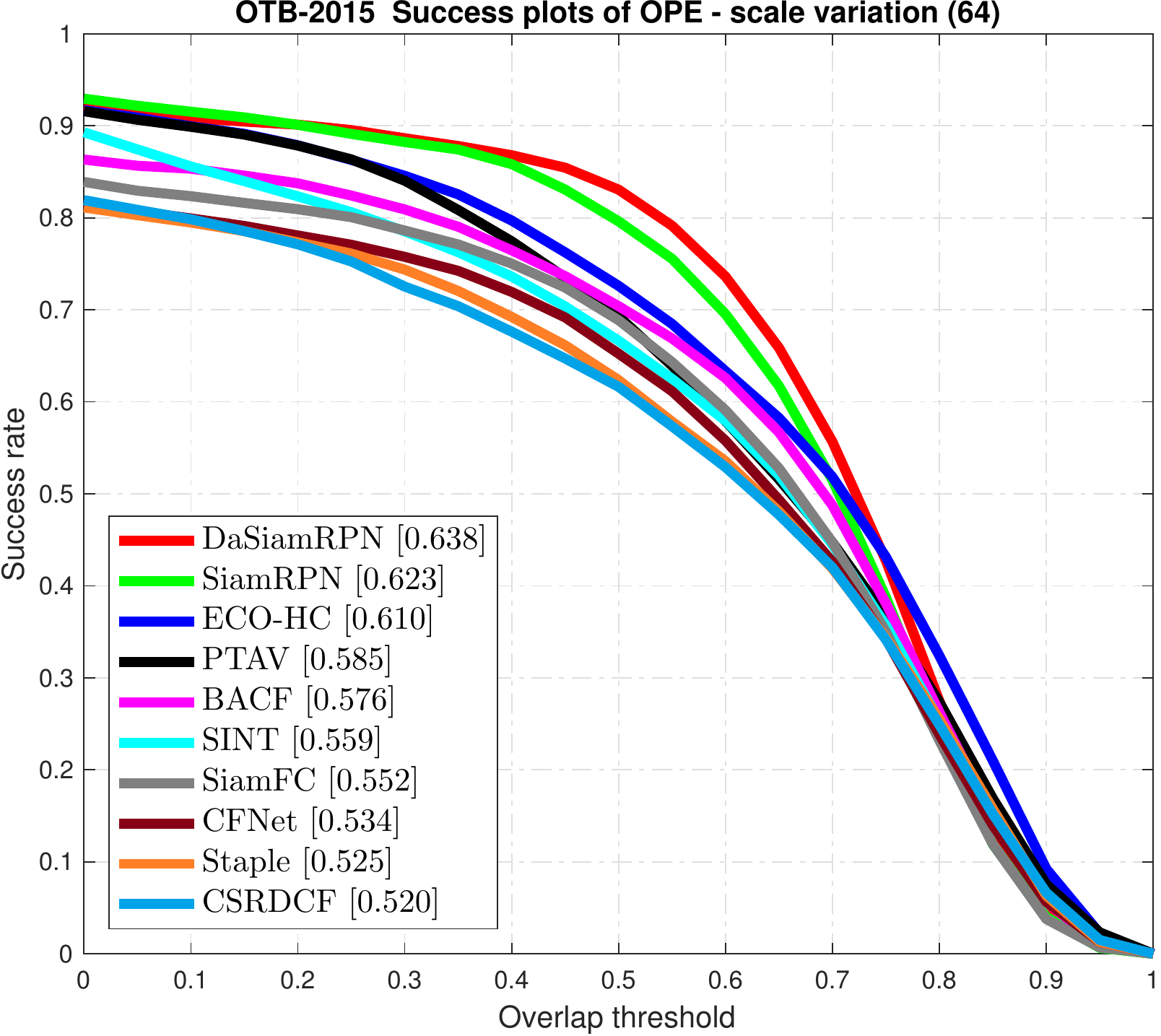}
\label{SV on OTB2015}}
\hfil
\subfloat[OV on OTB-2015]{\includegraphics[width=0.3\linewidth]{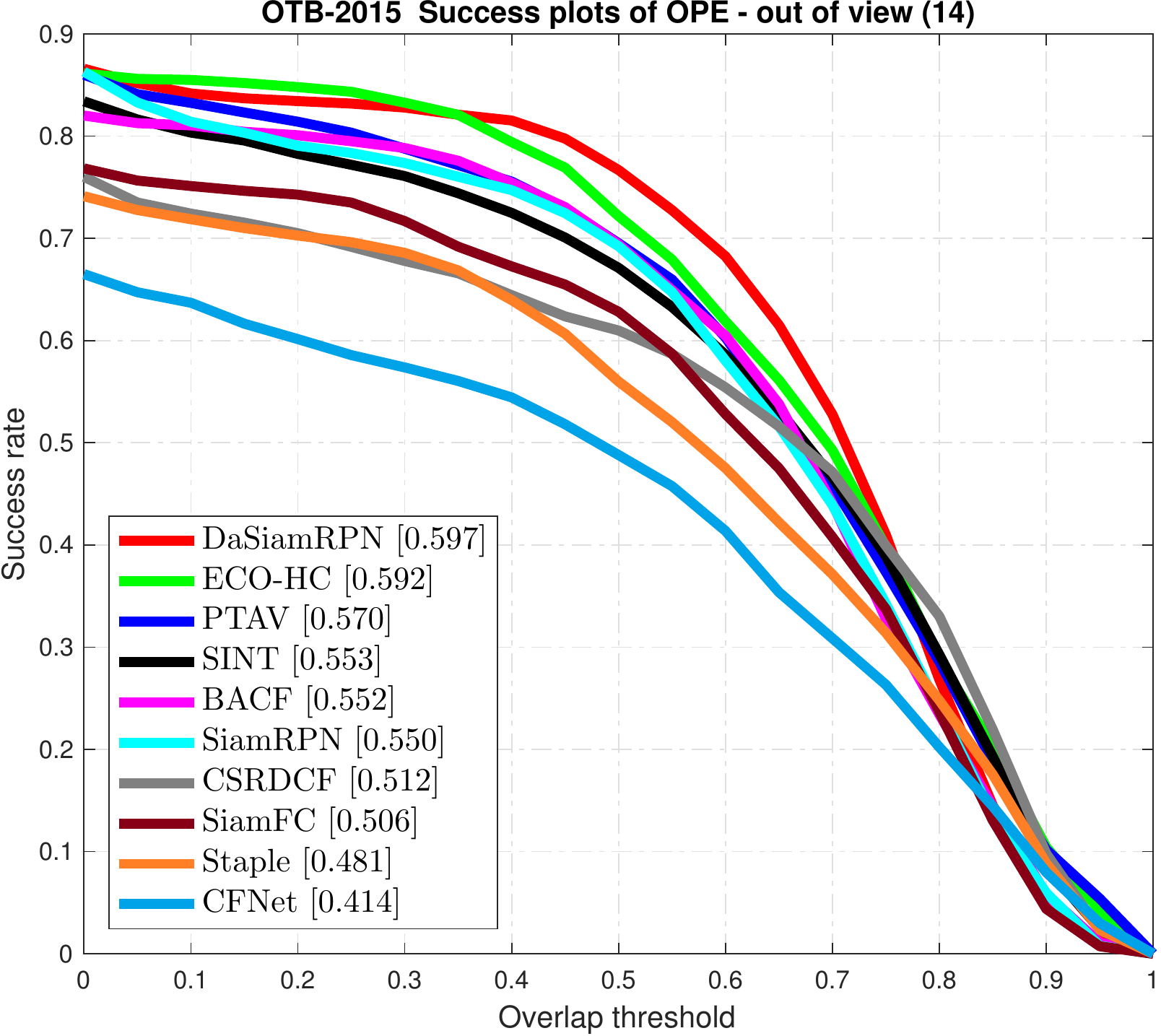}
\label{OV on OTB2015}}
\subfloat[OCC on OTB-2015]{\includegraphics[width=0.3\linewidth]{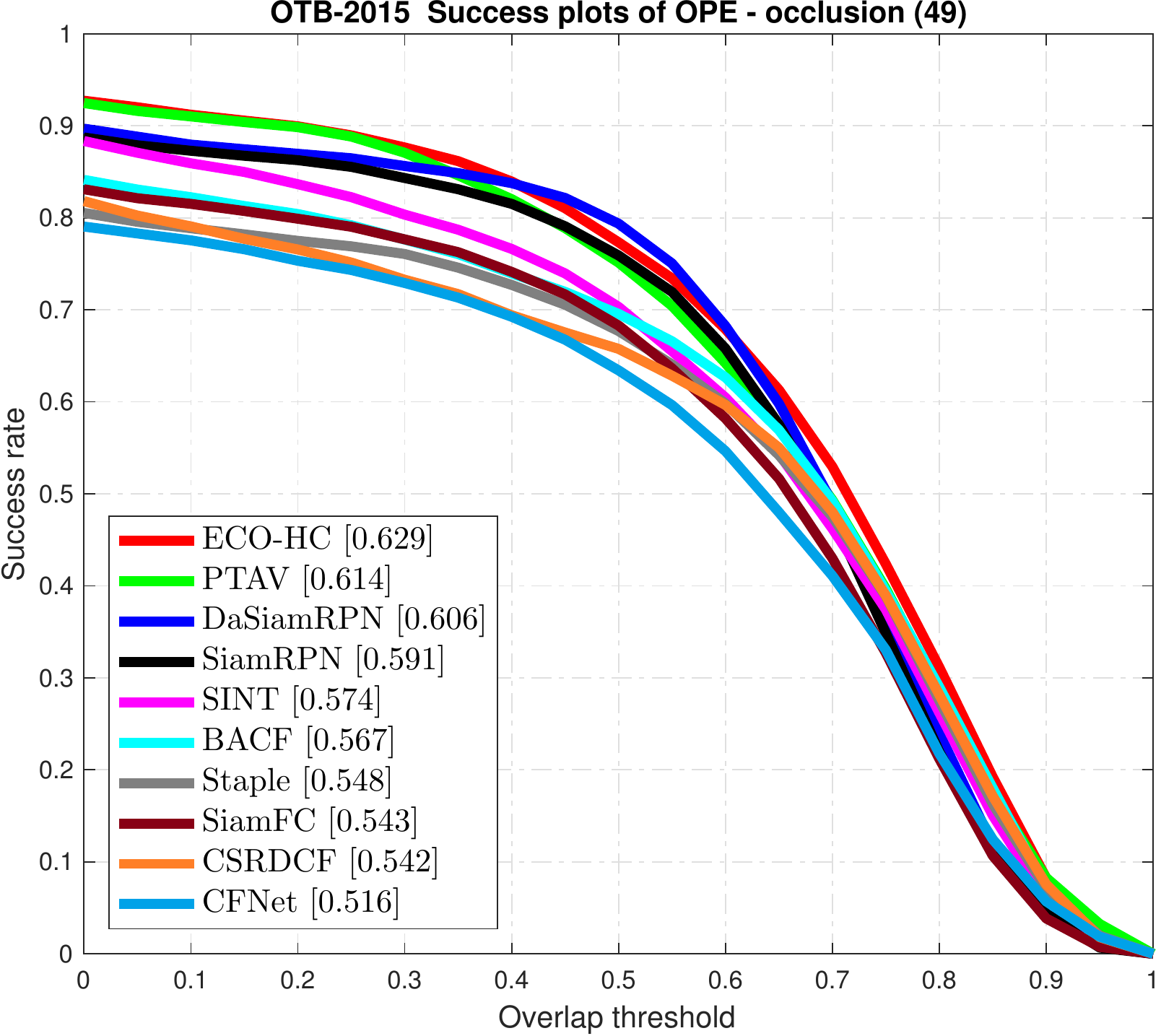}
\label{OCC on OTB2015}}
\subfloat[BC on OTB-2015]{\includegraphics[width=0.3\linewidth]{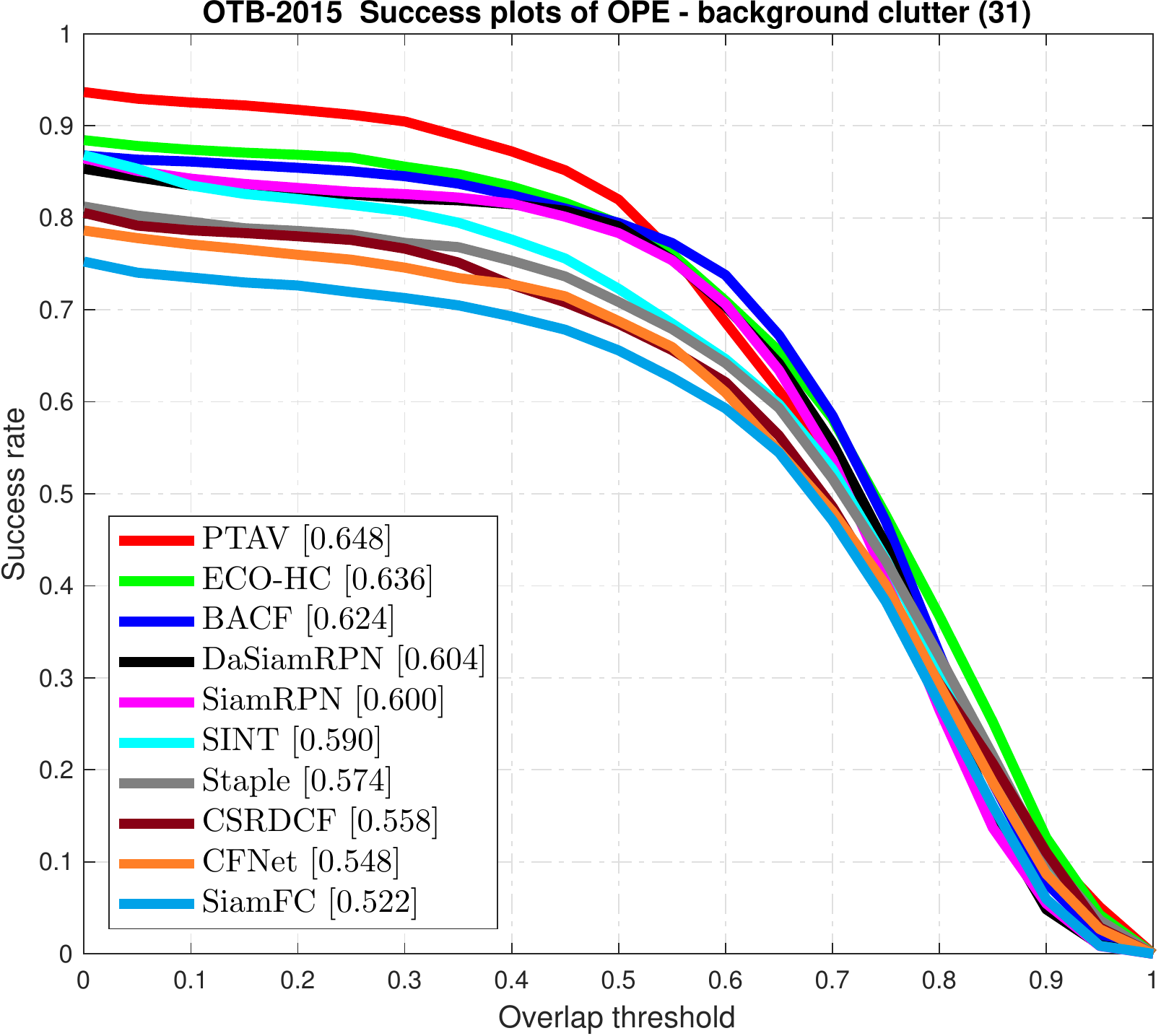}
\label{BC on OTB2015}}
\hfil
\subfloat[DEF on OTB-2015]{\includegraphics[width=0.3\linewidth]{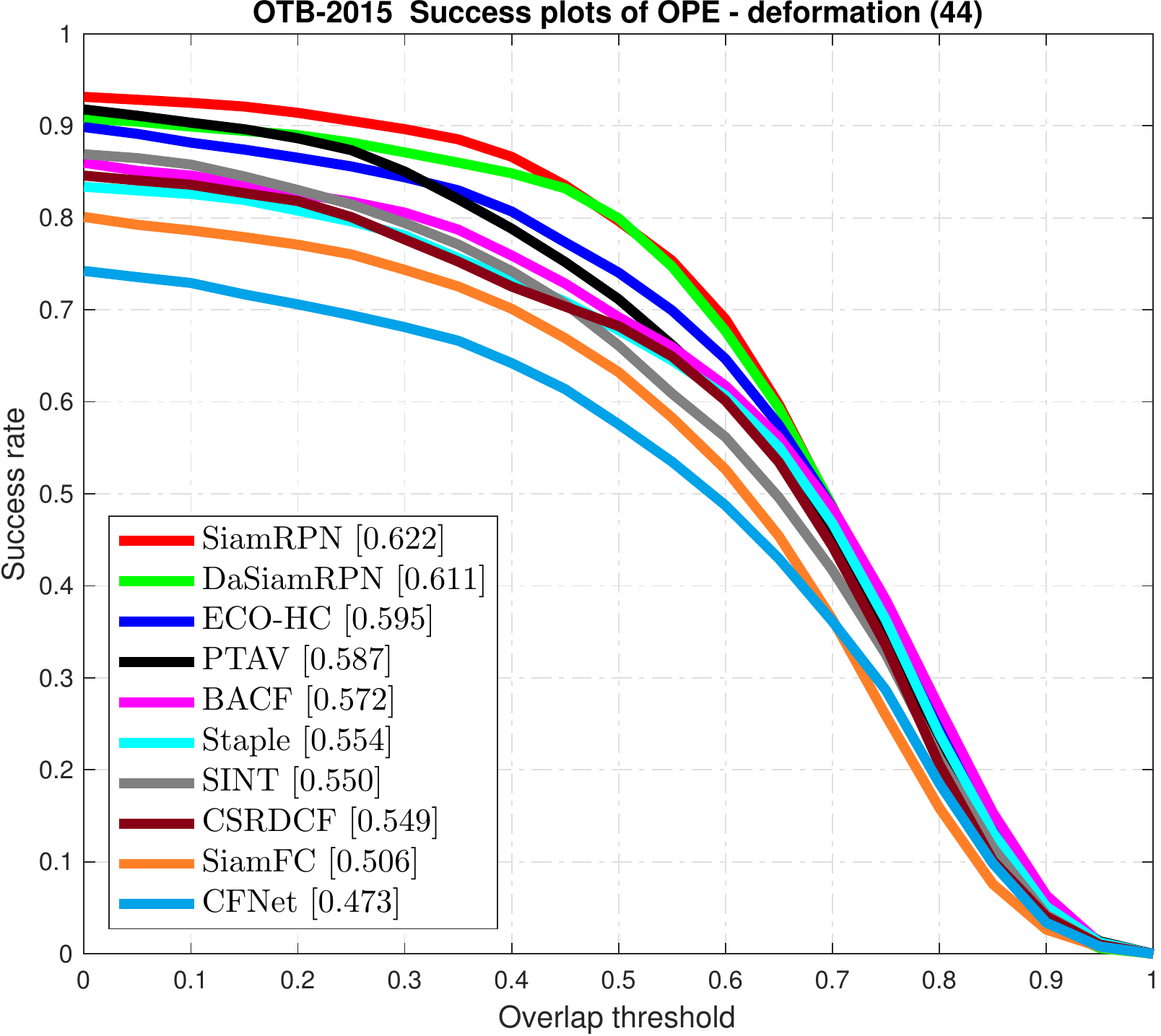}
\label{DEF on OTB2015}}
\subfloat[IV on OTB-2015]{\includegraphics[width=0.3\linewidth]{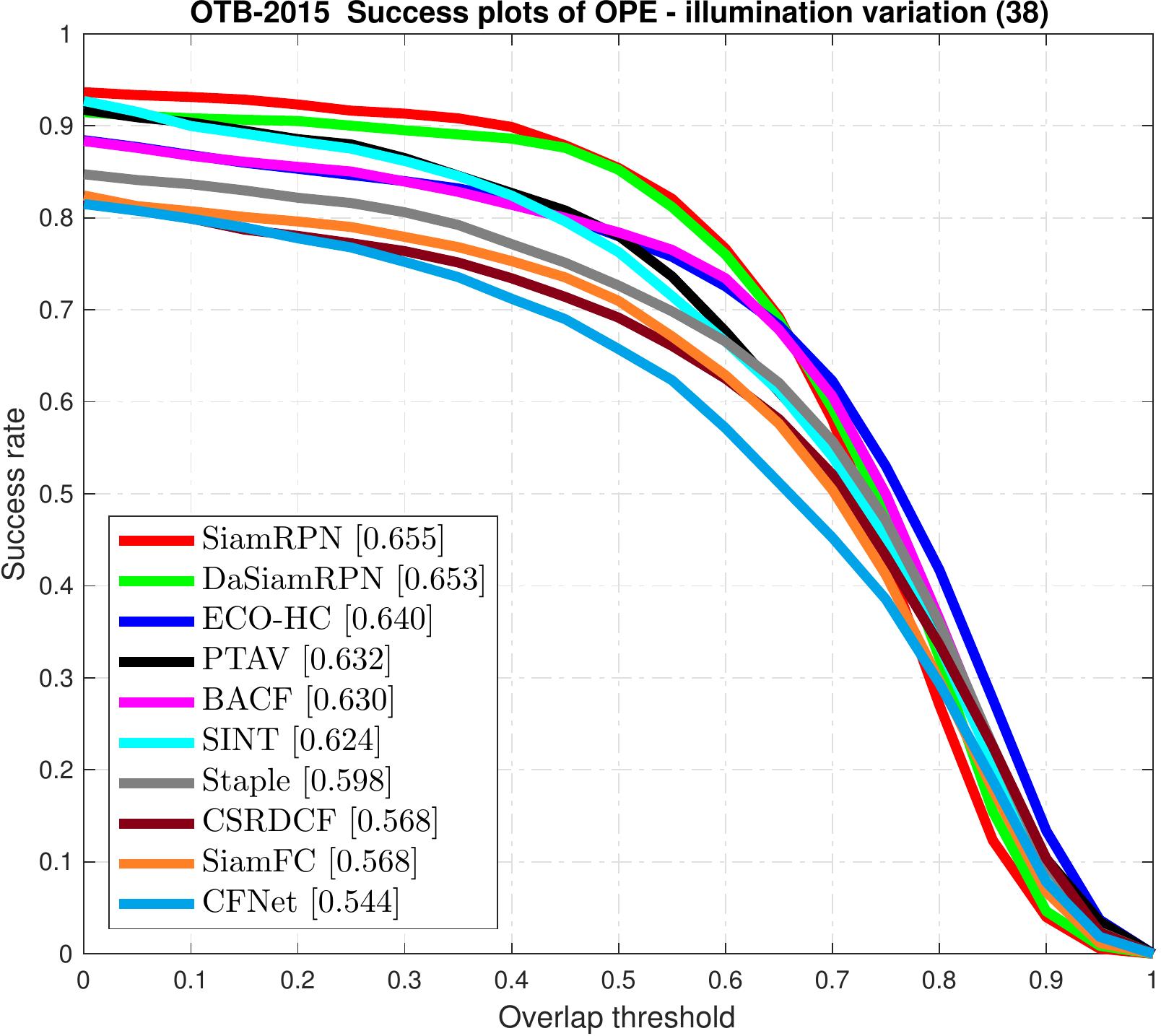}
\label{IV on OTB2015}}
\subfloat[LR on OTB-2015]{\includegraphics[width=0.3\linewidth]{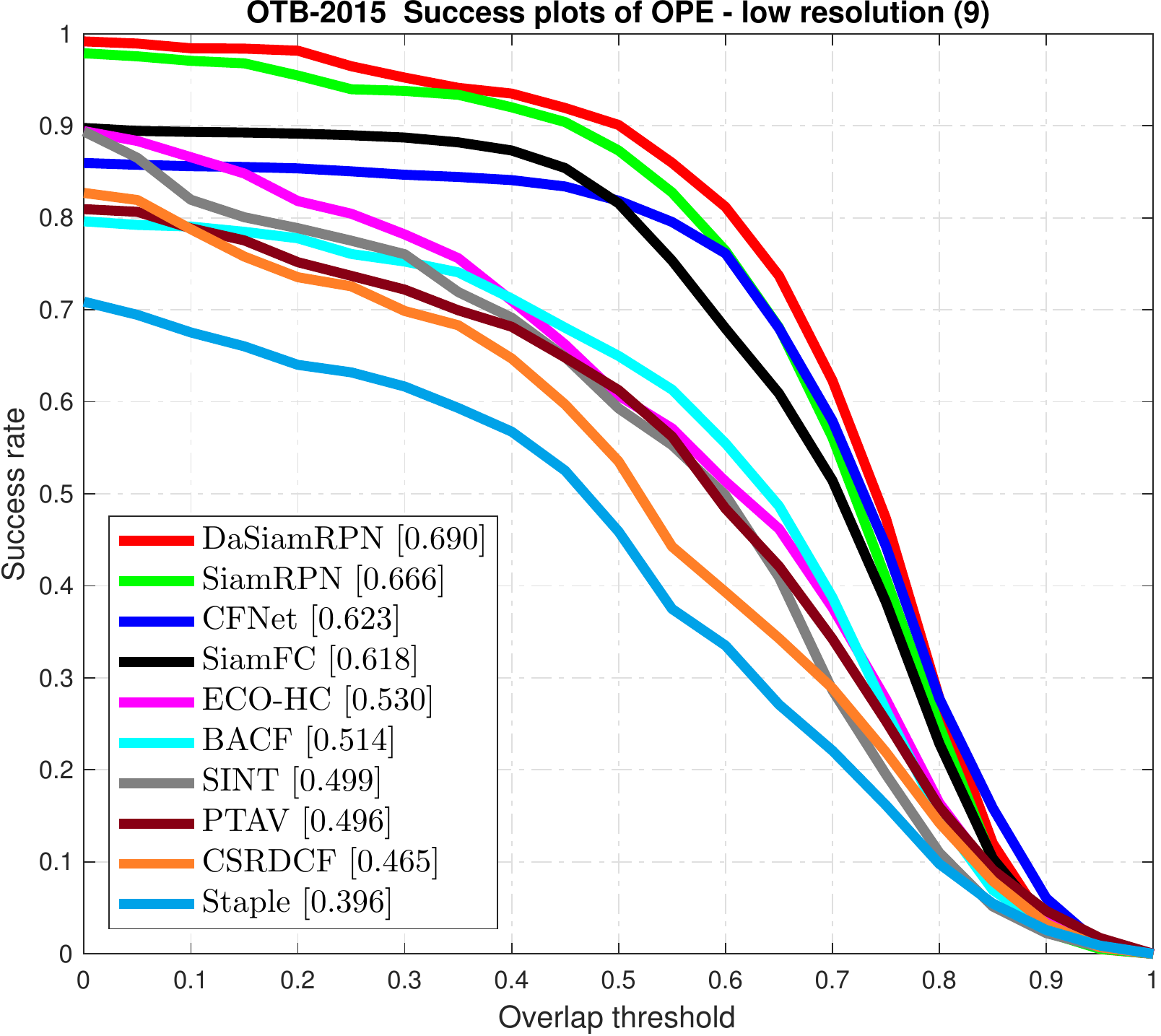}
\label{LR on OTB2015}}
\hfil
\subfloat[FM on OTB-2015]{\includegraphics[width=0.3\linewidth]{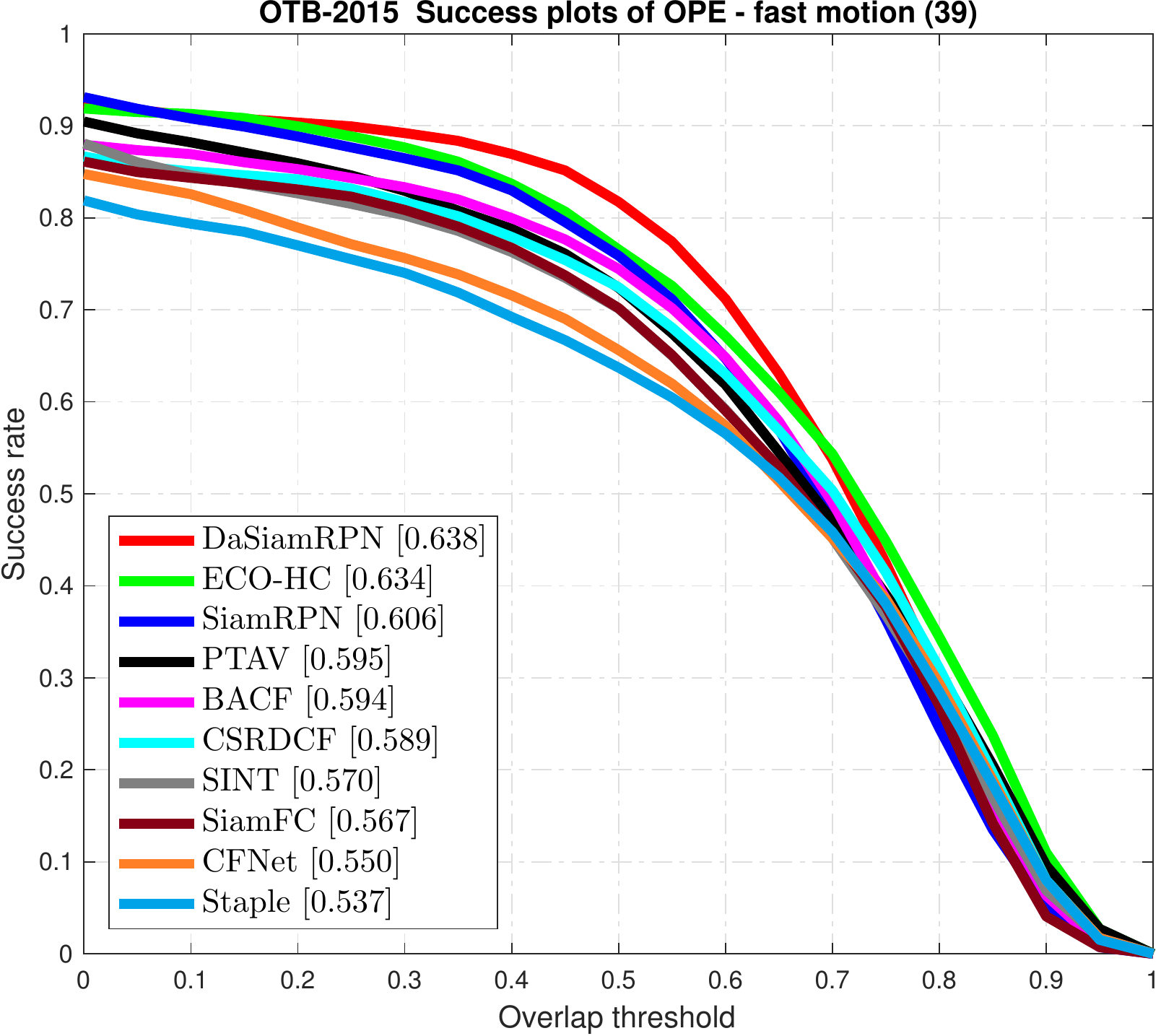}
\label{FM on OTB2015}}
\subfloat[MB on OTB-2015]{\includegraphics[width=0.3\linewidth]{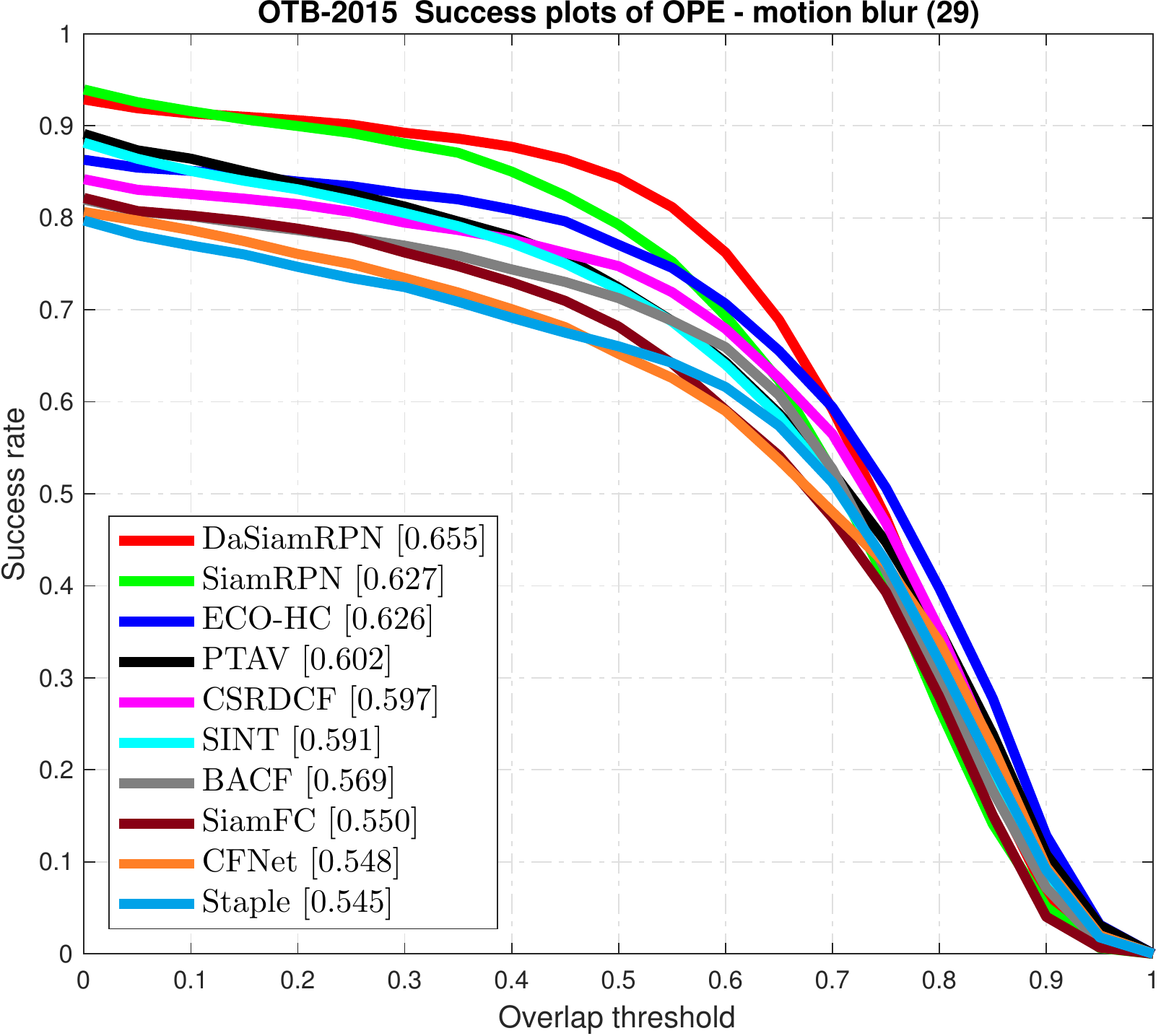}
\label{MB on OTB2015}}
\caption{The success plots on OTB-2015 for eleven challenge attributes: in-plain rotation, out-of-plane rotation, scale variation, out of view, occlusion, background clutter, deformation, illumination variation, low resolution, fast motion and motion blur.}
\label{fig:result2OTB}
\end{figure*}

In this section, detailed results on OTB are provided. Fig.~\ref{fig:result1OTB} and Fig.~\ref{fig:result2OTB} show the success plots for all 11 attributes on OTB2013 and OTB2015, respectively.

\subsection{Speed Analyses}
In the main paper, we report the tracking speed of our proposed DaSiamRPN on Titan X (160 FPS). We now provide some tracking time analyses additionally in Table~\ref{speed_table}. At the tracking stage, the computational bottleneck of our system is the cost of convolution operations existing in the forward process. However, our system is still more efficient than the traditional deep learning based trackers. The tracking speeds are reported on GTX 1060, GTX 1080, Titan X and Titan Xp.

\begin{table}[tp]
\scriptsize
  \centering
  \caption{Speed of the proposed tracker on different platforms.}
\begin{tabular}{ccccc}
\hline
\bf Trackers & \bf GTX 1060 & \bf GTX 1080 & \bf Titan X  & \bf Titan Xp\\
\hline
\textbf{SiamRPN}   &  160 FPS &  200 FPS     &  200 FPS     &  240 FPS\\\hline
\textbf{DaSiamRPN} &  130 FPS &  170 FPS     &  160 FPS     &  190 FPS\\\hline
\end{tabular}
   \label{speed_table}
\end{table}

\subsection{Qualitative Results}
\begin{figure}[htbp]
\centering
\includegraphics[width=1\linewidth]{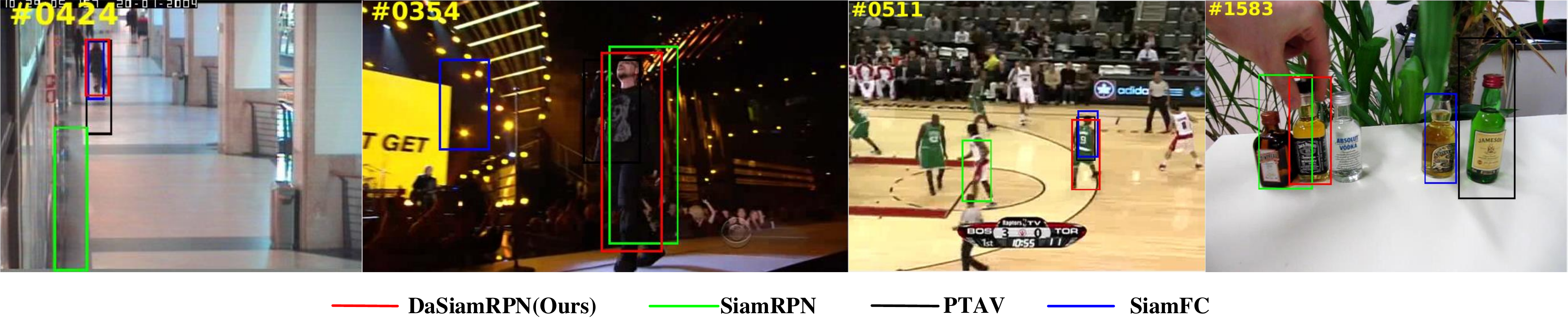}
\caption{The qualitative results of the DaSiamRPN and compared trackers.}
\label{vis}
%\vspace{-0.5cm}
\end{figure}

To visualize the superiority of the proposed framework, we show examples of the DaSiamRPN results compared to recent trackers (SiamRPN, SiamFC and PTAV) on challenging sample videos. As shown in Fig.~\ref{vis}, the proposed DaSiamRPN can handle the challenges while SiamRPN and SiamFC tend to drift to distractor. PTAV adopts long-term component, but fails to track in the second and last videos. The superiority performance of our algorithm can be attributed to the design of distractor-aware Siamese networks.% and local-to-global search strategy.

\end{document}